\documentclass{article}

\usepackage[preprint]{corl_2026}

\usepackage{algorithm}
\usepackage{algpseudocode}
\usepackage{booktabs}
\usepackage{graphicx} 
\usepackage{subcaption}
\usepackage{amsmath,amsfonts,bm}
\usepackage{multicol}
\usepackage{wrapfig}

\newcommand{\mT}{\mathbf{T}}
\newcommand{\mR}{\mathbf{R}}
\newcommand{\vp}{\mathbf{p}}

\usepackage{amsmath,amsfonts,bm}

\def\eqref#1{equation~\ref{#1}}

\def\1{\bm{1}}

\def\va{{\bm{a}}}

\def\vo{{\bm{o}}}
\def\vp{{\bm{p}}}

\def\mK{{\bm{K}}}

\def\mR{{\bm{R}}}

\def\mT{{\bm{T}}}

\DeclareMathAlphabet{\mathsfit}{\encodingdefault}{\sfdefault}{m}{sl}
\SetMathAlphabet{\mathsfit}{bold}{\encodingdefault}{\sfdefault}{bx}{n}

\DeclareMathOperator*{\argmax}{arg\,max}
\DeclareMathOperator*{\argmin}{arg\,min}

\title{What Matters When Cotraining Robot Manipulation Policies on Everyday Human Videos?}

\usepackage{authblk}


\author[1,$\dagger$]{Richard Li}
\author[2]{Aditya Prakash}
\author[1]{Andrew Wen}
\author[2]{Saurabh Gupta}
\author[3,$\star$]{Yilun Du}
\author[1,$\star$]{Pulkit Agrawal}
\affil[1]{Massachusetts Institute of Technology}
\affil[2]{University of Illinois Urbana-Champaign}
\affil[3]{Harvard University}

\date{}

\begin{document}
\maketitle
\renewcommand{\thefootnote}{\ifcase\value{footnote}\or\ensuremath{\star}\or\ensuremath{\dagger}\fi}
\setcounter{footnote}{0}
\footnotetext[1]{Equal advising.}
\footnotetext[2]{Corresponding author. Contact \texttt{rli14@mit.edu}}
\renewcommand{\thefootnote}{\arabic{footnote}}
\vspace{-8mm}
\begin{abstract}
Human video datasets used for cotraining robot manipulation policies largely consist of curated demonstrations where motions are orchestrated to resemble robot behavior and 3D hand poses are captured with specialized hardware. A more plentiful source of data is everyday Internet video, but it is an open question what factors enable transfer from such videos to robots. We investigate this using a new dataset of 532 human videos with 28 hours of high-quality triangulated hand labels and natural motions. We find that hand pose quality affects transfer, but even with accurate hands, the inherent motion gap hinders transfer unless the vision and policy networks specialize to each embodiment. Our cotraining recipe yields consistent improvements, with an absolute success rate gain of $29.7\%$ in the low-robot-data regime across six manipulation tasks.
\end{abstract}


\section{Introduction}
\label{sec:intro}

A central challenge in building generalist robotic foundation models is data scarcity. Teleoperated demonstrations are expensive and slow to collect, while reinforcement learning remains difficult to scale due to challenges in reward design, exploration, and sim-to-real transfer. In contrast, Internet videos of humans performing everyday activities are abundant and offer a potentially scalable route to physical intelligence.

\begin{figure}[t]
    \centering
    \includegraphics[width=\textwidth]{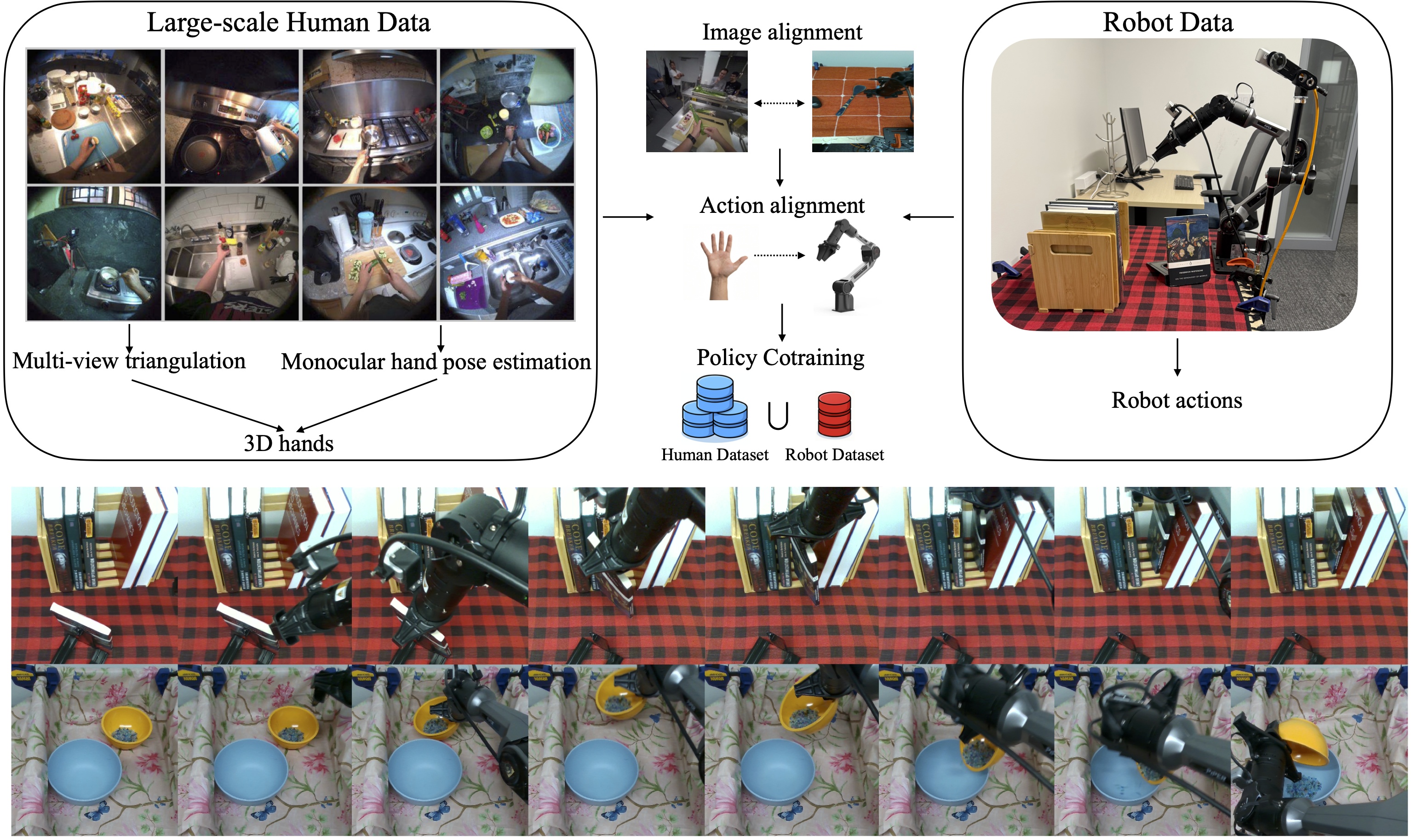}
    \vspace{-0.5em}
    \caption{\emph{Top:} System diagram showing data processing and policy cotraining steps. \emph{Bottom:} Rollouts from cotrained policy manipulating unseen objects in unseen scenes. See interactive visualizations and videos: \url{https://richardrl.github.io/what-matters-cotraining-human-videos/}.}
    \label{fig:teaser}
    \vspace{-1em}
\end{figure}

Prior work on learning from human video largely falls into two categories. The first uses Internet videos of everyday activities but relies on modular pipelines that decompose a policy into multiple components or stages \cite{shi2025zeromimic, bharadhwaj2024interactionplans,bharadhwaj2024track2act}. The second achieves human-to-robot transfer by cotraining end-to-end policies, but relies on \emph{aligned data}, where human demonstrations are carefully orchestrated to match robot motion, identical cameras are used, and specialized hardware provides accurate 3D hand pose \cite{kareer2024egomimic, punamiya2025egobridge}. This setting differs fundamentally from Internet video, which exhibits unconstrained motion, heterogeneous cameras, task mismatch, and noisy hand pose estimates.

A rigorous empirical study of cotraining on Internet video is difficult because ground-truth 3D hand labels are unavailable, and cotraining pipelines can break down with poor labels. We therefore study cotraining in a proxy setting that preserves the key challenges of Internet video---egocentric views, natural motions, camera and task mismatch---while using extrinsic cameras to triangulate high-quality hand labels. Our goal is not to produce a method directly applicable to Internet video today, but rather to understand \emph{what factors can enable} this regime, focusing on two key sources of difficulty: (1) 3D hand pose quality, and (2) the action gap induced by natural human motion and task mismatch. By identifying the bottlenecks, future work can concentrate efforts towards lifting these bottlenecks, e.g. by scaling up training data for hand pose estimators.

To enable controlled study, we construct a dataset of everyday human videos with high-quality 3D hand poses by triangulating EgoExo4D \cite{grauman2024ego} with a multi-view pipeline. These labels upper bound what monocular hand pose estimation can achieve and can serve as training data for future monocular estimators. We isolate the role of hand quality by comparing our triangulated hands against a state-of-the-art monocular estimator \cite{zhang2025hawor}.

Even with accurate 3D hand poses, successful transfer requires careful alignment across human and robot data. We identify \emph{scale inconsistency in image space} as a previously overlooked issue when cotraining across datasets with different cameras. Simple image-space scale alignment significantly improves transfer, despite large differences in camera hardware. Additionally, we observe that standard cotraining recipes designed for lab data with smaller motion gaps yield little to no improvement over robot-only training given our natural motion human data. Through ablations, we show that the common design choices of using image bottlenecks and shared action encoders limit transfer. Thus, we propose a cotraining recipe that explicitly accounts for embodiment differences via token-level fusion, embodiment-specific action encoders and decoders, and upweighting robot data.

We evaluate our approach across six real-world manipulation tasks by training on 532 human videos and 3,000 robot demonstrations, and conduct 3,480 real-world rollouts. Our cotraining recipe significantly improves scaling behavior: human data provides the largest benefits in low-robot-data regimes (+29.7\% absolute improvement) and continues to help as task-specific robot data scales.


Therefore, our main contributions are:
\begin{itemize}
\item We curate a large-scale dataset of everyday human videos with accurate 3D hands, and show our dataset has better robot transfer than a large-scale lab dataset \cite{hoque2025egodex}.
\item We show that higher-quality 3D hand labels yield greater transfer, with a large gap between monocular-estimated and triangulated hands on the same videos.
\item We demonstrate the benefit of scale alignment in image space when cotraining across datasets with different cameras, a factor not currently accounted for in leading VLAs \cite{black2410pi0}.
\item We show that architectural and loss function decisions enabling embodiment-specific specialization are necessary for significant transfer from natural motion human data, whereas design choices effective for lab demonstrations do not transfer to natural motion videos.
\end{itemize}

Our overall contribution is to chart a path toward robot cotraining on Internet video by identifying and addressing the key challenges---hand quality and natural motion---that distinguish such data from lab demonstrations.

\section{Related Work}

\newcommand{\boldparagraph}[1]{\noindent{\bf #1:}}\vspace{0.2cm}

\boldparagraph{Egocentric datasets with hand-object interaction (HOI)}
Large-scale egocentric video collections fall into two camps: datasets with accurate hand/head poses but limited motion and scene diversity, such as ARCTIC and EgoDex \cite{fan2023arctic, hoque2025egodex}, and natural-motion datasets with poor hand labels, such as HoloAssist \cite{wang2023holo}, EpicKitchens \citep{damen2018scaling}, Ego4D \citep{grauman2022ego4d}, and EgoExo4D \citep{grauman2024ego}. We build upon EgoExo4D by significantly processing and cleaning the hands to produce a new dataset with natural motions and clean hand labels.

\boldparagraph{Modular pipelines using HOI videos}
Existing works exploit hand motion from Internet videos through modular pipelines that learn intermediate representations such as 3D poses \cite{srirama2024hrp,kannan2023corl}, affordances \cite{shi2025zeromimic,srirama2024hrp,mendonca2023worldmodels,bahl2023vrb,kannan2023corl, goyal2022human, chang2023vidm}, or 2D tracks \cite{bharadhwaj2024track2act,bharadhwaj2024interactionplans,wen2024atm}. For example, \cite{shi2025zeromimic} learns visual affordances from human Internet videos for post-grasp policies, and \cite{bharadhwaj2024interactionplans} generates keyframe-level interaction plans to condition robot policies. We hypothesize the lack of strong 3D hand pose estimators until recently led to this focus on modularity over end-to-end cotraining.

\boldparagraph{Co-training policies with hand and robot data}
Another recent paradigm directly cotrains policies on both human and robot data, requiring 3D hand motion labels from lab datasets~\cite{yang2025egovla} or custom setups and data collection devices~\cite{hoque2025egodex,kareer2024egomimic,tao2025dexwild}. Egomimic and Egobridge \cite{kareer2024egomimic,punamiya2025egobridge} study techniques for human-robot transfer but rely on small-scale lab data with orchestrated motions, and make assumptions that break as the motion gap grows (Sec.~\ref{sec:results}, \cite{punamiya2025egobridge}). Other works scale up cotraining but use data with limited scene diversity \cite{qiu2025humanoid,hoque2025egodex} or noisy hand labels that yield minimal transfer \cite{yang2025egovla,wang2023holo}. Our work differs in that we explicitly investigate what can enable cotraining on Internet videos.


\section{TriHands Dataset}
\label{sec:trihands}
The main difficulty in studying the impact of hand label quality is the lack of a dataset of everyday human videos with clean hands. We build such a dataset by multiview triangulating such hands on 532 cooking videos from the EgoExo4D dataset \cite{grauman2024ego}. 

The 2D keypoints used by the EgoExo4D team \cite{sengupta2020mm} are the primary bottleneck. We found that 3D model-based pose estimators \cite{MANO:SIGGRAPHASIA:2017} project to much more accurate 2D keypoints in poor lighting conditions and when the hand partially leaves the field-of-view, which are both common in egocentric and in-the-wild videos (Supp. \ref{supp:keypoints_comparison}). Additionally, 2D hand keypoint estimators degrade significantly under self-occlusion, whereas 3D hand estimators always project to a complete set of 2D keypoints.

Given a strong 3D base model, an out-of-distribution dataset can be accurately annotated using a small number of partial (i.e. the visible subset of hand joints) 2D keypoint labels. We fine-tune the hand pose estimator with a reprojection loss, $\mathcal{L}_{\text{proj}} = \sum_{i \in \mathcal{I}} \lVert \vp_i^{2\mathrm{D}} - \pi_{\mK}\!\big(\hat{\vp}_i^{3\mathrm{D}}\big) \rVert_{1}$, where $\pi_{\mK}$ is the projection operator with intrinsics $\mK$, and $\mathcal{I}$ is the set of joints with 2D annotations. Fine-tuning is performed with this loss on 4407 frames with partial 2D keypoint labels provided by EgoExo4D, which represents less than $0.144\%$ of the final right-hand triangulated frames we can produce.

We triangulate 3D hands using the egocentric and $M \in \{4,5\}$ exocentric views via DLT \cite{hartley2003multiple} with nonlinear refinement, picking the hand that maximizes camera count subject to reprojection error thresholds (details in Supp. \ref{supp:trihands_triangulation_pipeline}). After filtering and 0.4s linear interpolation, we obtain 3,042,406 right-hand frames at 30fps (over 28 hours of training data).

\begin{figure}[t]
  \centering
  \includegraphics[width=\textwidth]{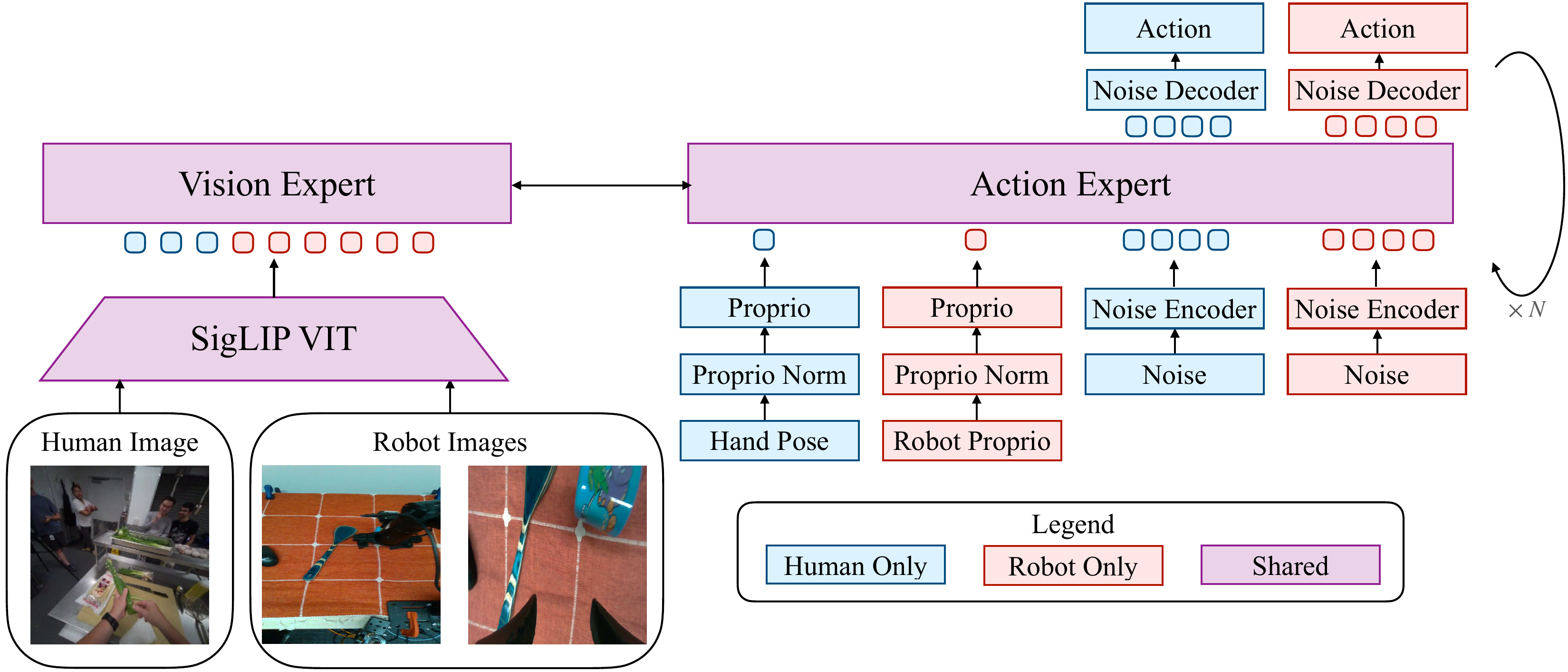}
  \caption{Input-output diagram for inference over human and robot data in our conditional flow matching architecture. Images shown with image-space alignment and resize. Color denotes weight sharing and $N$ denotes inference steps. See Supp. \ref{supp:transfusion_arch}.}
  \label{fig:architecture_diagram}
\end{figure}

\section{Framework}

Our framework for cotraining policies on human videos and robot trajectories balances two competing goals: encouraging positive transfer by aligning observations and actions (Sec.~\ref{sec:pixel_space_alignment},~\ref{sec:action_space_alignment}), while preventing harmful representation alignment by allowing the network to specialize to each embodiment (Sec.~\ref{sec:cross_embodiment_arch},~\ref{sec:cotraining_algorithm}). We adopt the spatial-algebra notation of \cite{manipulation}, where a homogeneous transform
${}^{b}\mT{}^{a}
=
\begin{bmatrix}
{}^{b}\mR{}^{a} & {}^{b}\vp{}^{a} \\
\mathbf{0}^\top & 1
\end{bmatrix}$
represents frame $a$ with respect to, and expressed in, frame $b$.

\subsection{Data streams}
We use images from an egocentric moving head-camera for the human data and we use a similarly-positioned egocentric camera for the robot data (Fig. \ref{fig:teaser}). The cameras have significantly different sensor sizes and focal lengths, which makes transfer difficult (Sec. \ref{sec:pixel_space_alignment}, Supp. \ref{supp:camera_specs}).

The human raw actions come from either triangulation (Sec. \ref{sec:trihands}) or monocular hand pose estimation (Sec. \ref{sec:experimental_design}) and must be transformed into a the robot action space to enable cotraining. The robot is a 6-DOF AgileX Piper arm with a parallel-jaw gripper, and its action space is the tool center point (TCP) pose and a discrete grasp command. Concretely, we define ${}^{R}\va = \bigl({}^{\text{camera}}\vp^{\text{TCP}}, {}^{\text{camera}}\mR^{\text{TCP}}, g\bigr)$, where $g$ is a discrete ternary variable representing \emph{open}, \emph{close}, and \emph{no-op} grasp commands.

\subsection{Image-space scale alignment}
\label{sec:pixel_space_alignment}
It is common for the cameras used in the robot and human datasets to be significantly different in terms of field of view and focal length. This motivates the need to align the image-space scale of objects in order to learn shared features. The pinhole equation $Z = \frac{f \Delta X}{\Delta u}$ shows that for a fixed 3D object extent $\Delta X$, the pixel extent $\Delta u$ is determined by the focal length $f$ and depth $Z$, so cameras with different focal lengths produce different image-space scales for an object that is the same distance away from the cameras.

We undistort the human fisheye images to a pinhole projection \cite{kannala2004generic,szeliski2022computer} and place the robot camera at $z_{\text{human}}$ from the workspace, where $z_{\text{human}}$ is the median wrist depth in the human dataset. However, since the robot camera FOV is much smaller, this distance causes the image to lose scene context. We therefore adjust the robot camera extrinsics: rotating it $90^{\circ}$ clockwise so the wider FOV captures the vertical extent of the scene\footnote{Hand-object interactions in the human dataset tend to occur towards the bottom of the image.}, and translating it to depth $z_{\text{robot}} = z_{\text{human}} \frac{f_s H_c}{f_c H_s}$, where $H_c, H_s$ are image heights and $f_c, f_s$ the focal lengths (Supp. \ref{supp:pix_extrinsic}). This approximately matches image-space object scale while retaining the 3D scene extent, and we show in Sec. \ref{sec:results} that it is crucial for transfer.

\subsection{Action space alignment}
\label{sec:action_space_alignment}

\begin{wrapfigure}{r}{0.4\textwidth}
  \vspace{0em}
  \centering
  \begin{subfigure}{0.18\textwidth}
    \centering
    \includegraphics[width=\linewidth]{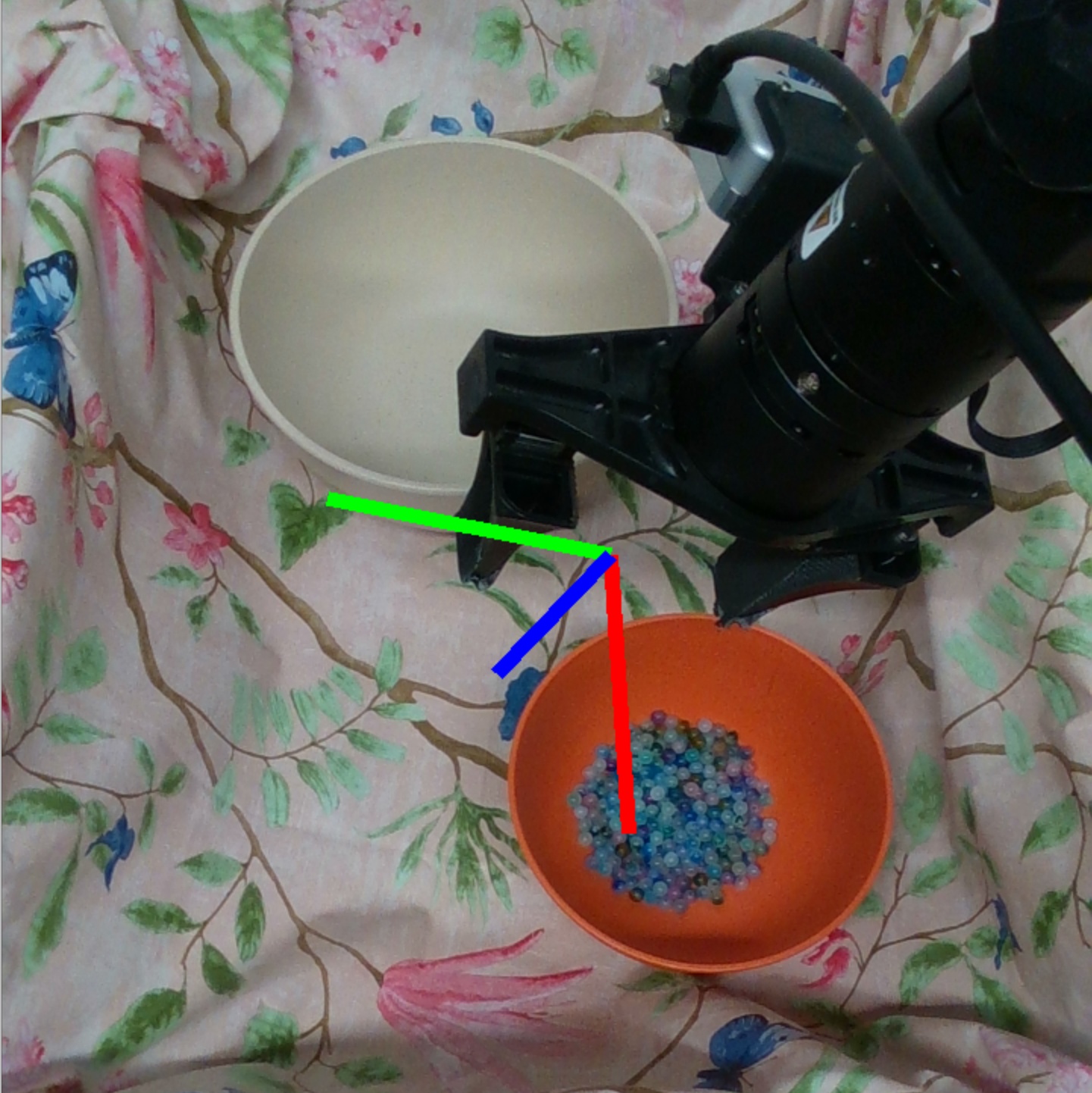}
    \caption{Piper TCP.}
    \label{fig:agilex_tcp}
  \end{subfigure}
  \hspace{0.3em}
  \begin{subfigure}{0.18\textwidth}
    \centering
    \includegraphics[width=\linewidth]{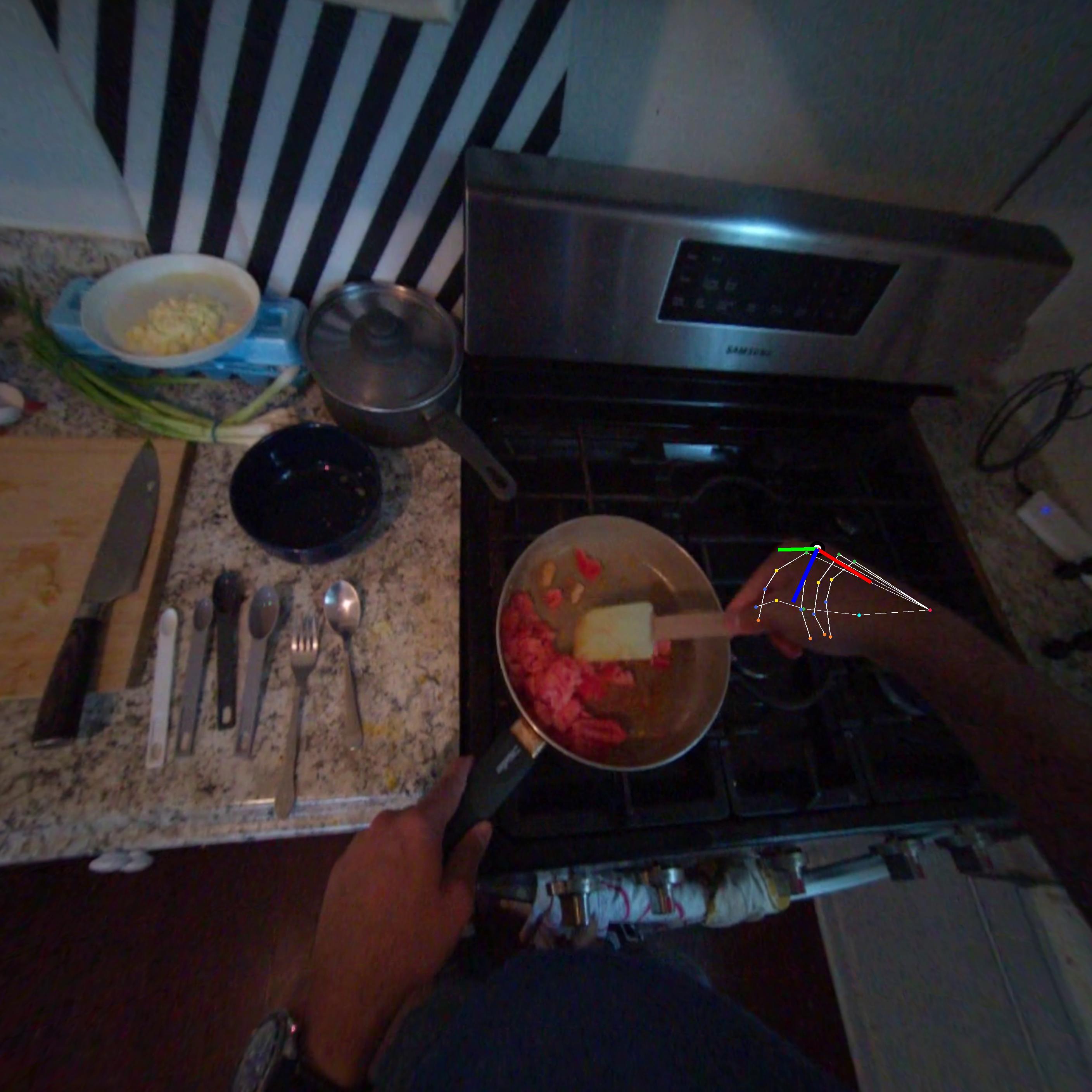}
    \caption{MANO TCP.}
    \label{fig:mano_gripper}
  \end{subfigure}
  \vspace{0.5em}
  \caption{Robot TCP frame vs. human TCP frame (not to scale).}
  \label{fig:mano_tcp}
  \vspace{-1em}
\end{wrapfigure}

For the network to learn shared representations across embodiments, the human and robot action marginals must share the same support. We map the human 3D hand pose to the robot action space by picking the ``middle finger proximal'' MANO joint frame with a rotation to match the robot TCP frame (Fig. \ref{fig:mano_tcp}), chosen because it best aligns contacts compared to the wrist frame. The remapped human action space in the current camera frame is ${}^H\va = \bigl( {}^{\text{camera}}\vp^{\text{middle}},\; {}^{\text{camera}}\mR^{\text{middle}}\!~\!~{}^{\text{middle}}\mR^{\text{TCP}},\; g_{\text{no-op}} \bigr)$.

Since imitation learning methods predict action chunks \cite{chi2023diffusion,black2410pi0}, we cannot use the above equation directly because head camera motion introduces extreme multimodality. We stabilize hand actions by transforming per-timestep camera-frame poses to the current head frame. The head poses for TriHands come from EgoExo4D \cite{grauman2024ego}, but could also be estimated from monocular video \cite{teed2021droid,maggio2025vggt,zhang2025hawor}. The human action chunk for observation $\vo_t$ is $\va_t, \ldots, \va_{t+H} = {}^{c_t}\!\va^{t}, \ldots, {}^{c_t}\!\mT^{w}\, {}^{w}\!\mT^{c_{t+H}}\, {}^{c_{t+H}}\!\va^{t+H}$.

Despite alignment, a bias remains between the two action distributions: human hand motions are biased towards a camera region corresponding to hand chirality, and camera placement differences cause differences in action depths (Supp. \ref{supp:marginal_dist}). We resolve this by mean-centering each dataset independently, filtering to the 1st--99th percentiles, and scaling to $\pm1$. Rotations need no normalization since $R \in \mathrm{SO}(3) \Rightarrow R_{ij} \in [-1,1] \;\forall i,j \in \{1,2,3\}.$

\subsection{Cross-embodiment architecture}
\label{sec:cross_embodiment_arch}
For an ideal cross-embodiment policy, for human observation $o_h$ and robot observation $o_r$ of the same object in the same pose, we would like to learn an invariant function $f_{\theta}(o_h) = f_{\theta}(o_r) = z$ mapping to the embodiment-agnostic object state. If the encoder succeeds, a shared deterministic decoder would be forced to output $g_{\phi}(z) = {}^R\va = {}^H\va$. This mapping is guaranteed to fail given the large motion gap in our TriHands dataset, where tasks and motions are natural and unconstrained. Thus, embodiment-specific weights are necessary.

Our Transfusion-inspired architecture uses separate FFN experts for visual and action tokens \cite{black2410pi0, zhou2024transfusion} (Fig. \ref{fig:architecture_diagram}, Supp. \ref{supp:transfusion_arch}). It retains the full grid of ViT \cite{dosovitskiy2020image} image tokens and concatenates them with noisy action tokens via shared attention (\textit{token-level fusion}). Embodiment-specific 3-layer MLPs serve as the action and proprioception encoders and decoders.

\subsection{Cotraining algorithm and weighted loss}
\label{sec:cotraining_algorithm}
To prevent harmful representational alignment, we implicitly upweight the influence of the robot data. For each minibatch, we sample an equal number of samples independently from each embodiment. This scheme is equivalent to weighting the robot loss by $\frac{N_R + N_H}{N_R}$ and the human loss by
$\frac{N_R + N_H}{N_H}$
relative to a uniform sampling strategy. Because the robot dataset is much smaller than the human dataset, this amplifies the influence of the robot data. See Supp. \ref{supp:cotraining_alg}.

\section{Experimental Design}
\label{sec:experimental_design}

We aim to validate four hypotheses: \textbf{H1}: Cotraining on human data increases zero-shot generalization to new objects and backgrounds. \textbf{H2}: Human data transfers motion knowledge, instead of only coarse visual features. \textbf{H3}: Higher hand quality leads to greater transfer. \textbf{H4}: Allowing the network to specialize to each embodiment leads to improved robot task performance.

\begin{wrapfigure}[17]{r}{0.45\textwidth}
  \vspace{-1.5em}
  \centering
  \includegraphics[width=0.43\textwidth]{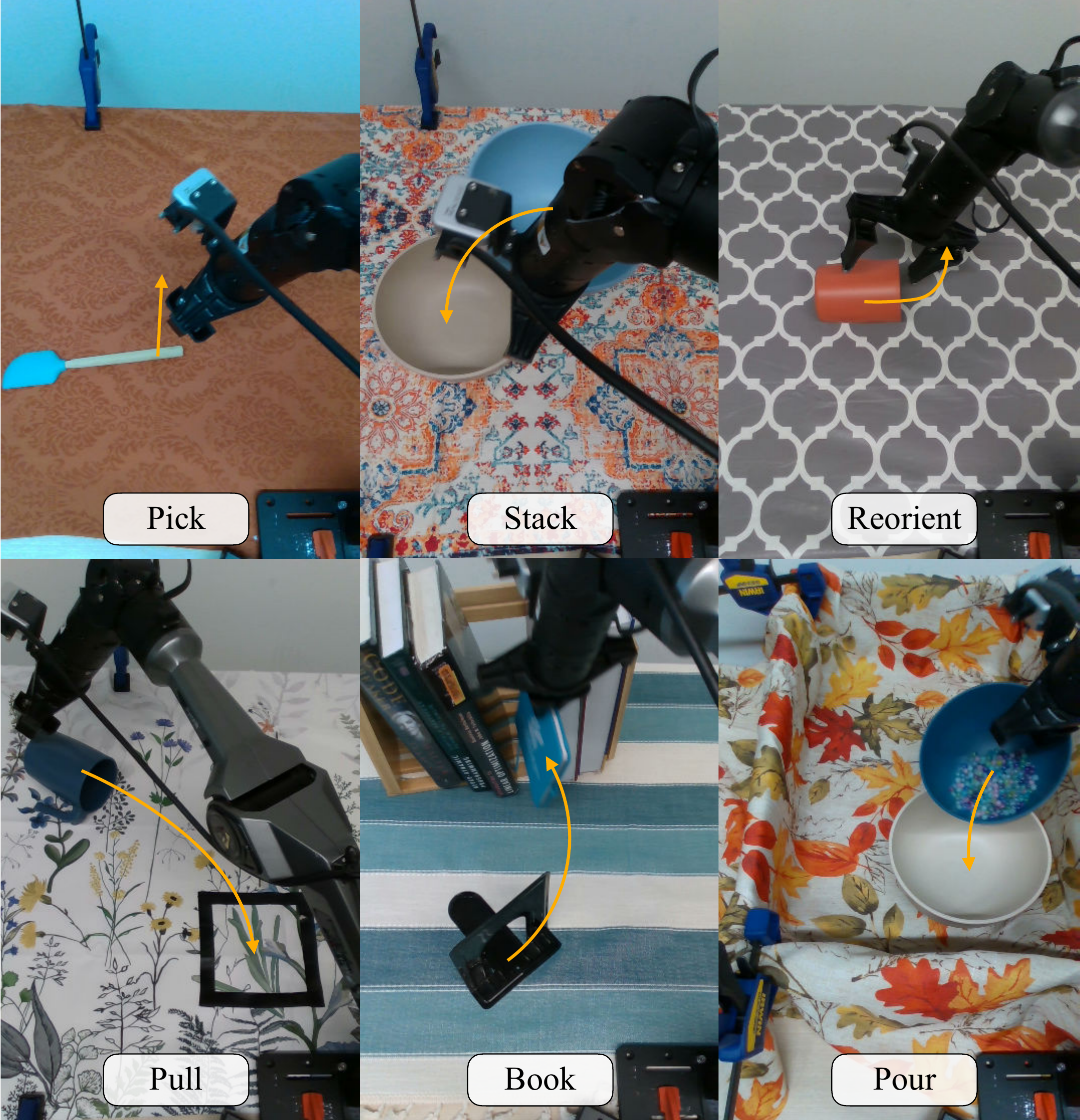}
  \vspace{-0.5em}
  \caption{\textbf{Task visualizations.} Orange arrow shows representative object motion.}
  \label{fig:task_image_kn}
  \vspace{-1em}
\end{wrapfigure}

\textbf{Tasks and environments}:  To validate cotraining transfer, we would like to see consistent transfer across a range of tasks. We provide results on six tasks (Fig. \ref{fig:task_image_kn}), ranging from pick-and-place to more intricate tasks requiring complicated rotations and/or reaching the limit of the robot's kinematic workspace. To test generalization (\textbf{H1}), we follow prior work \cite{hu2024data} and focus on the controlled setting of within-category object generalization and background generalization. Each environment thus consists of a unique object and background. See Supp. \ref{supp:tasks}.

\textbf{Data sources:}
\label{sec:exp_design_cotraining} Following data guidelines from prior work on robot scaling laws \cite{hu2024data}, we collect 50 demonstrations per environment. For each task, we have 10 training environments, resulting in a total of 500 demos per task. For approaches where we cotrain with human data in addition to the robot data, we always use RGB images sampled from all 532 videos from the TriHands dataset described above. The hand actions come from triangulation (TriHands) and monocular estimation \cite{zhang2025hawor} (\textbf{H3}). To further isolate the effect of hand quality, we add calibrated Gaussian noise (fit to the TriHands--HaWoR error distribution) to our triangulated hands at $0.5\times$ and $1.0\times$ the fitted standard deviation (Table~\ref{table:noise_ablation}).

\begin{figure}[t]
    \centering
    \includegraphics[width=\textwidth]{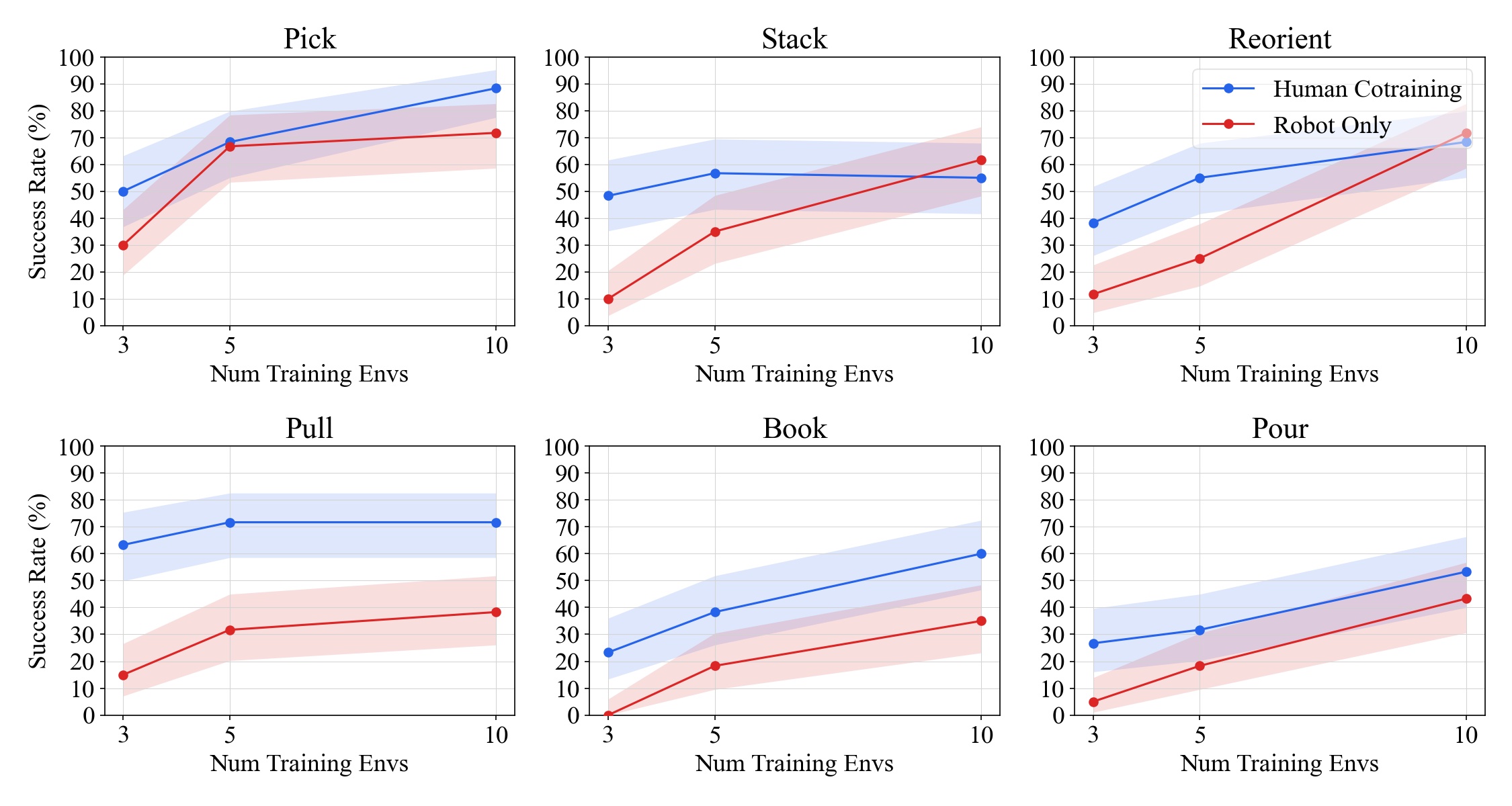}
    \caption{Per-task comparison between human cotraining with triangulated hands and robot-only training (95\% Clopper-Pearson CI).}
    \label{fig:human_vs_robot_scaling}
\end{figure}

\textbf{Architecture ablations:} To test \textbf{H4}, we ablate our token-level fusion and embodiment-specific encoders/decoders (Sec. \ref{sec:cross_embodiment_arch}) against two alternatives: (1) an image-bottleneck (\textit{CLS-token}) that pools ViT tokens to a single vector, as is common in prior work \cite{kareer2024egomimic, punamiya2025egobridge, chi2023diffusion}, and (2) shared action encoder/decoder weights across embodiments.

\textbf{Experiments:} We refer to training on both robot data and human data with our triangulated labels and recipe as \textit{Human Cotraining} (HC), and training on robot data alone as \textit{Robot Only} (RO). To test \textbf{H1}, we ablate the effect of human data at increasing robot data levels (Fig. \ref{fig:human_vs_robot_scaling}). For every task, we train on $\{3, 5, 10\}$ environments (50 demos each), and evaluate in 4 test environments (15 rollouts each). To test \textbf{H2}, we analyze 90 rollouts each (across all 6 tasks) for HC and RO, for environments where the HC has significantly higher mean success rate than RC. We categorize failures as \emph{global} (robot does not servo near the object) or \emph{local} (robot fails task by a few cm). If HC often resolves RO's local errors, that would indicate fine-grained motion transfer instead of coarse visual transfer.

\begin{table}[t]
\caption{Our cotraining recipe vs. others. Parentheses compare to Ours. All methods except Robot Only cotrained with TriHands, 3 envs of robot data. 95\% Gaussian CI.}
\label{table:our_cotraining_ablations}
\centering
\footnotesize
\setlength{\tabcolsep}{3pt}
\begin{tabular}{@{}lccccccc@{}}
\toprule
& \textbf{Mean} & \textbf{Pick} & \textbf{Stack} & \textbf{Pull} & \textbf{Reorient} & \textbf{Book} & \textbf{Pour} \\
\midrule
Ours & \textbf{41.5$\pm$12.1\%} & 50.0\% & \textbf{48.0\%} & \textbf{63.0\%} & \textbf{38.0\%} & \textbf{23.0\%} & \textbf{26.7\%} \\
\midrule
\multicolumn{8}{l}{\textbf{Ablations}} \\
\cmidrule(lr){1-8}
Robot only (vs.\ HC) & 12.0$\pm$8.2\% & 30.0\% & 10.0\% & 15.0\% & 11.7\% & 0.00\% & 5.00\% \\
CLS-token (vs.\ token fusion) & 14.7$\pm$18.6\% & 0.00\% & 6.67\% & 61.7\% & 5.00\% & 6.67\% & 8.33\% \\
\cmidrule(lr){1-8}
\multicolumn{8}{l}{\textbf{Datasets}} \\
\cmidrule(lr){1-8}
EgoDex \cite{hoque2025egodex} (vs.\ TriHands) & 19.4$\pm$13.4\% & 33.3\% & 45.0\% & 18.3\% & 8.33\% & 11.7\% & 0.00\% \\
\cmidrule(lr){1-8}
\multicolumn{8}{l}{\textbf{Methods}} \\
\cmidrule(lr){1-8}
HaWoR \cite{zhang2025hawor} (vs.\ TriHands) & 24.7$\pm$16.9\% & \textbf{53.3\%} & 10.0\% & 48.3\% & 15.0\% & 20.0\% & 1.70\% \\
EgoBridge \cite{punamiya2025egobridge} (vs.\ BC) & 16.6$\pm$10.7\% & 28.0\% & 10.0\% & 35.0\% & 20.0\% & 0.00\% & 6.67\% \\
PiZero \cite{black2410pi0} (vs.\ sep.\ enc/dec) & 17.5$\pm$6.0\% & 21.7\% & 21.7\% & 23.3\% & 15.0\% & 20.0\% & 3.30\% \\
\bottomrule
\end{tabular}
\end{table}

\textbf{Baselines: } \textit{CLS-token:} Vision tokens are attention-pooled to a single vector (image-bottleneck). \textit{EgoDex} \cite{hoque2025egodex}\textit{:} 145 hours of lab-collected egocentric data with limited scene diversity (vs.\ our 28 hours of natural scenes). \textit{HaWoR} \cite{zhang2025hawor}\textit{:} Human videos labelled with a monocular hand estimator instead of triangulation. \textit{EgoBridge} \cite{punamiya2025egobridge}\textit{:} Representational alignment loss based on action similarity; requires CLS-token architecture due to its optimal transport formulation. \textit{PiZero} \cite{black2410pi0}\textit{:} Shared action encoders and decoders across embodiments.

\section{Results}
\label{sec:results}

\begin{wraptable}{r}{0.52\textwidth}
\vspace{-1.5em}
\caption{Success rates for HC and RO (95\% Gaussian CI).}
\label{table:results_cotraining_means}
\centering
\footnotesize
\setlength{\tabcolsep}{3pt}
\begin{tabular}{@{}lccc@{}}
\toprule
\textbf{Method} & \textbf{3 envs} & \textbf{5 envs} & \textbf{10 envs} \\
\midrule
HC
& \textbf{41.7$\pm$12.1\%}
& \textbf{53.6$\pm$12.7\%}
& \textbf{66.1$\pm$10.5\%} \\
RO
& 12.0$\pm$8.2\%
& 32.5$\pm$14.5\%
& 53.6$\pm$13.4\% \\
\bottomrule
\end{tabular}
\vspace{-1em}
\end{wraptable}

\begin{wrapfigure}[19]{r}{0.51\textwidth}
    \vspace{-1.5em}
    \centering
    \includegraphics[width=0.49\textwidth]{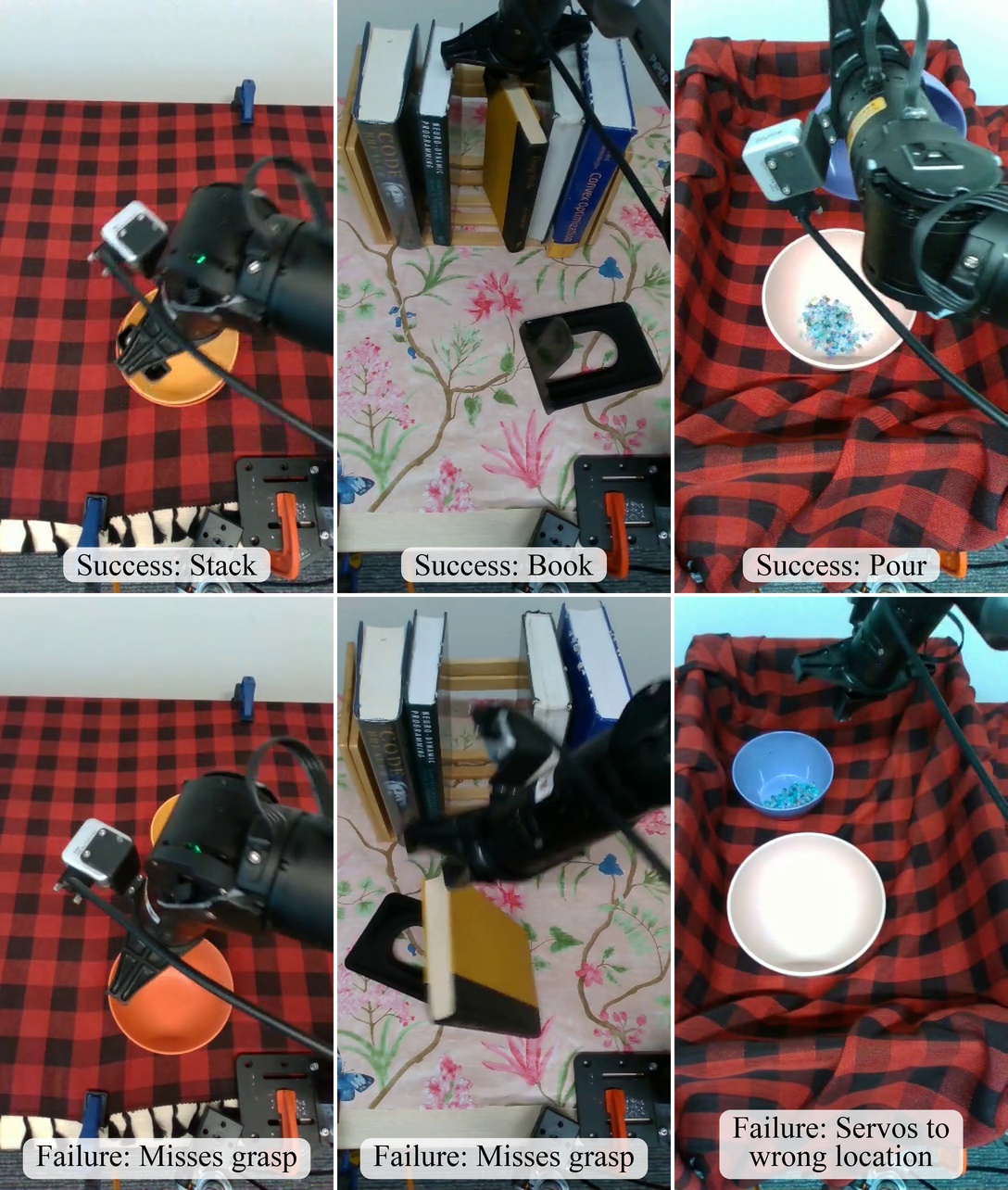}
    \caption{HC rollouts on unseen environments.}
    \label{fig:rollout_collage}
    \vspace{-1em}
\end{wrapfigure}

\textbf{Human cotraining significantly improves task performance over robot-only training.} From Table~\ref{table:results_cotraining_means}, we observe consistent gains across all robot data regimes using our recipe, with the largest improvements in the low-data setting: human cotraining improves absolute success rates by $20\%$--$48\%$ at the 3-env level (Figure~\ref{fig:human_vs_robot_scaling}). The Pull task benefits particularly strongly from our human data and preprocessing, even across different architectures and methods (Table~\ref{table:our_cotraining_ablations}). In Figure~\ref{fig:rollout_collage}, we provide a qualitative analysis of rollouts for the Human Cotraining model.

At higher robot data scales, the gap narrows---expected from prior work on cross-task cotraining \cite{barreiros2025careful}, since the robot-only baseline trains on same-task, same-distribution demos. Despite these diminishing returns, improving zero-shot transfer in the low-data regime remains practically useful when paired with RL post-training \cite{intelligence2025pi}. Finally, TriHands outperforms EgoDex across all tasks despite $5\times$ fewer hours of data, suggesting scene diversity matters more than scale (Table~\ref{table:our_cotraining_ablations}).


\textbf{There is motion transfer on simple tasks, but motion transfer is inconclusive for our most complex task.} We visually analyze the errors and find $100\%$ of the Robot Only errors on 5/6 tasks are local errors. On these tasks, Human Cotraining outperforms Robot Only $82.8\pm11.4\%$ to $29.2\pm13.7\%$, indicating motion transfer. For the remaining task (Pour), Robot Only struggles to even servo to the object, so we cannot use this argument. We believe the overall recipe may be less effective on complex tasks due to the reliance on kinematic retargeting and the larger motion gap, and leave further investigation to future work.

\begin{wraptable}{r}{0.45\textwidth}
\vspace{-1.5em}
\caption{HaWoR hand pose error metrics (mm) using triangulated 3D hands as ground-truth.}
\label{table:handpose_metrics}
\centering
\footnotesize
\setlength{\tabcolsep}{3pt}
\begin{tabular}{@{}cccc@{}}
\toprule
\textbf{MPJPE} & \textbf{PA-MPJPE} & \textbf{W-MPJPE} & \textbf{WA-MPJPE} \\
\midrule
185.67 & 15.72 & 161.85 & 77.86 \\
\bottomrule
\end{tabular}
\vspace{0.5em}
\caption{Effect of hand pose noise on transfer. 95\% Gaussian CI.}
\label{table:noise_ablation}
\centering
\footnotesize
\setlength{\tabcolsep}{3pt}
\begin{tabular}{@{}lccc@{}}
\toprule
\textbf{Noise level} & \textbf{Mean} & \textbf{Reorient} & \textbf{Pick bowl} \\
\midrule
0.0 & $\mathbf{48.3\pm19.6\%}$ & $\mathbf{38.3\%}$ & $\mathbf{58.3\%}$ \\
0.5 & $30.0\pm13.1\%$ & $23.3\%$ & $36.7\%$ \\
1.0 & $20.0\pm6.5\%$ & $16.7\%$ & $23.3\%$ \\
\bottomrule
\end{tabular}
\vspace{0.5em}
\caption{Pinhole camera ablation (95\% Clopper-Pearson CI). Supp.~\ref{supp:camera_ablation_details}.}
\label{table:camera_ablation}
\centering
\footnotesize
\setlength{\tabcolsep}{3pt}
\begin{tabular}{@{}lc@{}}
\toprule
\textbf{Human Camera Model} & \textbf{Mean SR} \\
\midrule
Extent-Matched Pinhole (Ours) & \textbf{51.2\% [39.8, 62.6]} \\
Medium Misalignment Pinhole & 40.0\% [29.2, 51.6] \\
Heavy Misalignment Pinhole & 25.0\% [16.0, 35.9] \\
Robot-Only Data & 26.2\% [17.0, 37.3] \\
\bottomrule
\end{tabular}
\vspace{-1em}
\end{wraptable}

\textbf{Hand quality matters, but the current state-of-the-art monocular hand estimator still shows transfer.} Monocular-estimated hands performance is worse when averaged across tasks compared to the triangulated hands (Table \ref{table:our_cotraining_ablations}). However, the mean success rate for HaWoR is still higher than  Robot Only ($24.7\%$ vs $11.9\%$) (Table \ref{table:results_cotraining_means}). This indicates even noisy hand labels provide some benefit across certain tasks but greater improvements in hand pose estimation quality will unlock better transfer across other tasks. Given the metrics in Table \ref{table:handpose_metrics}, it is unsurprising the monocular transfer is worse --- the hand joint translations in our 1-second prediction horizon can often be dominated by the noise in the depth values. Table~\ref{table:noise_ablation} shows that transfer degrades monotonically with noise level, further confirming that higher quality hands lead to greater transfer.

\textbf{Embodiment-specialization significantly increases human-to-robot transfer.} Prior works on human data with more robot-aligned motion use image-bottleneck architectures \cite{kareer2024egomimic, punamiya2025egobridge, chi2023diffusion} and shared action decoders across embodiments \cite{black2410pi0}. We find that our token-fusion architecture outperforms the CLS-token bottleneck on all tasks except Pull (Table \ref{table:our_cotraining_ablations}), likely because the large motion gap makes human and robot action chunks distinguishably different, and token-level fusion allows the model to specialize visual understanding per embodiment. Additionally, untying the action encoders and decoders leads to large improvements in 5 out of 6 tasks, for a similar reason.

\textbf{Other ablation experiments}
Scale-aligning the human fisheye images with an extent-matched pinhole camera doubles performance over Robot Only (Table \ref{table:camera_ablation}); a heavily-misaligned pinhole yields no transfer benefit. EgoBridge \cite{punamiya2025egobridge} improves over the CLS-token baseline on 4/6 tasks but lags behind our BC-only recipe (Table \ref{table:our_cotraining_ablations}), likely because the larger action gap in our data makes its optimal transport loss based on action similarity unreliable.

\section{Conclusion}
We present a dataset of everyday human videos with accurate triangulated hand labels, and present a framework for successfully training on such videos with natural motions. Extensive experiments show improved zero-shot generalization to novel objects and backgrounds when cotraining with our human videos. Our hand quality experiments show that modern hand-pose estimators are starting to show transfer on natural Internet-type videos, and further improvements in hand pose estimation should unlock greater transfer. For future work, we seek to find new representational alignment and action retargeting techniques that work with the complexity of everyday videos. Overall, we believe our work analyzes difficulties and suggests solutions for Internet video cotraining.

\clearpage
\acknowledgments{Richard Li was supported by the NSF Institute for Artificial Intelligence and Fundamental Interactions (Grant No. PHY-2019786) and the Felicis Scholars program. Aditya Prakash and Saurabh Gupta were supported by NSF Grant IIS-2007035. We thank John Marangola for his advice on the robot setup, and Antonia Bronars and Branden Romero for paper writing suggestions.}

\bibliography{references}

\clearpage
\appendix

\section{Hand label comparison}
\label{supp:keypoints_comparison}

Ego-Exo4D provides 2D keypoint labels but does not release its fine-tuned MMPose model. Therefore, to demonstrate that our 3D-projected keypoints (\textbf{Ours}) are superior to their 2D keypoints (\textbf{Ego-Exo4D MMPose}), we conduct a human pairwise evaluation with 1700 trials and 17 subjects. We allow ties and exclude them from calculations. $P(\textbf{Ours}\ \text{wins}) = 76.3\% \; (95\%\,\text{CI }[74.0\%,\,78.6\%])$.
\begin{figure}[h]
    \centering
    \includegraphics[width=1.0\linewidth]{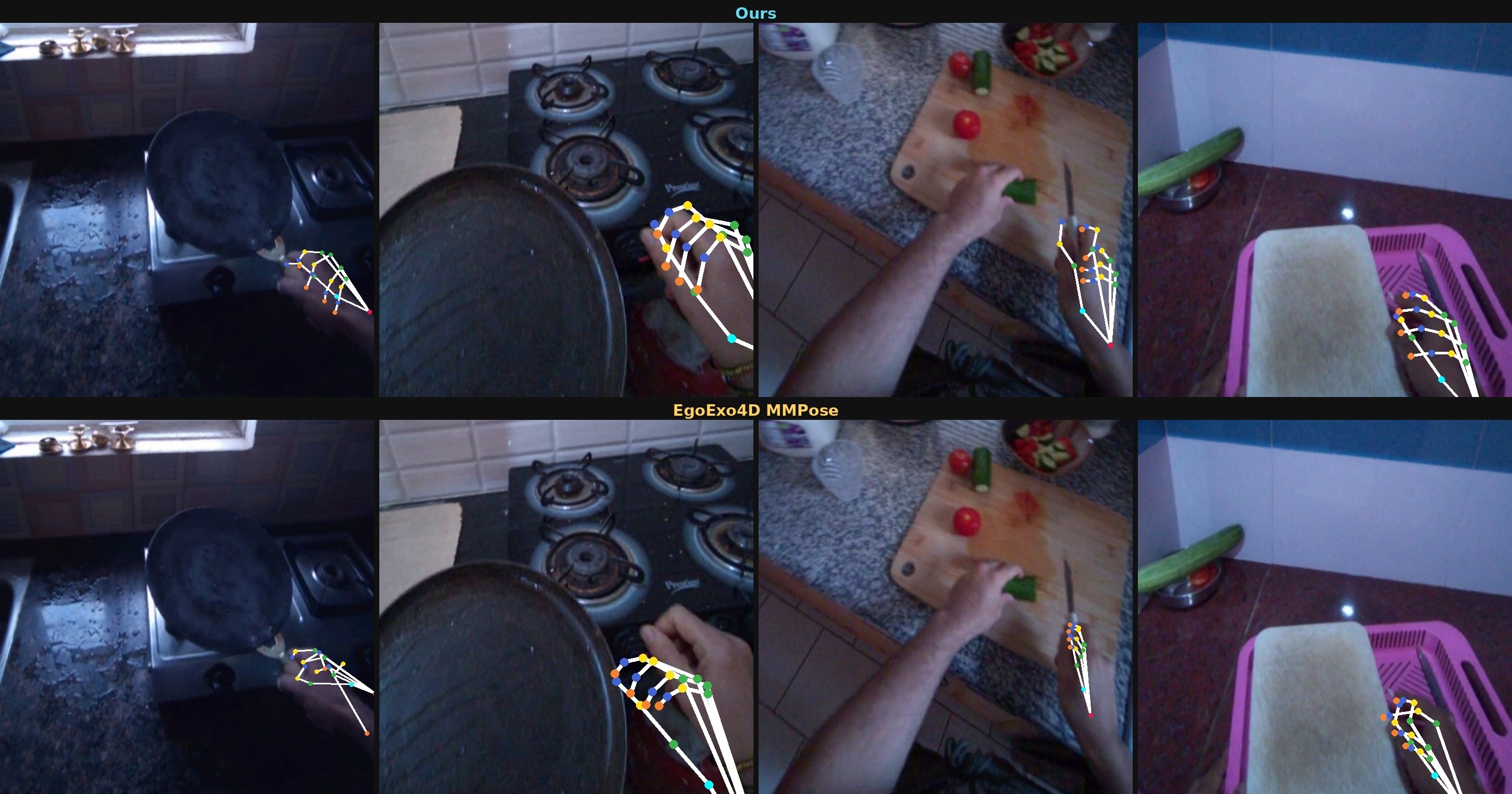}
    \caption{Top-row are ego-view projected keypoints from our 3D hands, and bottom row are corresponding projected keypoints from the default EgoExo4D 3D hand keypoints.}
    \label{fig:keypoints_comparison}
\end{figure}

\section{TriHands triangulation pipeline}
\label{supp:trihands_triangulation_pipeline}

Each episode in EgoExo4D comes with one egocentric view from a head-mounted camera and $M \in \{ 4, 5 \}$ exocentric views. For a given set of 2D hand-keypoints associated with a set of cameras, we use the direct-linear transformation (DLT) algorithm \cite{hartley2003multiple} as an efficient initial triangulation estimate then refine triangulated points with reprojection error above a threshold using nonlinear optimization.

To select the most accurate triangulation among all camera subsets, we incorporate several safeguards. Minimizing reprojection error alone is insufficient, as a set of cameras can exhibit low reprojection error yet yield poor 3D reconstructions due to small 2D keypoint errors. Instead, we maximize the number of cameras used while enforcing reprojection error thresholds. In addition, EgoExo4D videos often contain secondary individuals near the ego-camera wearer. To avoid their hands, we require the selected camera set to include valid hand detections for the egocentric view.

For a subset $S \subset C$ of the total set of cameras $C$, we set the reprojection threshold that must be satisfied for each camera $c$ to be $0.01 \cdot \max(H_c, W_c)$, where $H_c$ and $W_c$ are the camera image dimensions in pixels. To simplify experiments, we only use right-hand human hand labels to train a single robot arm. After applying up to 0.4s linear interpolation between 3D joints of triangulated hands, we end up with 3,042,406 frames of reconstructed right hands at 30fps, which corresponds to over 28 hours of usable data for training.

\begin{algorithm}[h]
\caption{Multiview Hand Triangulation Pipeline}
\label{alg:triangulation}
\begin{algorithmic}[1]
\Require Multi-view video $\mathcal{V} = \{V_{\text{ego}}, V_{\text{exo}_1}, \ldots, V_{\text{exo}_N}\}$, intrinsics $\{\mathbf{K}_c\}$, extrinsics $\{\mathbf{T}_c\}$ for each camera $c$
\Ensure 3D keypoints $\mathbf{J}_{3D}$, MANO parameters $\boldsymbol{\theta}$
\Statex where $c$ denotes both a camera and its associated image, and
\Statex $\rho_{\text{H}}(r) = \begin{cases} \frac{1}{2}r^2 & |r| \leq \delta \\ \delta|r| - \frac{1}{2}\delta^2 & |r| > \delta \end{cases}$ (Huber loss)

\Statex \textit{// Stage 1: Hand Detection}
\ForAll{frame $f$, camera $c$}
    \State $\mathbf{B}_c^f \gets \textsc{HandDetector}(V_c[f])$ \Comment{bounding boxes}
\EndFor

\Statex \textit{// Stage 2: Single-View 3D Estimation}
\ForAll{frame $f$, camera $c$, hand $h \in \{L, R\}$}
    \State $I_{\text{patch}} \gets \textsc{Crop}(V_c[f], \mathbf{B}_c^f[h])$
    \State $\mathbf{J}_{2D}, \boldsymbol{\theta}_{\text{init}} \gets \textsc{WiLoR}(I_{\text{patch}})$ \Comment{2D joints \& MANO}
\EndFor

\Statex \textit{// Stage 3: Multiview Triangulation}
\ForAll{frame $f$, hand $h$}
    \State $\mathcal{C} \gets \{c : \mathbf{B}_c^f[h] \text{ exists}\}$
    \If{$|\mathcal{C}| < 2$} \textbf{continue} \EndIf
    \ForAll{$c \in \mathcal{C}$}
        \State $\mathbf{P}_c \gets \mathbf{K}_c [\mathbf{I} | \mathbf{0}] \mathbf{T}_c^{-1}$
    \EndFor
    \ForAll{subset $\mathcal{S} \subseteq \mathcal{C}$ with $c_{\text{ego}} \in \mathcal{S}$, $|\mathcal{S}| \geq 2$}
        \ForAll{joint $j = 1, \ldots, 21$}
            \State $\mathbf{X}_j \gets \textsc{DLT}(\{\mathbf{P}_c, \mathbf{J}_{2D}[c,j]\}_{c \in \mathcal{S}})$
            \State $\mathbf{X}_j \gets \argmin_{\mathbf{X}} \sum_{c} \rho_{\text{H}}\bigl(\|\pi_c(\mathbf{X}) - \mathbf{J}_{2D}[c,j]\|\bigr)$
            \State $\epsilon_j[c] \gets \|\pi_c(\mathbf{X}_j) - \mathbf{J}_{2D}[c,j]\|$
        \EndFor
    \EndFor
    \State $\mathcal{S}^* \gets \argmax_{|\mathcal{S}|}$ s.t.\ $\geq$95\% joints: $\epsilon_j[c] < \tau_c$
    \State $\mathbf{J}_{3D}[h,f] \gets \mathbf{X}[\mathcal{S}^*]$
\EndFor

\Statex \textit{// Stage 4: Temporal Interpolation}
\State $\mathbf{J}_{3D} \gets \textsc{Slerp}(\mathbf{J}_{3D}, g_{\max}=12)$ \Comment{max gap: 12 frames}

\Statex \textit{// Stage 5: Inverse Kinematics}
\ForAll{frame $f$, hand $h$}
    \State $\boldsymbol{\theta}_0 \gets \boldsymbol{\theta}_{\text{init}}[c_{\text{ego}}, h, f]$ \Comment{HaWoR init}
    \State $\mathbf{J}_{\text{tgt}} \gets \mathbf{T}_{\text{cam}}^{-1} \mathbf{J}_{3D}[h,f]$
    \State $\boldsymbol{\theta}[h,f] \gets \argmin_{\boldsymbol{\theta}} \|\mathcal{M}(\boldsymbol{\theta}) - \mathbf{J}_{\text{tgt}}\|^2$
\EndFor
\State \Return $\mathbf{J}_{3D}, \boldsymbol{\theta}$
\end{algorithmic}
\end{algorithm}

Some additional factors we found to be important for our triangulation pipeline:
\begin{enumerate}
    \item Most modern 3D hand pose estimators begin reconstruction from a cropped image of just the hand(s). We found the default Ultralytics YOLO v11 bounding box estimator included with our 3D hand model WiLoR \cite{potamias2025wilor} to have an extremely high rate of false negatives (missing bounding box detections) as well as chirality errors. We addressed this by swapping the Ultralytics model for a version of Hands23 \cite{cheng2023towards} finetuned on EpicKitchens.
    \item Due to our swapping of the bounding box estimator, and due to different padding conventions for Ultralytics YOLO and Hands23, the cropped images going into WiLoR became out-of-distribution, and 3D hand reconstructions were of poor quality. We addressed this by training a small bounding box translation model that took the chirality, bounding box extents, and bounding box location of a Hands23 bounding box and outputted corresponding Ultralytics YOLO bounding box extents. We supervised this translation model by running Ultralytics YOLO and Hands23 on a subset of the videos and finding the intersection set of frames where both models detected hands.
    \item Our triangulation pipeline produces triangulated 3D joints. To convert this into MANO parameters, we run GPU-batched inverse kinematics through the MANO model. Due to nonconvexity, this optimization often fails from random initialization - we initialize the optimization by using MANO parameter prediction from HaWoR \citep{zhang2025hawor}, which has relatively accurately metric space parameters compared to WiLoR \citep{potamias2025wilor}.
\end{enumerate}

\section{Triangulation visualizations}

\begin{figure}[t]
    \centering
    \includegraphics[
        width=0.98\textwidth,
        keepaspectratio
    ]{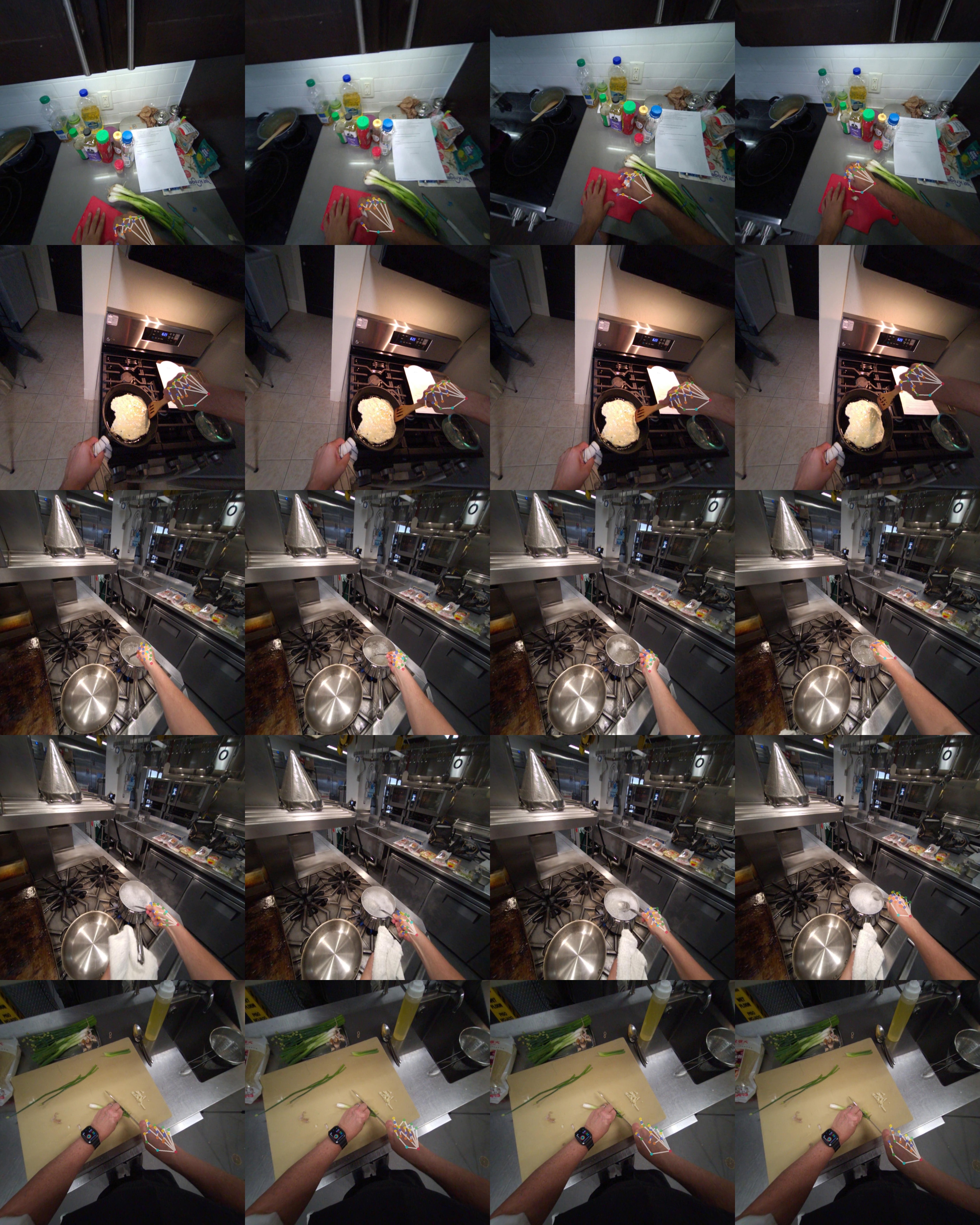}
    \caption{TriHands triangulation visualizations (first five scenes).}
    \label{fig:triangulation_collage_first}
\end{figure}

\begin{figure}[t]
    \centering
    \includegraphics[
        width=0.98\textwidth,
        keepaspectratio
    ]{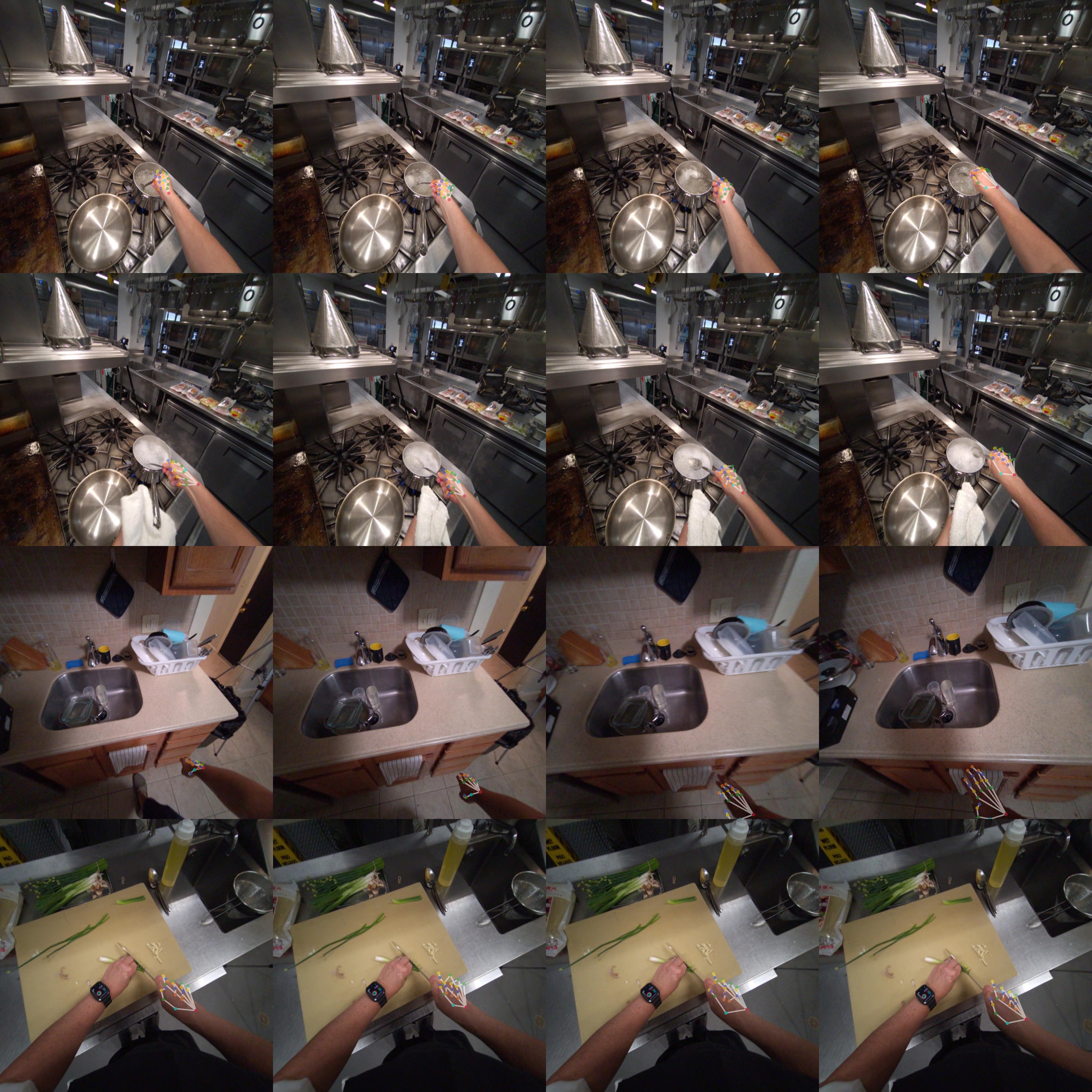}
    \caption{TriHands triangulation visualizations (remaining four scenes).}
    \label{fig:triangulation_collage_last}
\end{figure}

We visualize 2D projections in the egocamera of our triangulations in Fig. \ref{fig:triangulation_collage_first}, \ref{fig:triangulation_collage_last}.

\section{Camera specifications}
\label{supp:camera_specs}
The human camera is an RGB fisheye camera with $110^{\circ}$ HFOV and $110^{\circ}$ VFOV, while the robot camera is an RGB RealSense D435i camera with $87^{\circ}$ HFOV and $58^{\circ}$ VFOV.

\section{Camera ablation details}
\label{supp:camera_ablation_details}
We compare results cotraining on the pick spatula task, with the following intrinsics $(h, w, f_x, f_y)$ per baseline: extent-matched $(1280, 720, 690, 627)$, medium $(1280, 720, 460, 460)$, heavy $(720, 720, 211, 211)$. In addition, for the heavy misalignment model, we do not crop to fix the aspect ratio differences between human and robot camera.

\section{Image-space alignment extrinsic derivation}
\label{supp:pix_extrinsic}
Consider a human camera with focal length $f_c$ and vertical resolution $H_c$, and a robot camera with parameters $f_s$ and $H_s$. Suppose the relevant portion of the human scene lies at depth $z_{\text{human}}$ with respect to the camera frame (OpenCV camera frame convention). The top and bottom image rows, $v^c_T$ and $v^c_B$, correspond to 3D heights:
\[
Y^c_T = \frac{z_{\text{human}}\, v^c_T}{f_c}, \qquad
Y^c_B = \frac{z_{\text{human}}\, v^c_B}{f_c},
\]
so the vertical 3D span of the visible region is
\[
L^c = Y^c_T - Y^c_B
    = \frac{z_{\text{human}}\, H_c}{f_c}.
\]

We seek a robot-camera depth $z_{\text{robot}}$ such that its captured region has the same vertical 3D span. Because its pixel span is $H_s = v^s_T - v^s_B$,
\[
\frac{z_{\text{robot}}\, H_s}{f_s} = L^c.
\]
Solving gives
\[
z_{\text{robot}}
= z_{\text{human}}\, \frac{H_c}{f_c}\, \frac{f_s}{H_s}.
\]

This depth translation aligns the robot camera so that both cameras observe an equivalent 3D vertical extent.

\section{Action marginal distributions visualization}
\label{supp:marginal_dist}
\begin{figure}[p]
    \centering

    \begin{subfigure}{0.6\textwidth}
        \centering
        \includegraphics[width=\linewidth]{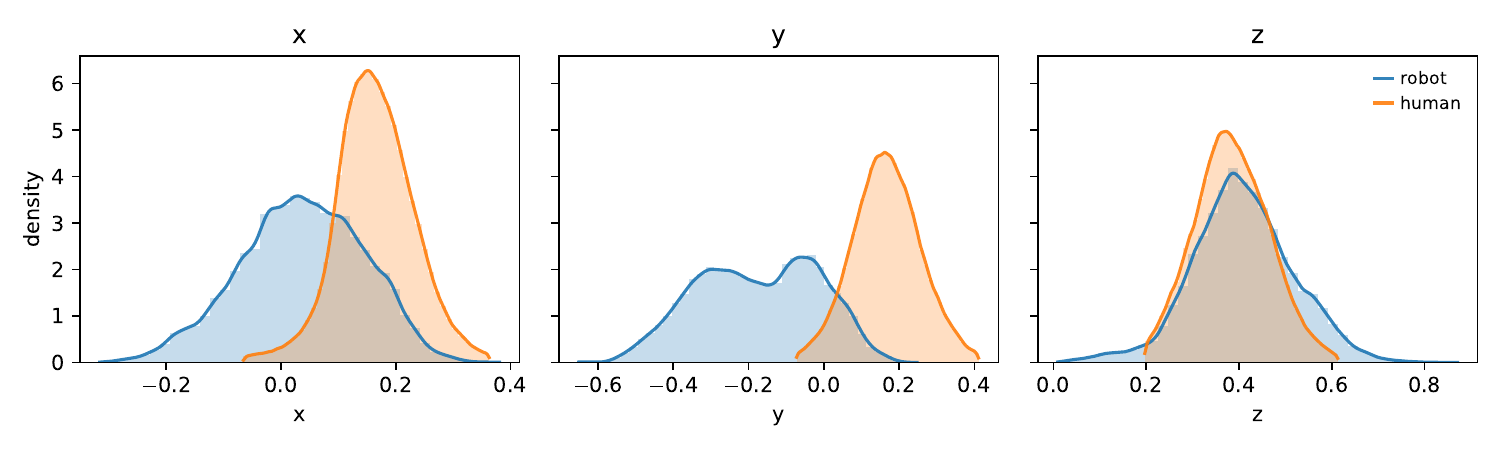}
        \caption{Pick}
        \label{fig:action-marginals-pick}
    \end{subfigure}

    \begin{subfigure}{0.6\textwidth}
        \centering
        \includegraphics[width=\linewidth]{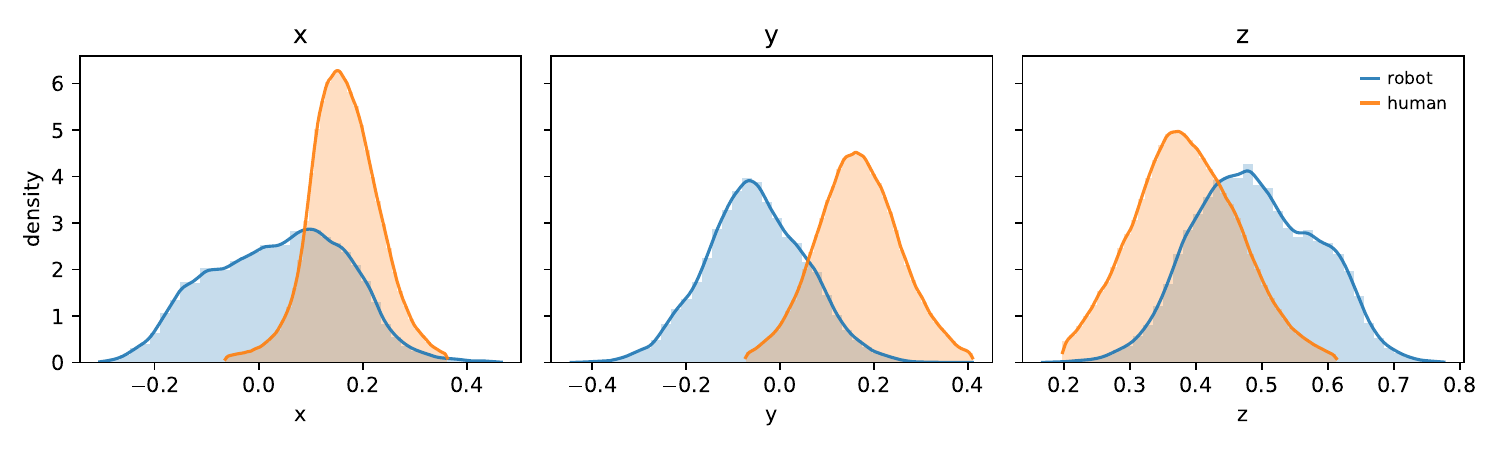}
        \caption{Stack}
        \label{fig:action-marginals-stack}
    \end{subfigure}

    \begin{subfigure}{0.6\textwidth}
        \centering
        \includegraphics[width=\linewidth]{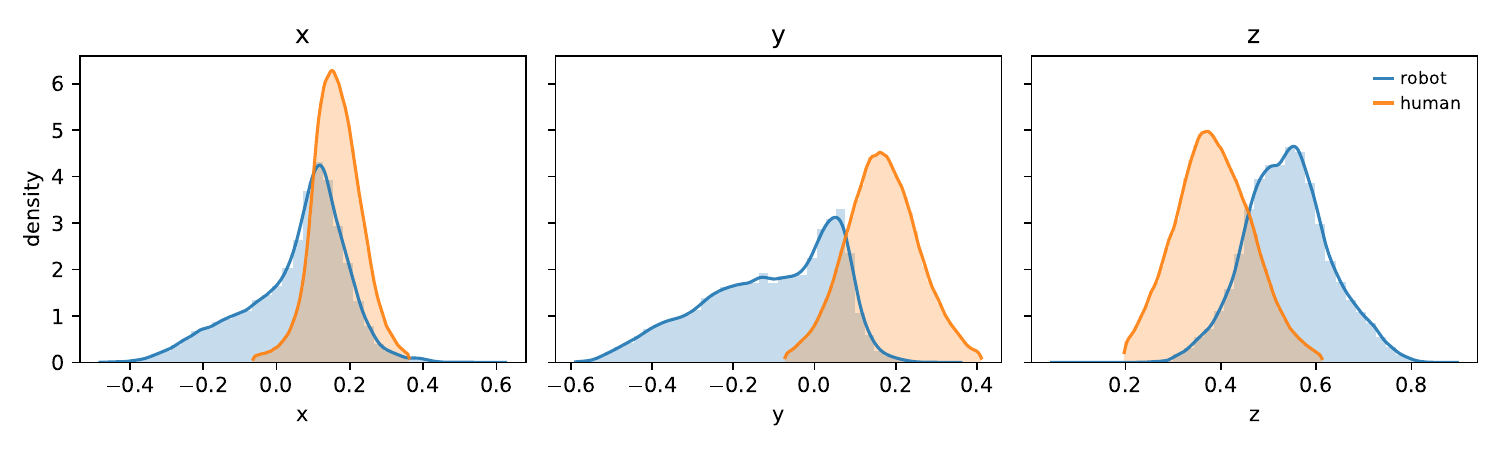}
        \caption{Pull}
        \label{fig:action-marginals-pull}
    \end{subfigure}

    \begin{subfigure}{0.6\textwidth}
        \centering
        \includegraphics[width=\linewidth]{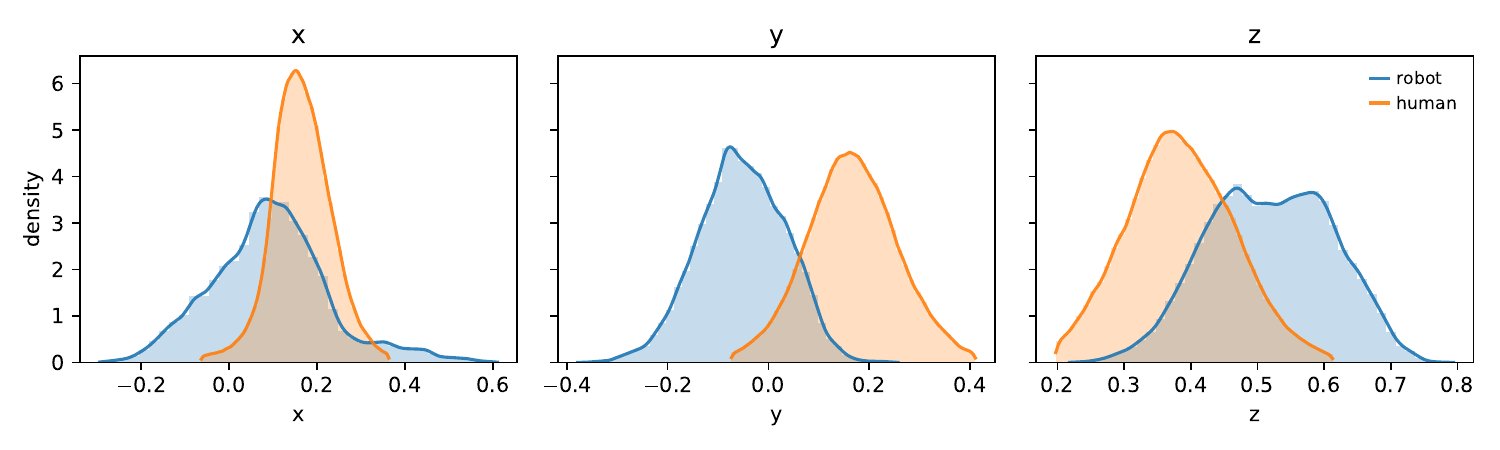}
        \caption{Reorient}
        \label{fig:action-marginals-reorient}
    \end{subfigure}

    \begin{subfigure}{0.6\textwidth}
        \centering
        \includegraphics[width=\linewidth]{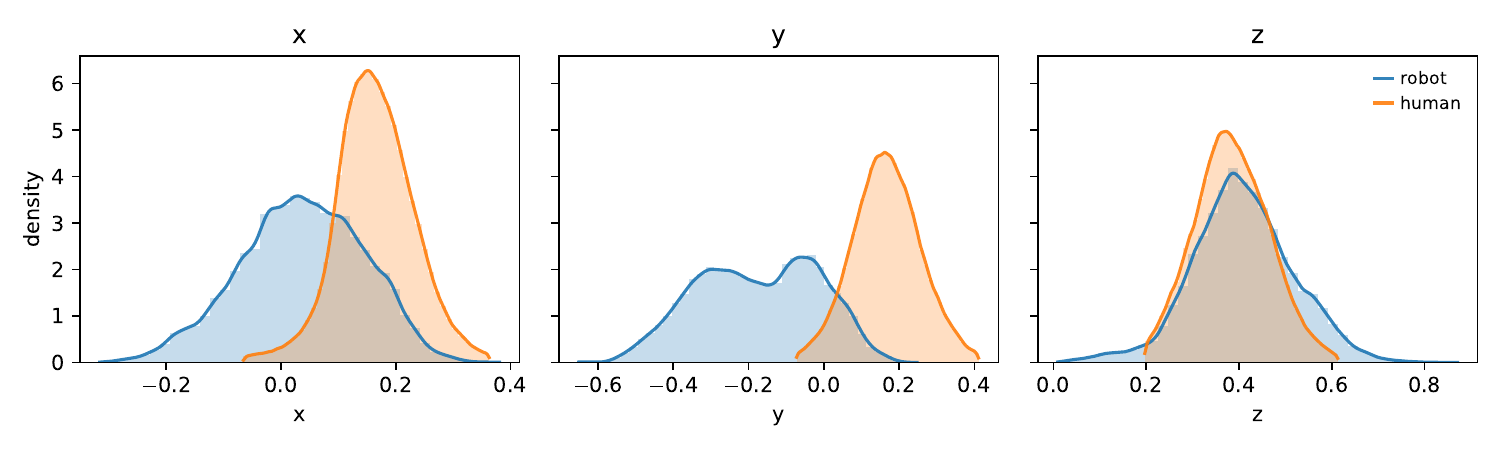}
        \caption{Book}
        \label{fig:action-marginals-book}
    \end{subfigure}

    \begin{subfigure}{0.6\textwidth}
        \centering
        \includegraphics[width=\linewidth]{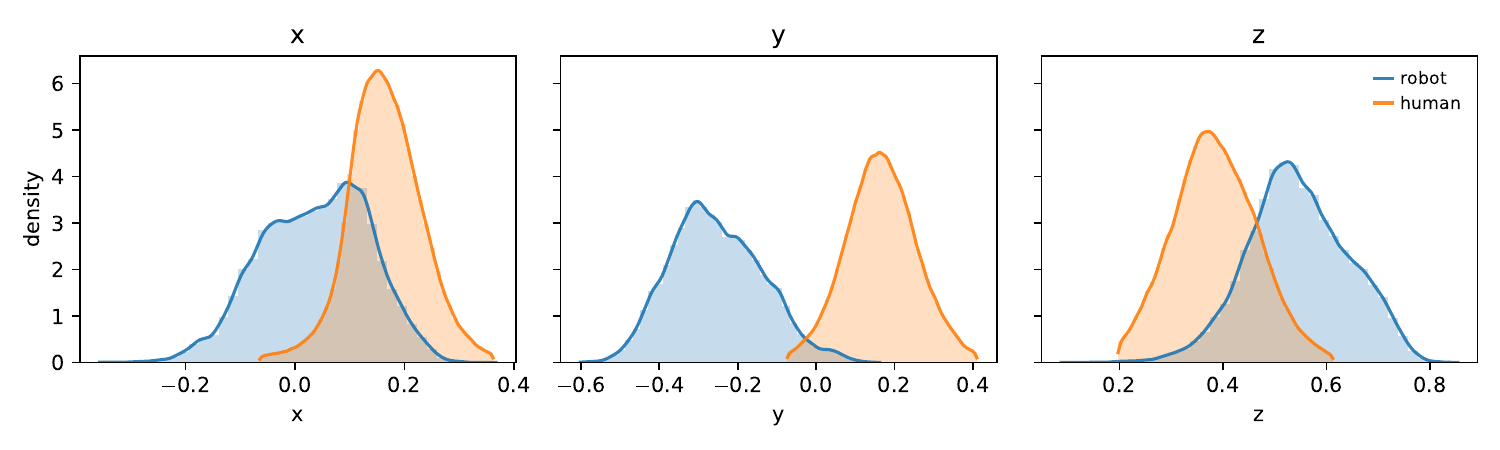}
        \caption{Pour}
        \label{fig:action-marginals-pour}
    \end{subfigure}

    \caption{Comparison of robot and human action marginals in XYZ camera coordinates across all six tasks.}
    \label{fig:action-marginals-all}
\end{figure}

We visualize the distribution of human hand (MANO TCP) translations and robot TCP frame translations, before our mean-centering and unit-scaling normalization, in Fig. \ref{fig:action-marginals-all}.

\begin{algorithm}[t]
\caption{Cotraining conditional flow-matching policy}
\label{alg:cotraining}
\begin{algorithmic}[1]
\Require Human dataset $\mathcal{D}_H$, robot dataset $\mathcal{D}_R$
\Require Flow-matching policy $v_\theta$, target field $u(\cdot \mid A_t)$
\Require Action noising distribution $q(\cdot \mid A_t)$
\Require Timestep distribution $\mathrm{Beta}(\alpha,\beta)$
\Require Batch sizes $B_H, B_R$
\Require Learning rates $\eta_{\text{vis}}$ (SigLIP encoder) and $\eta_{\text{rest}}$ (remaining modules)

\State Initialize $v_\theta$: vision encoder initialized with SigLIP, remaining modules from scratch
\State Initialize Adam optimizer with parameter groups:\\
\hspace{1.5em}$\mathrm{Adam}(\{(\theta_{\text{vis}},\eta_{\text{vis}}),\, (\theta_{\text{rest}},\eta_{\text{rest}})\})$

\While{stopping criterion not met}

    \State \textbf{// Human batch}
    \State Sample mini-batch $\{(o_t^{H,i}, A_t^{H,i})\}_{i=1}^{B_H} \sim \mathcal{D}_H$
    \State Sample timesteps $\{\tau^{H,i}\}_{i=1}^{B_H} \sim \mathrm{Beta}(\alpha,\beta)$
    \State Sample noisy actions $\{A_{t}^{\tau,H,i}\}_{i=1}^{B_H} \sim q(\cdot \mid A_t^{H,i})$
    \State Compute targets:\\
           \hspace{1.5em}$u^{H,i} \gets u(A_{t}^{\tau,H,i} \mid A_t^{H,i})$
    \State $\mathcal{L}_H \gets \frac{1}{B_H} \sum_{i=1}^{B_H}
        \| v_\theta(A_{t}^{\tau,H,i},\, o_t^{H,i}) - u^{H,i} \|_2^2$

    \State

    \State \textbf{// Robot batch}
    \State Sample mini-batch $\{(o_t^{R,i}, A_t^{R,i})\}_{i=1}^{B_R} \sim \mathcal{D}_R$
    \State Sample timesteps $\{\tau^{R,i}\}_{i=1}^{B_R} \sim \mathrm{Beta}(\alpha,\beta)$
    \State Sample noisy actions $\{A_{t}^{\tau,R,i}\}_{i=1}^{B_R} \sim q(\cdot \mid A_t^{R,i})$
    \State Compute targets:\\
           \hspace{1.5em}$u^{R,i} \gets u(A_{t}^{\tau,R,i} \mid A_t^{R,i})$
    \State $\mathcal{L}_R \gets \frac{1}{B_R} \sum_{i=1}^{B_R}
        \| v_\theta(A_{t}^{\tau,R,i},\, o_t^{R,i}) - u^{R,i} \|_2^2$

    \State

    \State \textbf{// Cotraining update}
    \State $\mathcal{L} \gets \mathcal{L}_H + \mathcal{L}_R$
    \State \text{Adam.update}$(\nabla_\theta \mathcal{L})$

\EndWhile
\State \Return $\theta$
\end{algorithmic}
\end{algorithm}

\section{Cotraining algorithm and loss function}
\label{supp:cotraining_alg}
The conditional flow-matching imitation learning loss for a single dataset $\mathcal{D}_i$, with observations $o_t$ and action chunks $A_t$, is:
\[
\mathcal{L}_{\mathcal{D}_i}(\theta)
=
\mathbb{E}_{\substack{
(o_t,A_t)\sim\mathcal{D}_i\\
\tau \sim \mathrm{Beta}(\alpha,\beta) \\
A_t^\tau\sim q(\cdot\mid A_t)
}}
\left[
\bigl\|
v_\theta(A_t^\tau,\, o_t)
-
u(A_t^\tau \mid A_t)
\bigr\|_2^{\,2}
\right]
\]

Uniformly sampling from the concatenated human and robot datasets is equivalent to sampling from a mixture-of-empirical-distributions:
\[
\begin{aligned}
D_R \cup D_H
&= \frac{1}{n_R + n_H}
   \left(
      \sum_{i=1}^{n_R} \delta(x - x_i)
      +
      \sum_{j=1}^{n_H} \delta(x - x_j)
   \right) \\[0.75em]
&= \frac{n_R}{n_R+n_H} (
   \frac{1}{n_R} \sum_{i=1}^{n_R} \delta(x - x_i))
   \\[-0.2em]
&\quad
   +\;
   \frac{n_H}{n_R+n_H} (
   \frac{1}{n_H} \sum_{j=1}^{n_H} \delta(x - x_j)) \\[0.75em]
&= \frac{n_R}{n_R+n_H}\, D_R(x)
 \;+\;
   \frac{n_H}{n_R+n_H}\, D_H(x).
\end{aligned}
\]

The overall loss would then be:
\[
\mathcal{L}_{\mathcal{D}_H \bigcup \mathcal{D}_R}(\theta) = \frac{n_R}{n_R + n_H} \mathcal{L}_{D_R} + \frac{n_H}{n_R+n_H} \mathcal{L}_{D_H}
\]

By independently sampling uniform batches from each dataset and summing the losses, we optimize instead: \[
\mathcal{L}_{\text{weighted}}(\theta) = \mathcal{L}_{D_R} + \mathcal{L}_{D_H}
\]

The full pseudo-code for our cotraining algorithm is in Alg. \ref{alg:cotraining}.

\begin{table}[t]
\caption{Model and training hyperparameters.}
\label{table:supp_architecture}
\centering
\begin{tabular}{l l}
\toprule
\textbf{Component} & \textbf{Configuration} \\
\midrule
\multicolumn{2}{l}{\textbf{Optimization}} \\
Visual Encoder LR & $3.0\times10^{-5}$ \\
MoE Parameters LR & $3.0\times10^{-4}$ \\
Adam Betas & $(0.9,\;0.95)$ \\
\midrule
\multicolumn{2}{l}{\textbf{Self-attention}} \\
Hidden Layers & 6 \\
Attention Heads & 8 \\
Key--Value Heads & 1 \\
Head Dim & 256 \\
\midrule
\multicolumn{2}{l}{\textbf{SigLIP Vision Encoder}} \\
Hidden Size & $1152$ \\
Intermediate Size & $4304$ \\
Hidden Layers & $27$ \\
Attention Heads & $16$ \\
Image Size & $224$ \\
Patch Size & $14$ \\
\midrule
\multicolumn{2}{l}{\textbf{EgoBridge Loss}} \\
Alignment Weight $\lambda$ & $1\times10^{-4}$ \\
DTW $\gamma$ & $0.1$ \\
OT Loss Weight & $1.0$ \\
\midrule
\multicolumn{2}{l}{\textbf{Vision Expert}} \\
Hidden Size & $2048$ \\
Intermediate Size & $4096$ \\
\midrule
\multicolumn{2}{l}{\textbf{Proprioception Expert}} \\
Hidden Size & $1024$ \\
Intermediate Size & $4096$ \\
\midrule
\multicolumn{2}{l}{\textbf{Action Expert}} \\
Hidden Size & $1024$ \\
Intermediate Size & $4096$ \\
\midrule
\multicolumn{2}{l}{\textbf{Flow Matching Parameters}} \\
Beta-distribution $\alpha$ & $1.5$ \\
Beta-distribution $\beta$ & $1.0$ \\
\midrule
\multicolumn{2}{l}{\textbf{Action/Proprio Encoder/Decoder MLPs}} \\
Num Layers & $3$ \\
Hidden Size & $1024$ \\
Activation & LeakyReLU \\
\bottomrule
\end{tabular}
\end{table}

\section{Architecture details}
\label{supp:transfusion_arch}

\noindent
Our architecture follows a Transfusion-inspired design that jointly models visual observations and continuous actions using a shared transformer backbone with modality-specific weights. The model operates on trajectories from two embodiments, human ($H$) and robot ($R$). RGB images are encoded using a shared SigLIP ViT encoder $E_{\mathrm{img}}$, producing visual tokens $\mathbf{z}_{\mathrm{img}}^d = E_{\mathrm{img}}(\mathbf{x}^d)\in\mathbb{R}^{N\times D}$ for each dataset $d\in\{H,R\}$. Actions are corrupted by a flow-matching process at timestep $t$, yielding $\tilde{\mathbf{a}}_t^d$, and embedded using dataset-specific action encoders $\mathbf{z}_{\mathrm{act}}^d = E_{\mathrm{act}}^{\,d}(\tilde{\mathbf{a}}_t^d)\in\mathbb{R}^{T\times D}$. Image and action tokens are concatenated into a single sequence $\mathbf{z}^d = [\mathbf{z}_{\mathrm{img}}^d;\mathbf{z}_{\mathrm{act}}^d]\in\mathbb{R}^{(N+T)\times D}$, which forms the input to the Transformer.

Within each Transformer layer, self-attention is computed jointly over image and action tokens using modality-specific projections. For each modality $m\in\{\mathrm{img},\mathrm{act}\}$, queries, keys, and values are obtained as $\mathbf{Q}_m^d = \mathbf{z}_m^d W_Q^{m}$, $\mathbf{K}_m^d = \mathbf{z}_m^d W_K^{m}$, and $\mathbf{V}_m^d = \mathbf{z}_m^d W_V^{m}$, where $W_{\{Q,K,V\}}^{m}$ depend only on modality and are shared across datasets. Concatenating these across modalities yields $\mathbf{Q}^d = [\mathbf{Q}_{\mathrm{img}}^d;\mathbf{Q}_{\mathrm{act}}^d]$, $\mathbf{K}^d = [\mathbf{K}_{\mathrm{img}}^d;\mathbf{K}_{\mathrm{act}}^d]$, and $\mathbf{V}^d = [\mathbf{V}_{\mathrm{img}}^d;\mathbf{V}_{\mathrm{act}}^d]$, and cross-modality self-attention is computed as $\mathbf{H}^d = \mathrm{softmax}(\mathbf{Q}^d(\mathbf{K}^d)^\top/\sqrt{D})\mathbf{V}^d$. This design allows image and action tokens to attend jointly. The final transformer output $\mathbf{H}^d$ is passed to a dataset-specific flow-matching action decoder $G_{\mathrm{act}}^{\,d}$, which outputs the denoised action at timestep $t-1$.

For robot trajectories, we provide both egocentric and wrist camera images; for human trajectories, only egocentric images are available, and all token positions corresponding to the wrist stream are masked out. Egocentric images are represented as $16\times16$ grids of image patches, while wrist images are spatially downsampled to $8\times8$ grids, with wrist positional embeddings obtained via bicubic interpolation. The complete model contains $1{,}362{,}151{,}162$ parameters, of which $412{,}442{,}352$ belong to the SigLIP visual encoder. Full architectural details are provided in Table~\ref{table:supp_architecture}, where the \textit{Self-attention} refers to the modality-specific self-attention layers.

\section{Policy inference hyperparameters} The action chunks used in conditional flow matching (CFM) are sampled at 30hz for a 1 second chunk, for a horizon during training of $H^t=30$. At inference, we execute open loop actions at half the training horizon, with $H^i = 15$. We also perform the same domain randomization at test time as train time (color jitter and random cropping) as we find this avoids an error mode of CFM, where the policy becomes stuck in a stationary point.

\section{Task Details}
\label{supp:tasks}

\subsection{Definitions}
\textbf{Pick:} The task involves grasping a spatula by its handle and lifting it off the table.

\noindent\textbf{Stack:} The task entails lifting one of two bowls and placing it on top of the other to form a stable stack.

\noindent\textbf{Pull:} The task involves pulling a mug into a fixed 6\,in\,$\times$\,6\,in square region marked on the table.

\noindent\textbf{Reorient:} The task entails reorienting a mug from a sideways pose into a stable upright configuration.

\noindent\textbf{Book:} The task involves picking up a book resting on a book holder and inserting it into an empty slot in a bookshelf.

\noindent\textbf{Pour:} The task entails lifting a small bowl filled with beads and pouring the beads into a larger bowl.

\subsection{Success Metrics}
\textbf{Pick:} A full success requires (1) lifting the spatula completely off the table, (2) securing the grasp on the handle, and (3) maintaining a stable hold after lifting.

\noindent\textbf{Stack:} A full success requires (1) grasping and lifting exactly one bowl and (2) placing it so that it rests stably on the other without either bowl tipping over.

\noindent\textbf{Pull:} A full success requires (1) the mug ending fully within the 6\,in\,$\times$\,6\,in square and (2) remaining within the region after all motion stops.

\noindent\textbf{Reorient:} A full success requires (1) lifting the mug from its sideways orientation and (2) placing it right-side up such that it remains upright after release.

\noindent\textbf{Book:} A full success requires (1) lifting the book without disturbing the book holder and (2) inserting it fully into the target slot so that it is closed and aligned with neighboring books.

\noindent\textbf{Pour:} A full success requires (1) lifting the small bowl without spilling beads during pickup and (2) transferring more than 50\% of the beads into the larger bowl.

\subsection{Initial Condition Sampling Distributions}
\textbf{Pick:} The spatula is initialized with a random pose (position and rotation each sampled uniformly at random) within the camera field of view (FOV).

\noindent\textbf{Stack:} Both bowls are initialized right-side up with positions sampled uniformly within the camera FOV.

\noindent\textbf{Pull:} The mug's position and rotation are sampled uniformly within the camera FOV; the square region remains fixed.

\noindent\textbf{Reorient:} The mug's pose is sampled uniformly within the camera FOV with a randomized rotation.

\noindent\textbf{Book:} The book's position and orientation are sampled randomly within the camera FOV.

\noindent\textbf{Pour:} The small and large bowls are initialized with positions sampled uniformly at random within the camera FOV.

\section{Environments}
See Tables \ref{tab:train-objects-grid1}, \ref{tab:train-objects-grid2}, \ref{tab:test-objects-grid} for exact environment descriptions, to be paired with the following visualizations.
\label{supp:environment_vis}
\subsection{Environment Visualizations}
Figures \ref{fig:env-grid}, \ref{fig:env-grid-1}, \ref{fig:env-grid-2}, \ref{fig:env-grid-0} exhibit the training environments, and Figures \ref{fig:env-grid-3}, \ref{fig:env-grid-4} exhibit the unseen testing environments, across all six tasks.

\newcommand{\envcell}[2]{%
  \begingroup
  \renewcommand{\arraystretch}{1}%
  \begin{tabular}[c]{@{}l@{}}#1\\[-0.15em]#2\end{tabular}%
  \endgroup
}

\begin{table}[t]
\caption{Training objects and tablecloths for each of the six tasks and first five training environments.}
\label{tab:train-objects-grid1}
\centering
\renewcommand{\arraystretch}{1.6}
\setlength{\tabcolsep}{4pt}
\small
\begin{tabular}{llllll}
\hline
\textbf{Task} & \textbf{Env 1} & \textbf{Env 2} & \textbf{Env 3} & \textbf{Env 4} & \textbf{Env 5} \\
\hline

\textbf{Pick} &
\envcell{Spatula 1}{Tablecloth 1} &
\envcell{Spatula 2}{Tablecloth 2} &
\envcell{Spatula 4}{Tablecloth 3} &
\envcell{Spatula 3}{Tablecloth 5} &
\envcell{Spatula 5}{Tablecloth 4} \\

\textbf{Stack} &
\envcell{Bowls (2,3)}{Tablecloth 1} &
\envcell{Bowls (2,8)}{Tablecloth 2} &
\envcell{Bowls (3,9)}{Tablecloth 3} &
\envcell{Bowls (1,3)}{Tablecloth 5} &
\envcell{Bowls (9,10)}{Tablecloth 4} \\

\textbf{Pull} &
\envcell{Mug 2}{Tablecloth 1} &
\envcell{Mug 4}{Tablecloth 2} &
\envcell{Mug 5}{Tablecloth 3} &
\envcell{Mug 3}{Tablecloth 11} &
\envcell{Mug 1}{Tablecloth 4} \\

\textbf{Reorient} &
\envcell{Mug 2}{Tablecloth 1} &
\envcell{Mug 4}{Tablecloth 2} &
\envcell{Mug 5}{Tablecloth 3} &
\envcell{Mug 3}{Tablecloth 5} &
\envcell{Mug 1}{Tablecloth 4} \\

\textbf{Book} &
\envcell{Book 1}{Tablecloth 1} &
\envcell{Book 4}{Tablecloth 2} &
\envcell{Book 3}{Tablecloth 3} &
\envcell{Book 5}{Tablecloth 11} &
\envcell{Book 2}{Tablecloth 4} \\

\textbf{Pour} &
\envcell{Bowls (4,8)}{Tablecloth 1} &
\envcell{Bowls (3,8)}{Tablecloth 2} &
\envcell{Bowls (1,9)}{Tablecloth 3} &
\envcell{Bowls (5,9)}{Tablecloth 11} &
\envcell{Bowls (2,10)}{Tablecloth 4} \\
\hline
\end{tabular}
\end{table}

\begin{table}[t]
\caption{Training objects and tablecloths for each of the six tasks and last five training environments.}
\label{tab:train-objects-grid2}
\centering
\renewcommand{\arraystretch}{1.6}
\setlength{\tabcolsep}{4pt}
\small
\begin{tabular}{llllll}
\hline
\textbf{Task} & \textbf{Env 6} & \textbf{Env 7} & \textbf{Env 8} & \textbf{Env 9} & \textbf{Env 10} \\
\hline

\textbf{Pick} &
\envcell{Spatula 6}{Tablecloth 6} &
\envcell{Spatula 7}{Tablecloth 7} &
\envcell{Spatula 8}{Tablecloth 8} &
\envcell{Spatula 9}{Tablecloth 9} &
\envcell{Spatula 10}{Tablecloth 10} \\

\textbf{Stack} &
\envcell{Bowls (5,6)}{Tablecloth 6} &
\envcell{Bowls (4,5)}{Tablecloth 7} &
\envcell{Bowls (6,7)}{Tablecloth 8} &
\envcell{Bowls (4,7)}{Tablecloth 9} &
\envcell{Bowls (8,9)}{Tablecloth 10} \\

\textbf{Pull} &
\envcell{Mug 6}{Tablecloth 6} &
\envcell{Mug 7}{Tablecloth 7} &
\envcell{Mug 8}{Tablecloth 8} &
\envcell{Mug 9}{Tablecloth 9} &
\envcell{Mug 10}{Tablecloth 10} \\

\textbf{Reorient} &
\envcell{Mug 6}{Tablecloth 6} &
\envcell{Mug 7}{Tablecloth 7} &
\envcell{Mug 8}{Tablecloth 8} &
\envcell{Mug 9}{Tablecloth 9} &
\envcell{Mug 10}{Tablecloth 10} \\

\textbf{Book} &
\envcell{Book 6}{Tablecloth 6} &
\envcell{Book 7}{Tablecloth 7} &
\envcell{Book 8}{Tablecloth 8} &
\envcell{Book 9}{Tablecloth 9} &
\envcell{Book 10}{Tablecloth 10} \\

\textbf{Pour} &
\envcell{Bowls (5,8)}{Tablecloth 6} &
\envcell{Bowls (3,9)}{Tablecloth 7} &
\envcell{Bowls (4,9)}{Tablecloth 8} &
\envcell{Bowls (5,9)}{Tablecloth 9} &
\envcell{Bowls (6,10)}{Tablecloth 10} \\
\hline
\end{tabular}
\end{table}

\begin{table}[t]
\caption{Testing objects and tablecloths for each of the six tasks and four test environments.}
\label{tab:test-objects-grid}
\centering
\renewcommand{\arraystretch}{1.6}
\setlength{\tabcolsep}{4pt}
\small
\begin{tabular}{lllll}
\hline
\textbf{Task} & \textbf{Env 1} & \textbf{Env 2} & \textbf{Env 3} & \textbf{Env 4} \\
\hline

\textbf{Pick} &
\envcell{Spatula 11}{Tablecloth 12} &
\envcell{Spatula 12}{Tablecloth 12} &
\envcell{Spatula 11}{Tablecloth 11} &
\envcell{Spatula 12}{Tablecloth 11} \\[0.4em]

\textbf{Stack} &
\envcell{Bowls (5,7)}{Tablecloth 12} &
\envcell{Bowls (4,6)}{Tablecloth 12} &
\envcell{Bowls (5,7)}{Tablecloth 11} &
\envcell{Bowls (4,6)}{Tablecloth 11}\\[0.4em]

\textbf{Pull} &
\envcell{Mug 11}{Tablecloth 12} &
\envcell{Mug 12}{Tablecloth 12} &
\envcell{Mug 11}{Tablecloth 5} &
\envcell{Mug 12}{Tablecloth 5} \\[0.4em]

\textbf{Reorient} &
\envcell{Mug 12}{Tablecloth 12} &
\envcell{Mug 13}{Tablecloth 12} &
\envcell{Mug 12}{Tablecloth 11} &
\envcell{Mug 11}{Tablecloth 11} \\[0.4em]

\textbf{Book} &
\envcell{Book 11}{Tablecloth 12} &
\envcell{Book 12}{Tablecloth 12} &
\envcell{Book 11}{Tablecloth 5} &
\envcell{Book 12}{Tablecloth 5} \\[0.4em]

\textbf{Pour} &
\envcell{Bowls (3,10)}{Tablecloth 12} &
\envcell{Bowls (2,8)}{Tablecloth 12} &
\envcell{Bowls (3,10)}{Tablecloth 5} &
\envcell{Bowls (2,8)}{Tablecloth 5} \\
\hline
\end{tabular}
\end{table}

\newlength{\snapw}
\setlength{\snapw}{0.19\textwidth}
\newlength{\gap}
\setlength{\gap}{1pt}
\newcommand{\envsnap}[1]{\includegraphics[width=\snapw]{#1}}

\setlength{\tabcolsep}{0pt}

\begin{figure}[t]
    \centering
    {%
    \renewcommand{\arraystretch}{0}%

    \begin{tabular}{@{}c@{\hspace{\gap}}c@{\hspace{\gap}}c@{\hspace{\gap}}c@{\hspace{\gap}}c@{}}

        \envsnap{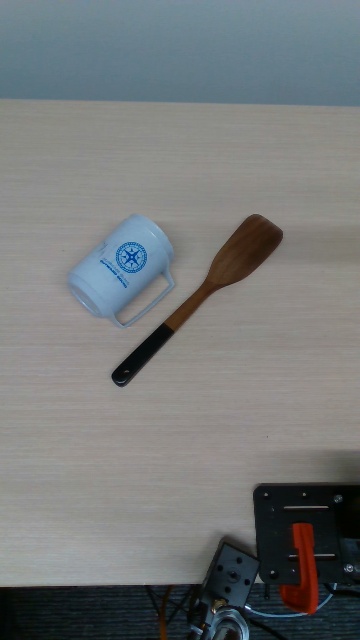} &
        \envsnap{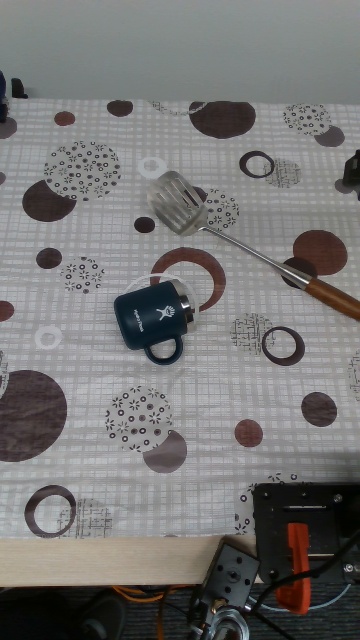} &
        \envsnap{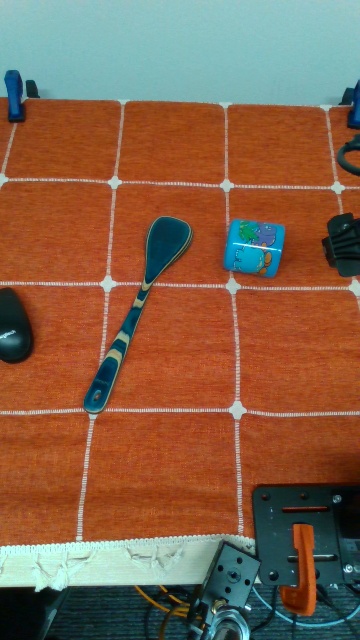} &
        \envsnap{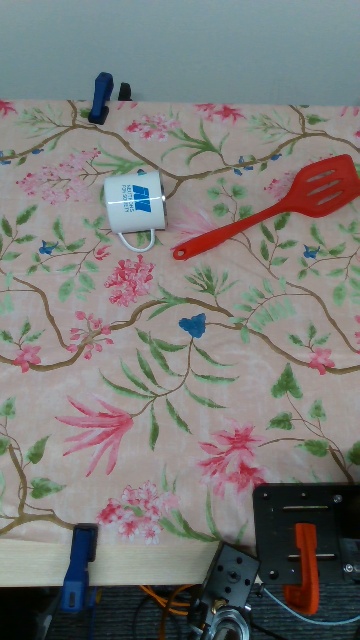} &
        \envsnap{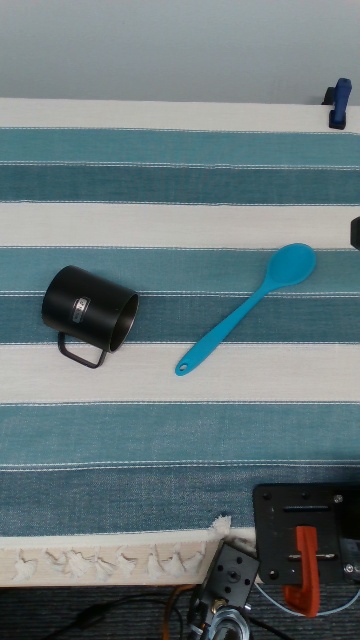} \\
        [\gap]

        \envsnap{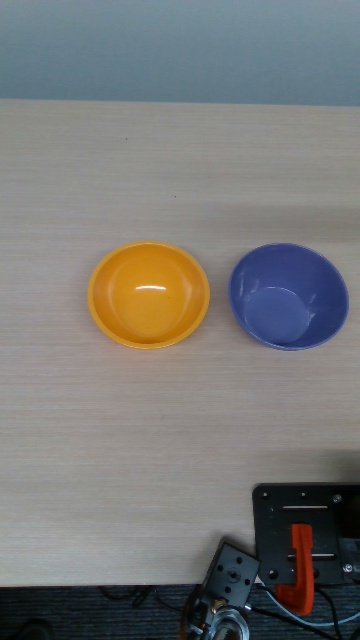} &
        \envsnap{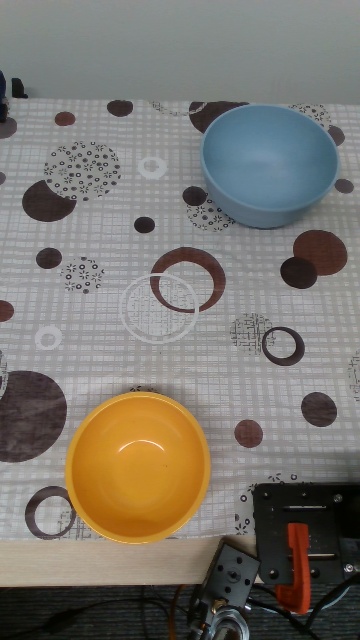} &
        \envsnap{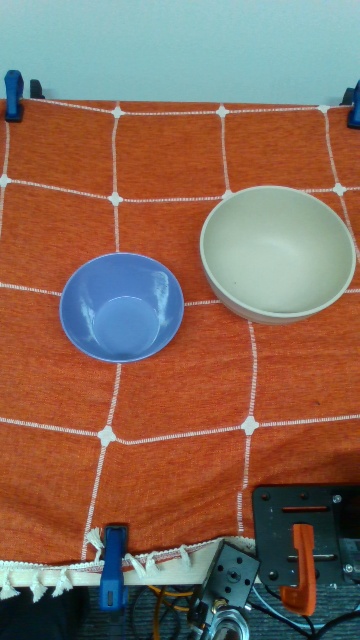} &
        \envsnap{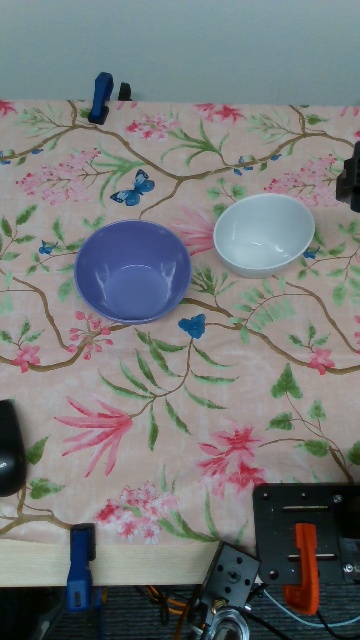} &
        \envsnap{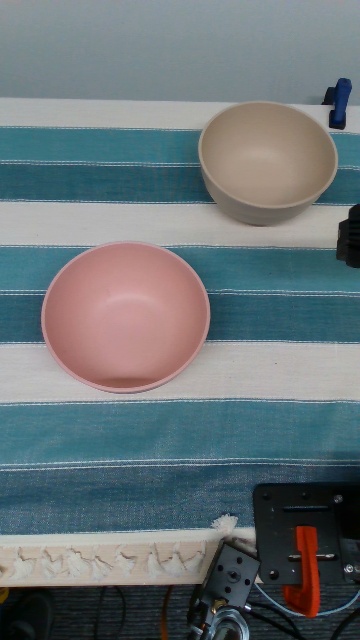} \\
        [\gap]

        \envsnap{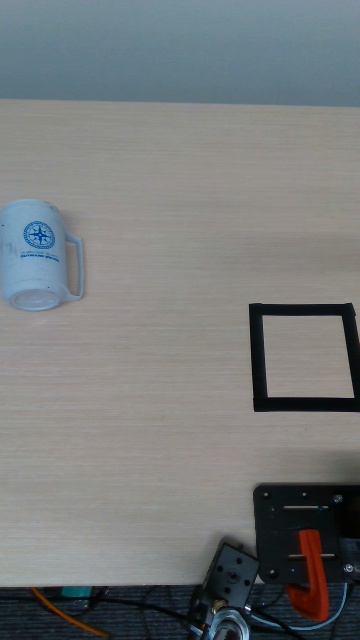} &
        \envsnap{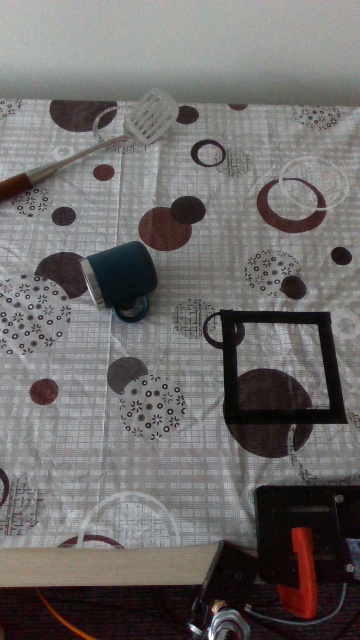} &
        \envsnap{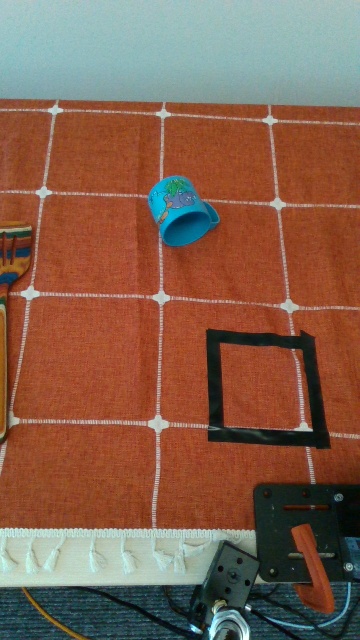} &
        \envsnap{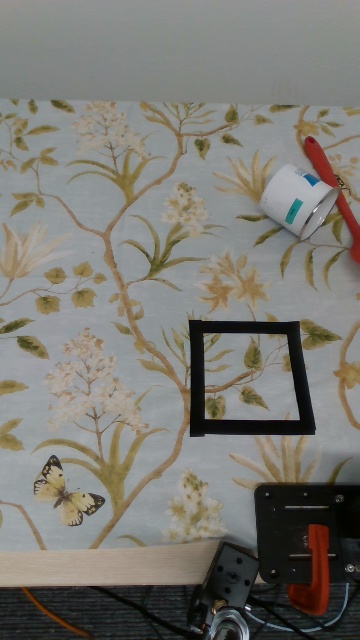} &
        \envsnap{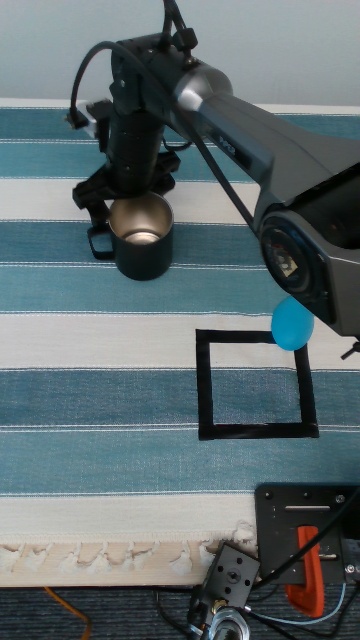} \\
    \end{tabular}%
    }

    \caption{Training environments across first three tasks (\textbf{Pick}, \textbf{Stack}, \textbf{Pull} top to bottom) and first five scenes per task (columns).}
    \label{fig:env-grid}
\end{figure}

\begin{figure}[t]
    \centering
    {%
    \renewcommand{\arraystretch}{0}%

    \begin{tabular}{@{}c@{\hspace{\gap}}c@{\hspace{\gap}}c@{\hspace{\gap}}c@{\hspace{\gap}}c@{}}

        \envsnap{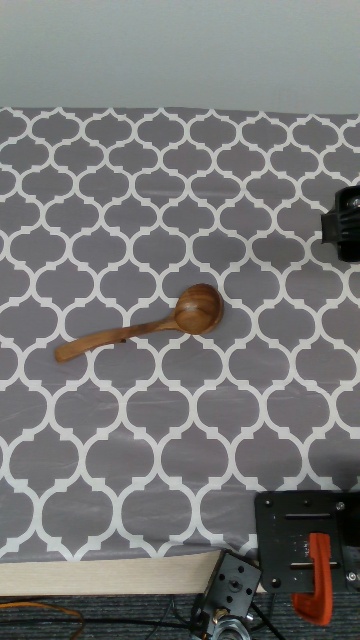} &
        \envsnap{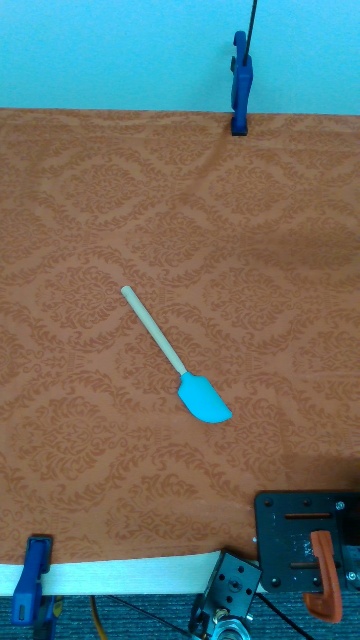} &
        \envsnap{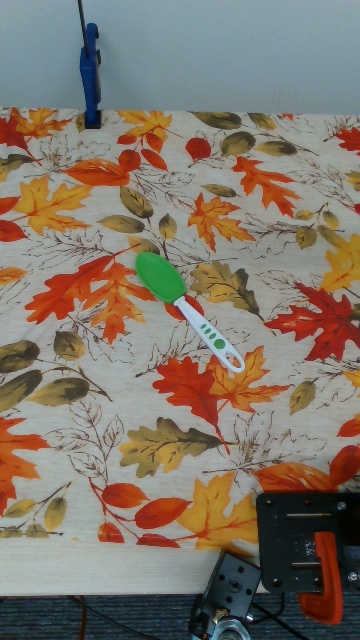} &
        \envsnap{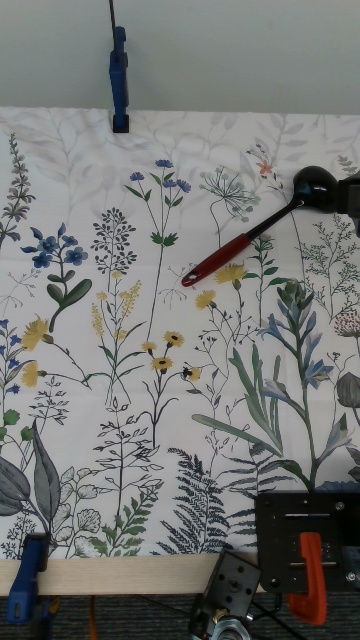} &
        \envsnap{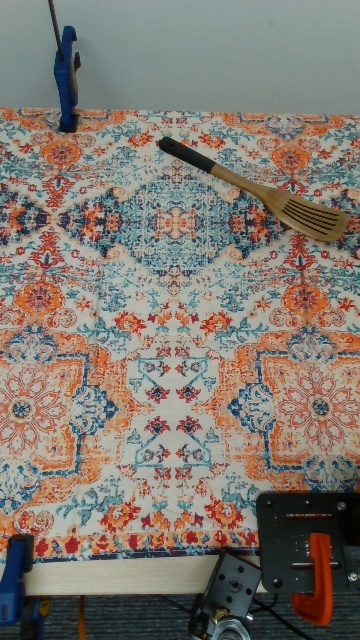} \\
        [\gap]

        \envsnap{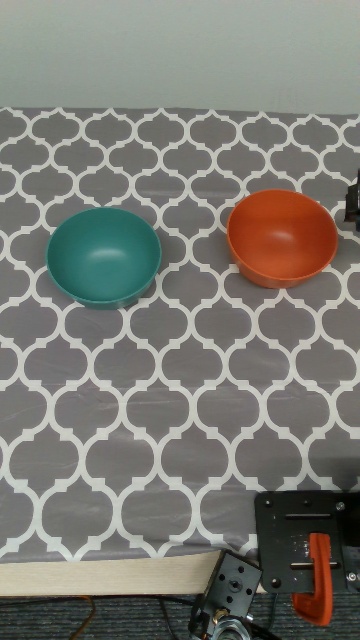} &
        \envsnap{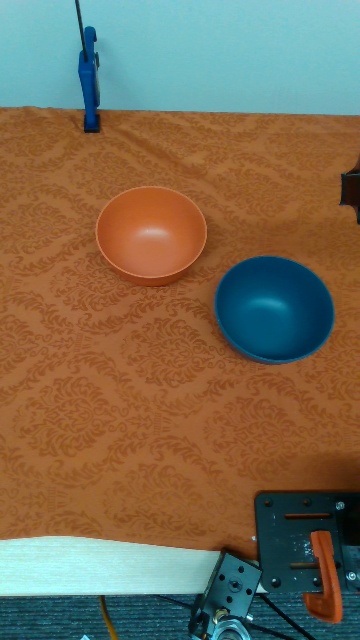} &
        \envsnap{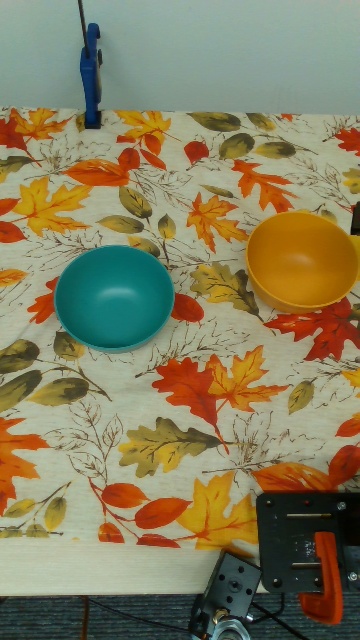} &
        \envsnap{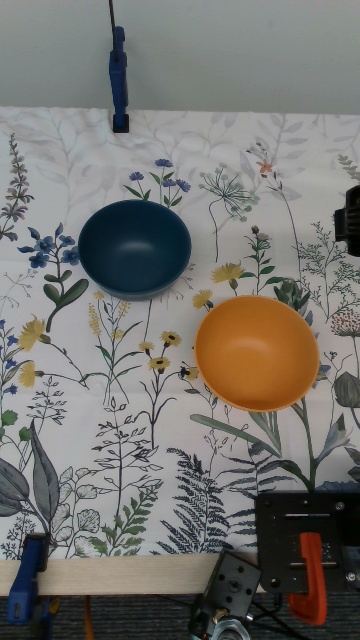} &
        \envsnap{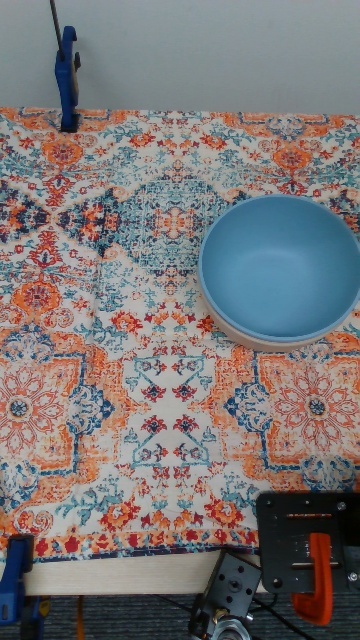} \\
        [\gap]

        \envsnap{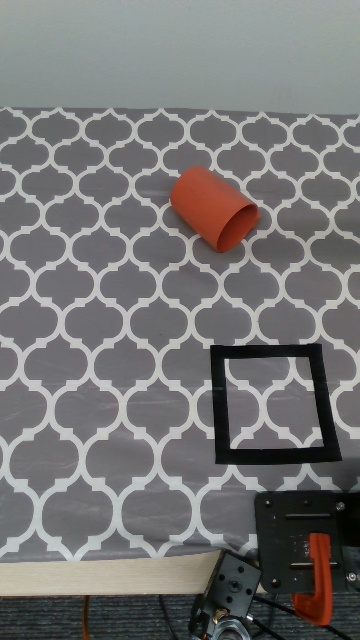} &
        \envsnap{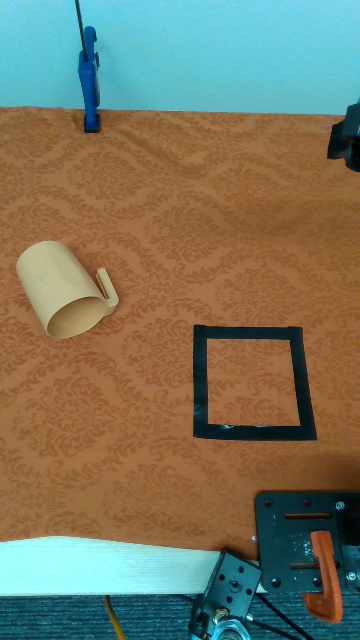} &
        \envsnap{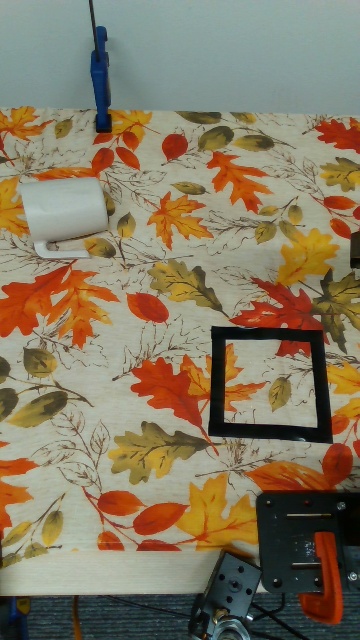} &
        \envsnap{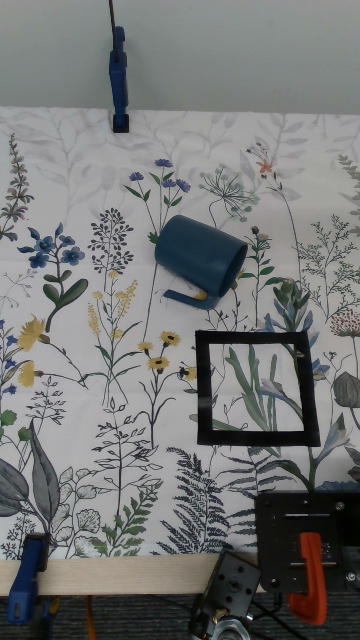} &
        \envsnap{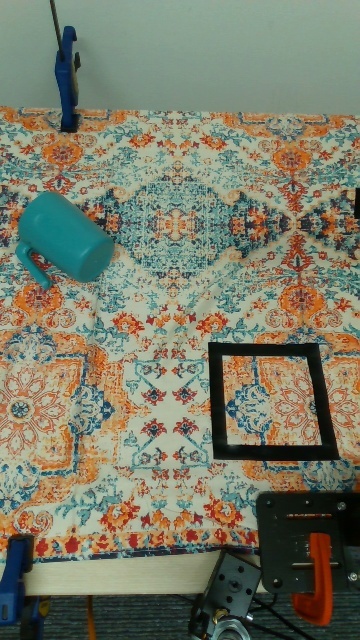} \\
    \end{tabular}%
    }

    \caption{Training environments across first three tasks (\textbf{Pick}, \textbf{Stack}, \textbf{Pull} top to bottom) and last five scenes per task (columns).}
    \label{fig:env-grid-1}
\end{figure}

\begin{figure}[t]
    \centering
    {%
    \renewcommand{\arraystretch}{0}%

    \begin{tabular}{@{}c@{\hspace{\gap}}c@{\hspace{\gap}}c@{\hspace{\gap}}c@{\hspace{\gap}}c@{}}

        \envsnap{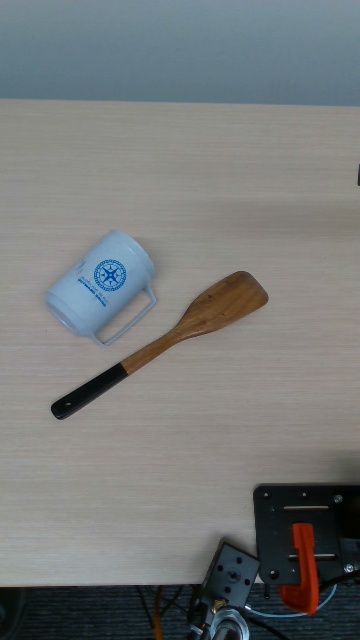} &
        \envsnap{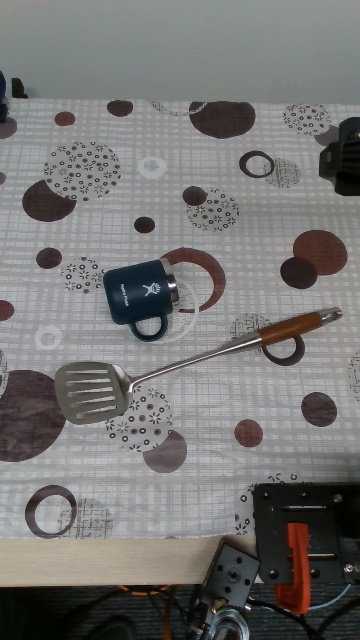} &
        \envsnap{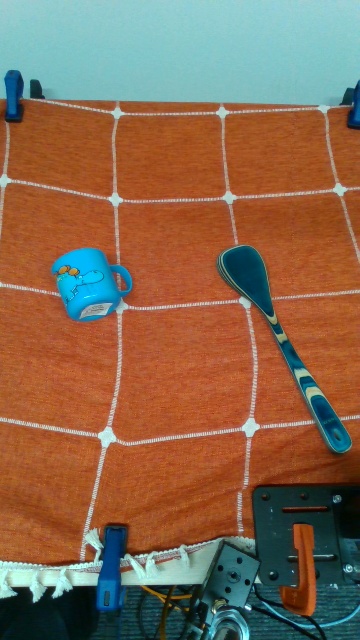} &
        \envsnap{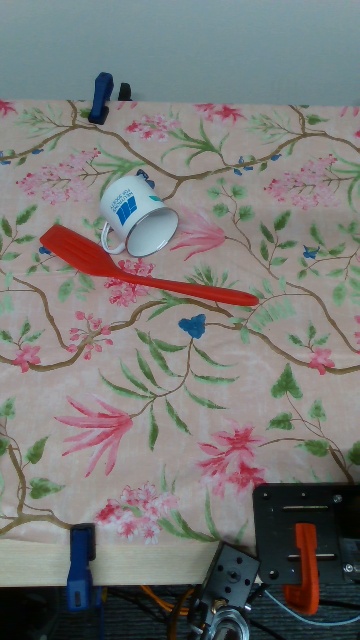} &
        \envsnap{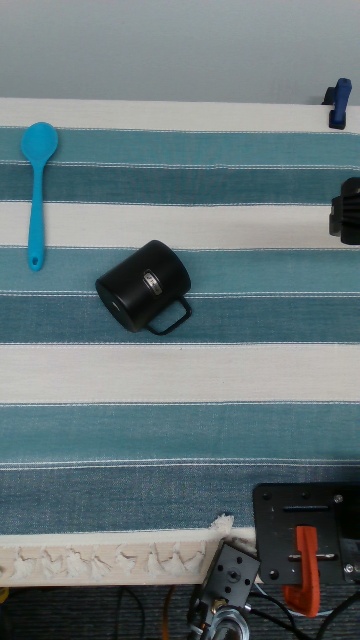} \\
        [\gap]

        \envsnap{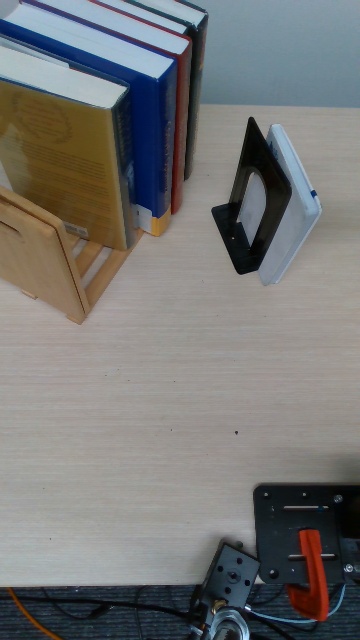} &
        \envsnap{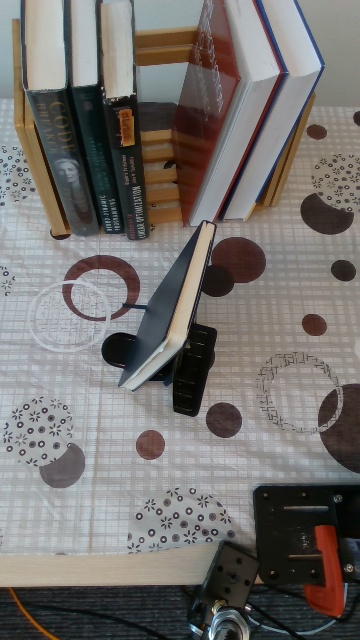} &
        \envsnap{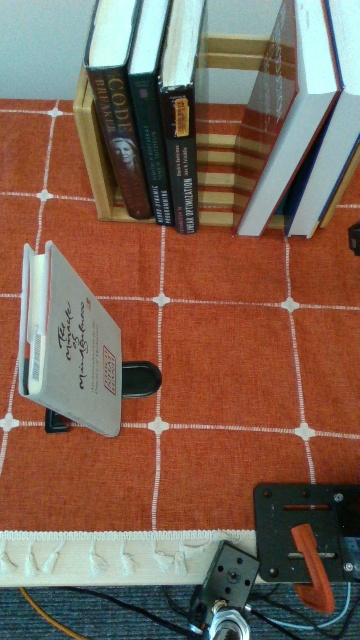} &
        \envsnap{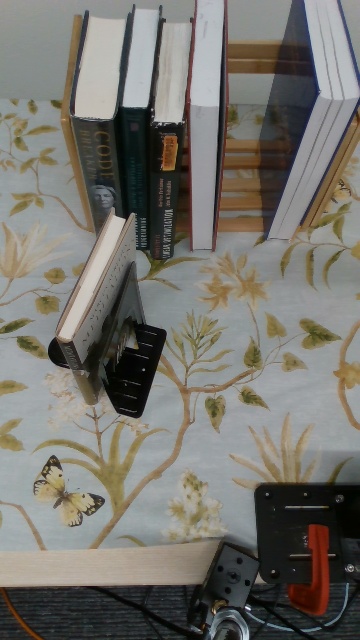} &
        \envsnap{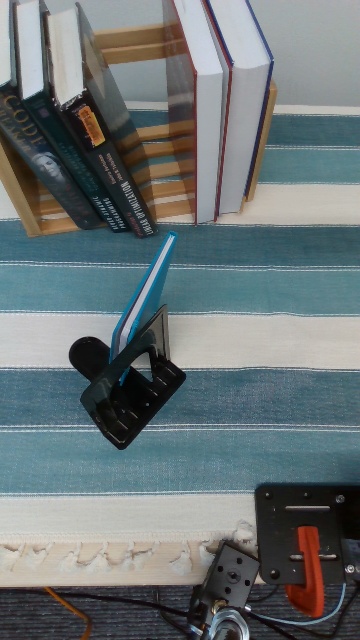} \\
        [\gap]

        \envsnap{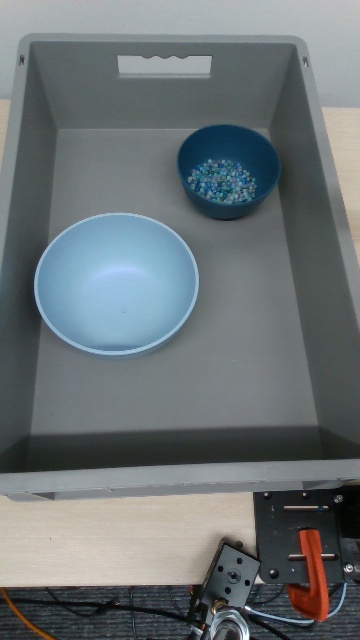} &
        \envsnap{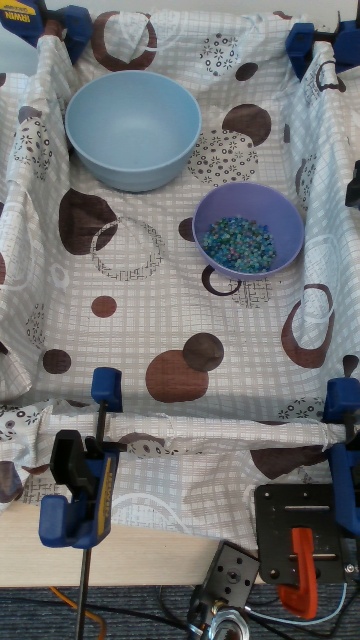} &
        \envsnap{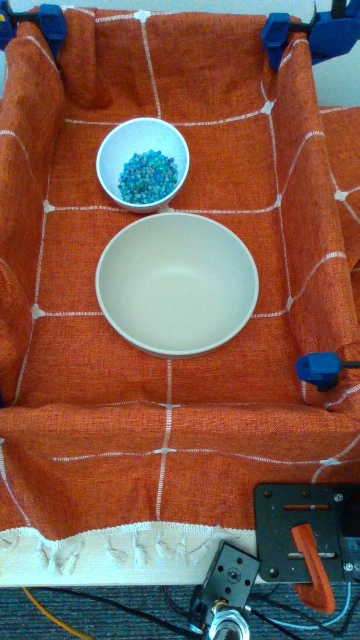} &
        \envsnap{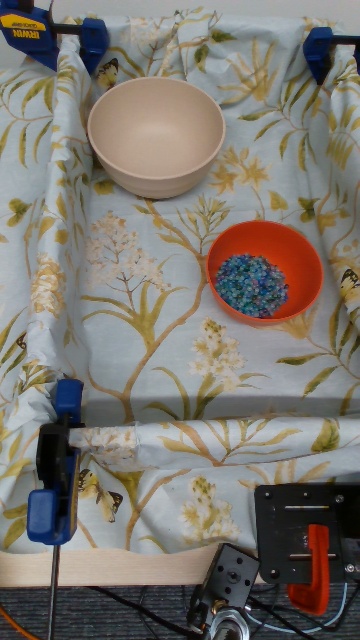} &
        \envsnap{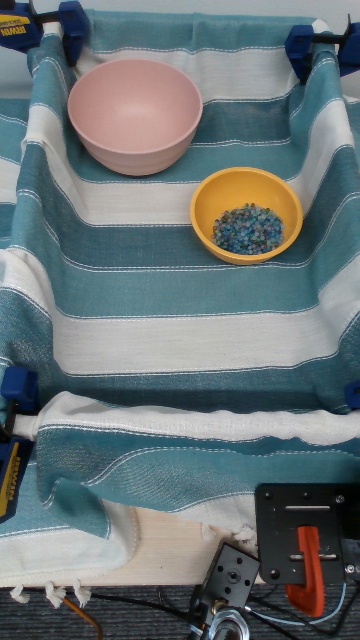}
    \end{tabular}%
    }

    \caption{Training environments across last three tasks (\textbf{Reorient}, \textbf{Book}, \textbf{Pour} top to bottom) and first five scenes per task (columns).}
    \label{fig:env-grid-2}
\end{figure}

\begin{figure}[t]
    \centering
    {%
    \renewcommand{\arraystretch}{0}%

    \begin{tabular}{@{}c@{\hspace{\gap}}c@{\hspace{\gap}}c@{\hspace{\gap}}c@{\hspace{\gap}}c@{}}

        \envsnap{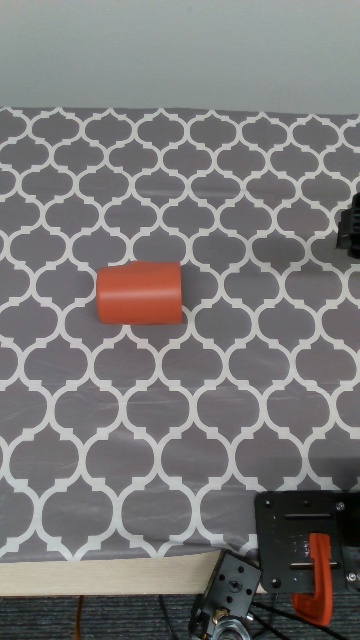} &
        \envsnap{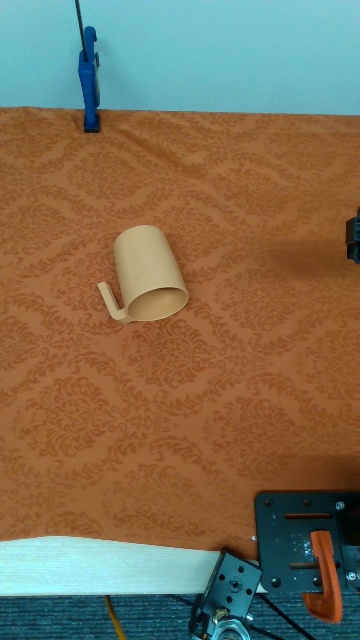} &
        \envsnap{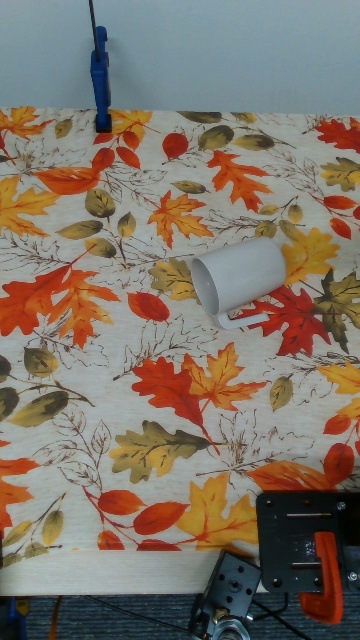} &
        \envsnap{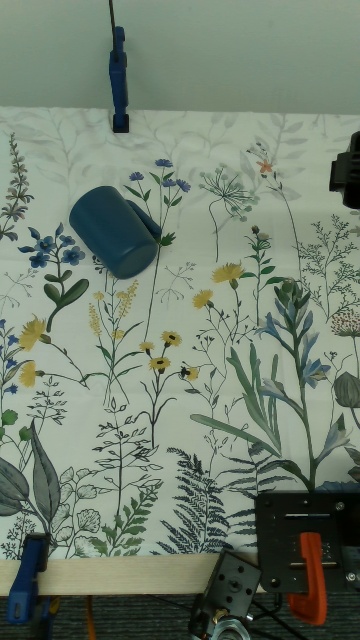} &
        \envsnap{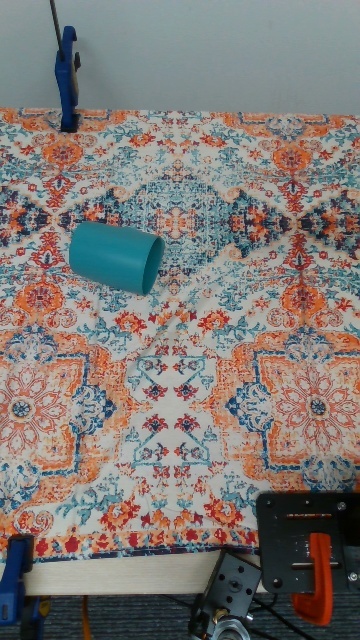} \\
        [\gap]

        \envsnap{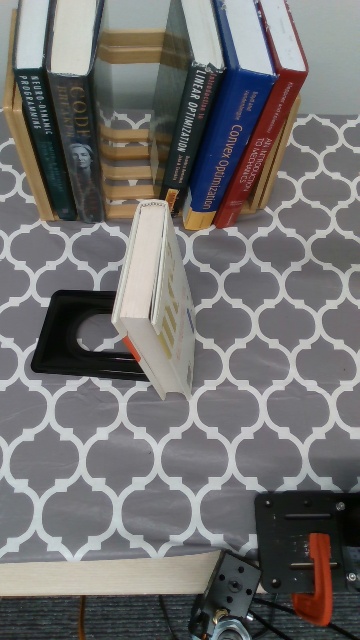} &
        \envsnap{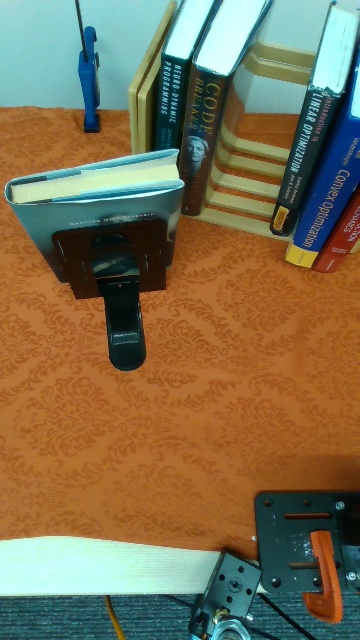} &
        \envsnap{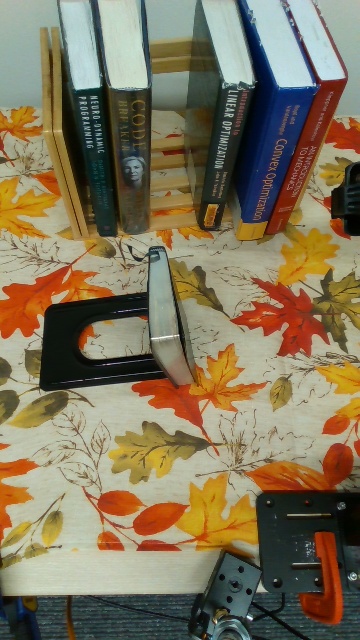} &
        \envsnap{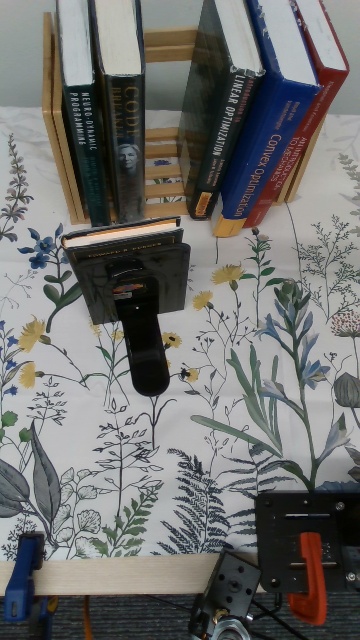} &
        \envsnap{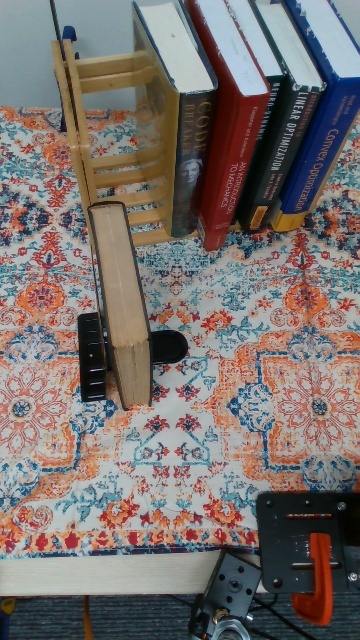} \\
        [\gap]

        \envsnap{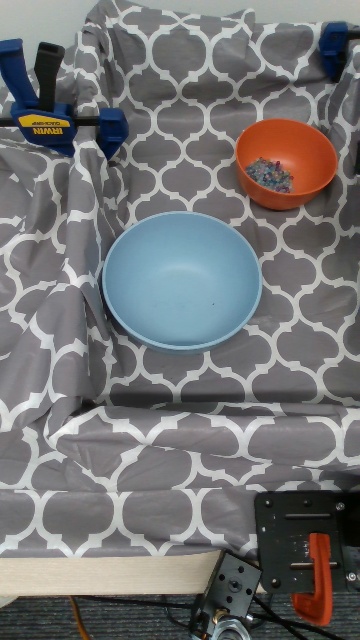} &
        \envsnap{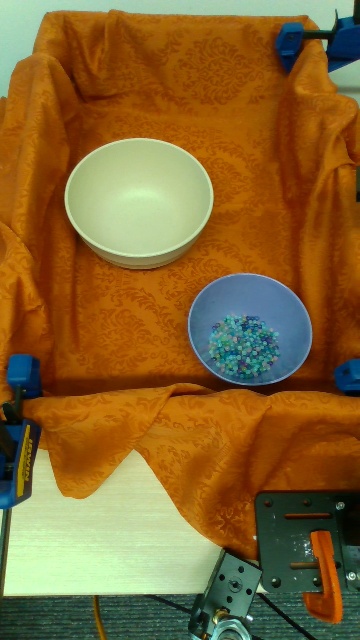} &
        \envsnap{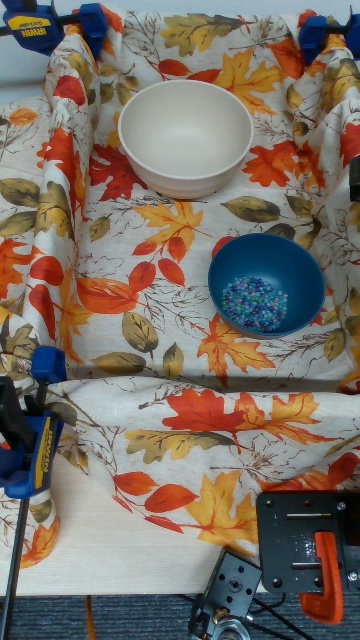} &
        \envsnap{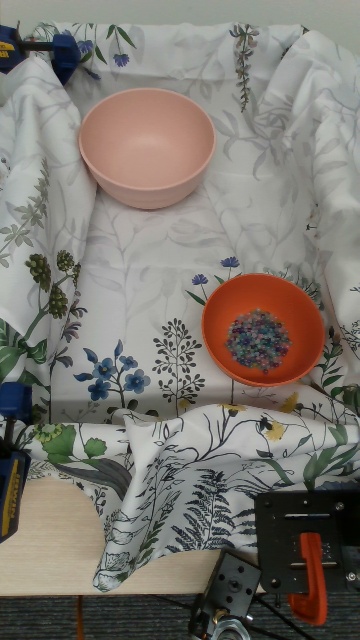} &
        \envsnap{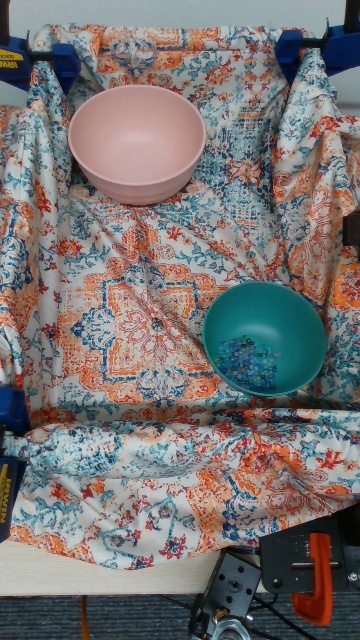}
    \end{tabular}%
    }

    \caption{Training environments across last three tasks (\textbf{Reorient}, \textbf{Book}, \textbf{Pour} top to bottom) and last five scenes per task (columns).}
    \label{fig:env-grid-0}
\end{figure}

\begin{figure}[t]
    \centering
    {%
    \renewcommand{\arraystretch}{0}%

    \begin{tabular}{@{}c@{\hspace{\gap}}c@{\hspace{\gap}}c@{\hspace{\gap}}c@{}}

        \envsnap{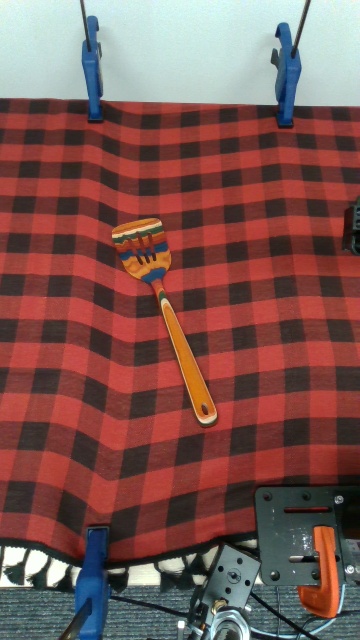} &
        \envsnap{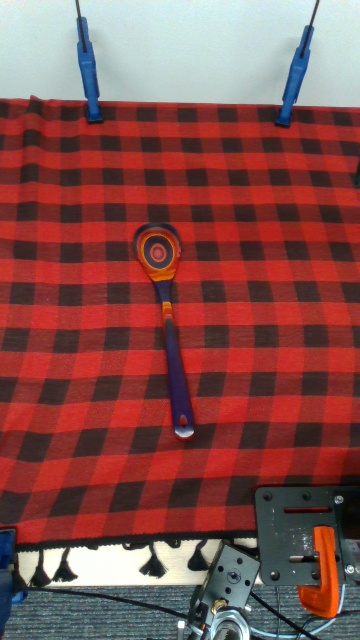} &
        \envsnap{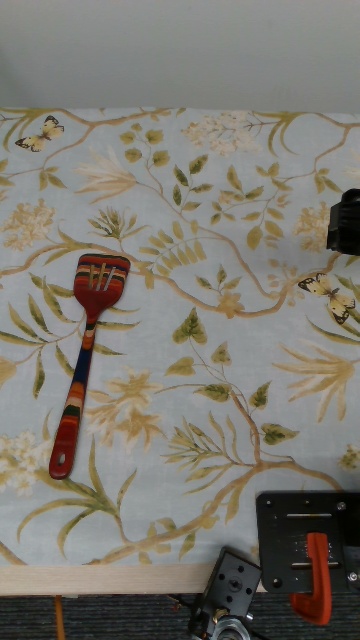} &
        \envsnap{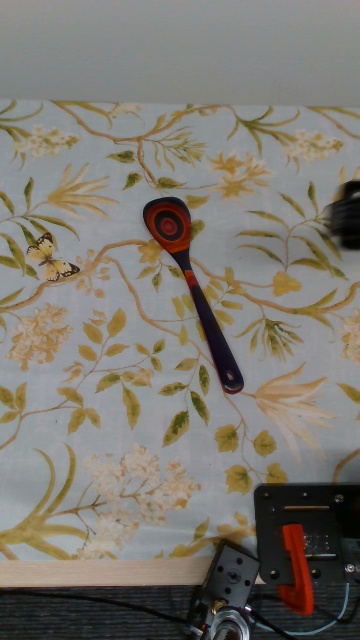} \\
        [\gap]

        \envsnap{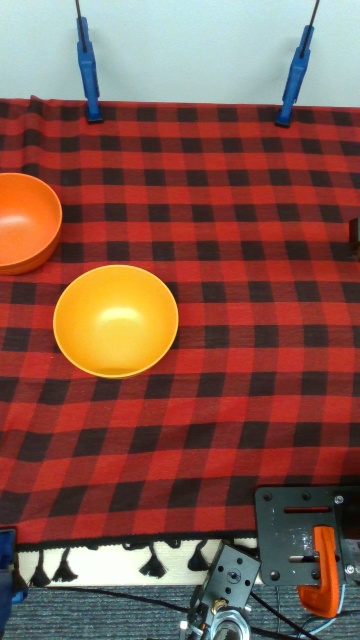} &
        \envsnap{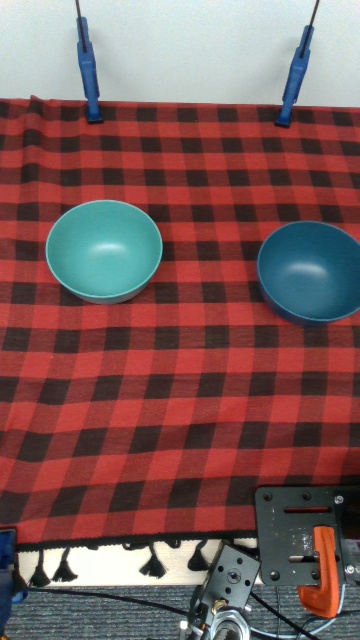} &
        \envsnap{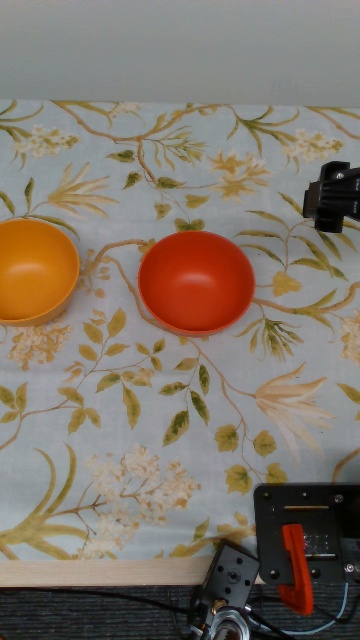} &
        \envsnap{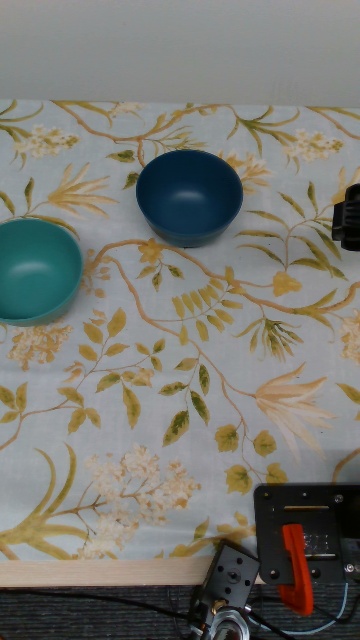} \\
        [\gap]

        \envsnap{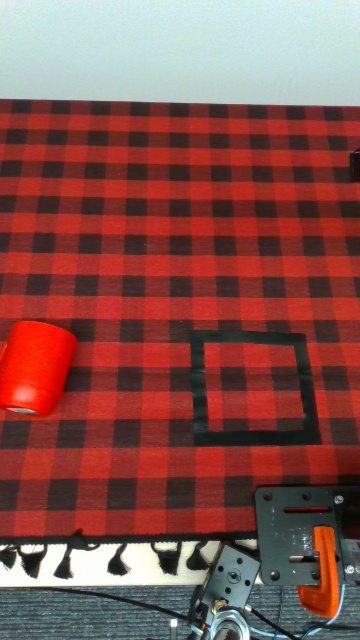} &
        \envsnap{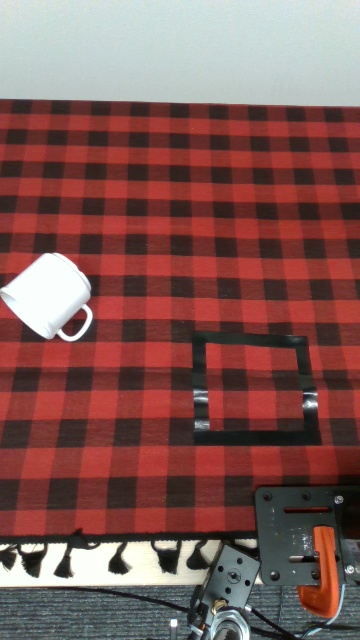} &
        \envsnap{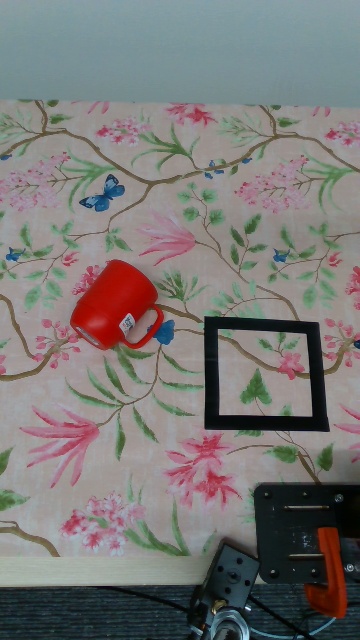} &
        \envsnap{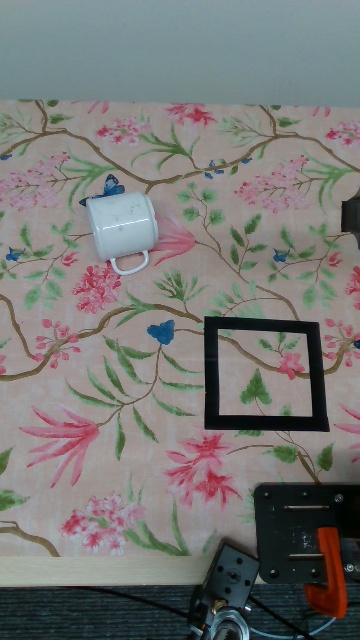}
    \end{tabular}%
    }

    \caption{Test environments across first three tasks (\textbf{Pick}, \textbf{Stack}, \textbf{Pull} top to bottom) and four scenes per task (columns).}
    \label{fig:env-grid-3}
\end{figure}

\begin{figure}[t]
    \centering
    {%
    \renewcommand{\arraystretch}{0}%

    \begin{tabular}{@{}c@{\hspace{\gap}}c@{\hspace{\gap}}c@{\hspace{\gap}}c@{}}

        \envsnap{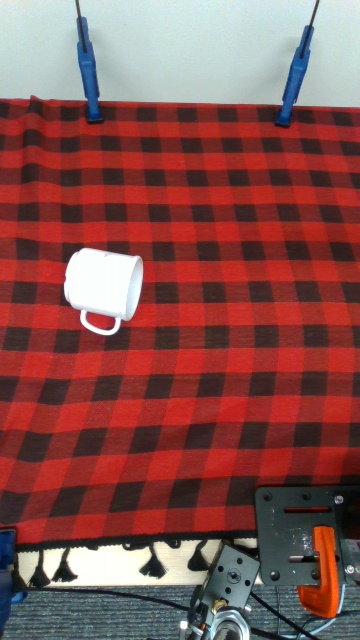} &
        \envsnap{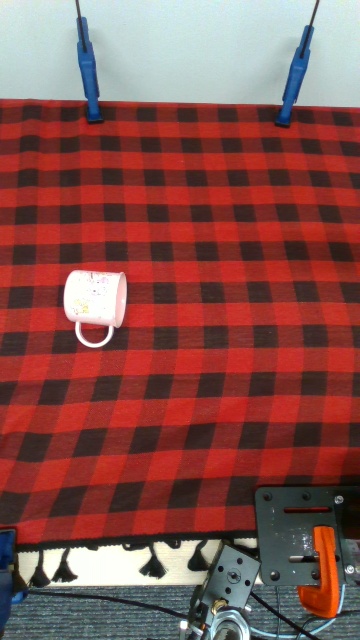} &
        \envsnap{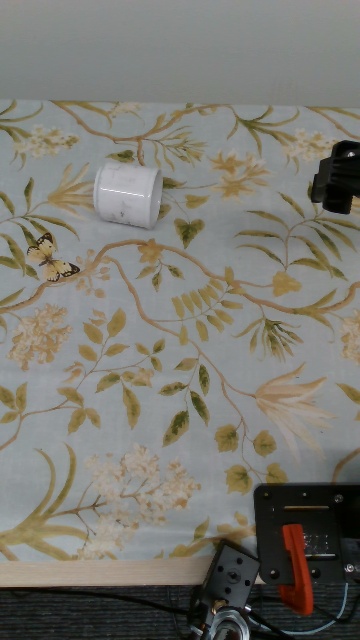} &
        \envsnap{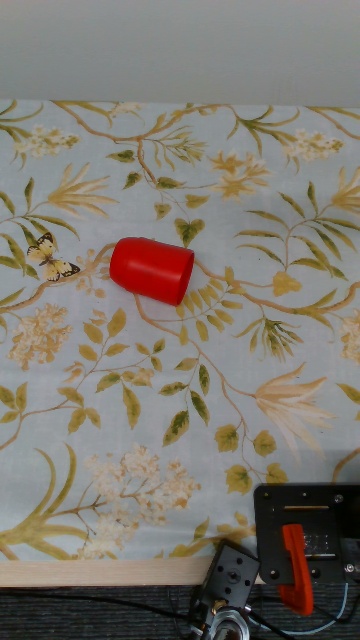} \\
        [\gap]

        \envsnap{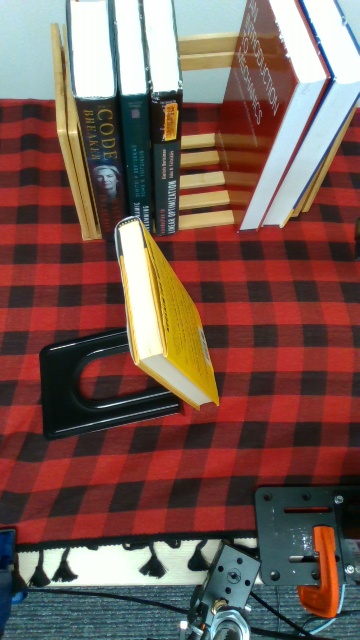} &
        \envsnap{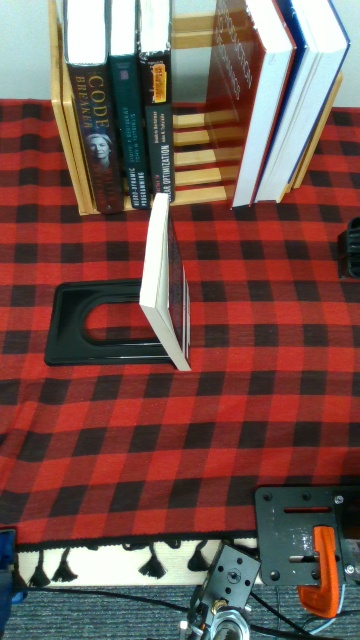} &
        \envsnap{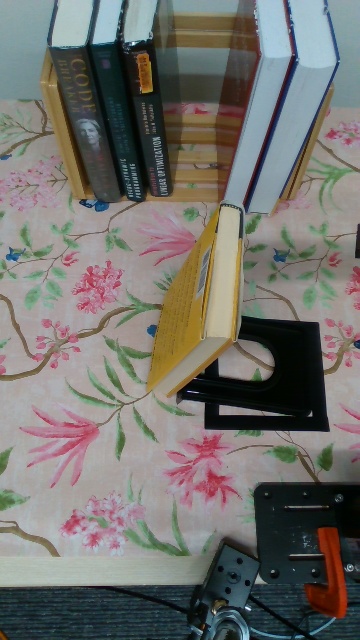} &
        \envsnap{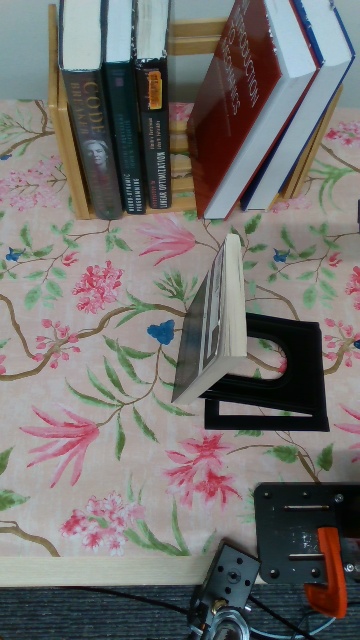} \\
        [\gap]

        \envsnap{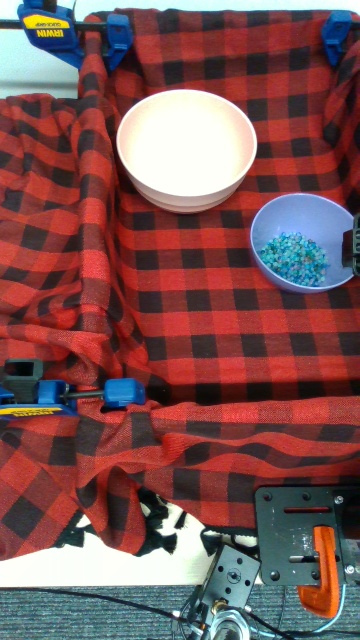} &
        \envsnap{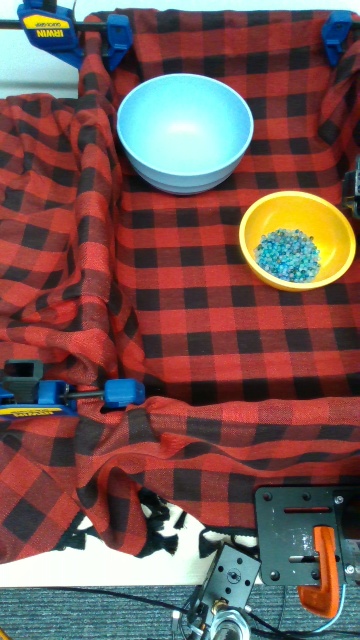} &
        \envsnap{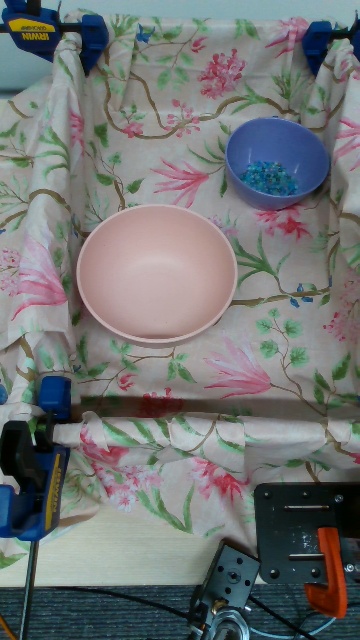} &
        \envsnap{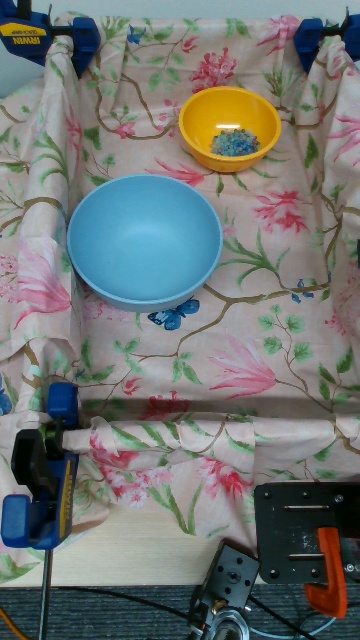}
    \end{tabular}%
    }

    \caption{Test environments across last three tasks (\textbf{Reorient}, \textbf{Book}, \textbf{Pour} top to bottom) and four scenes per task (columns).}
    \label{fig:env-grid-4}
\end{figure}

\subsection{Object Visualizations}
Figures~\ref{fig:tablecloths-train-test},
\ref{fig:spatulas-train-test},
\ref{fig:mugs-train-test},
\ref{fig:bowls-train-test},
\ref{fig:books-train-test},
 exhibit the training and testing objects for each of the six tasks. For each task, we detail the full set of objects actively involved in completing that task. This includes both objects the agent directly manipulates and objects that serve as targets or receivers of the manipulation. For example, in \textbf{Stack} this includes both bowls, in \textbf{Pour} it includes the small bead-filled bowl and the larger receiving bowl, and in \textbf{Book} it includes the book and the bookshelf slot.

\setlength{\fboxsep}{3pt}
\setlength{\fboxrule}{1.5pt}

\newlength{\imgw}
\newlength{\gapH}
\newlength{\gapV}
\newlength{\totw}

\newcommand{\objimg}[2]{%
  \begin{minipage}[b]{\imgw}%
    \centering
    \includegraphics[width=\linewidth]{#1}\\[-0.3em]
    {\scriptsize #2}%
  \end{minipage}%
}

\begin{figure}[!t]
    \setlength{\imgw}{0.18\textwidth}
    \setlength{\gapH}{2pt}
    \setlength{\gapV}{1pt}
    \setlength{\totw}{5\imgw}
    \addtolength{\totw}{4\gapH}

    \fbox{%
      \begin{minipage}{\totw}
        \setlength{\parskip}{0pt}

        \noindent
        \objimg{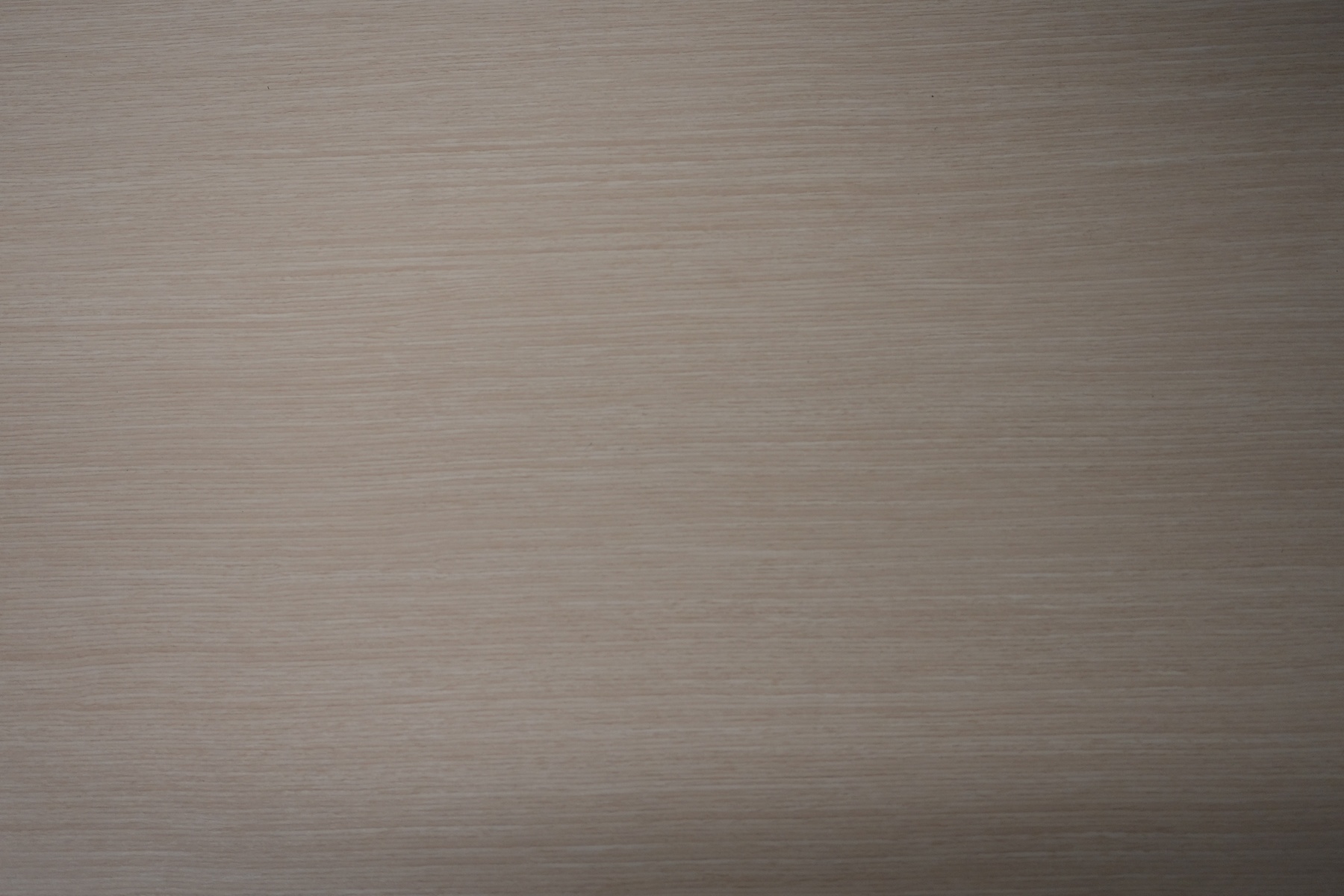}{1}\hspace{\gapH}%
        \objimg{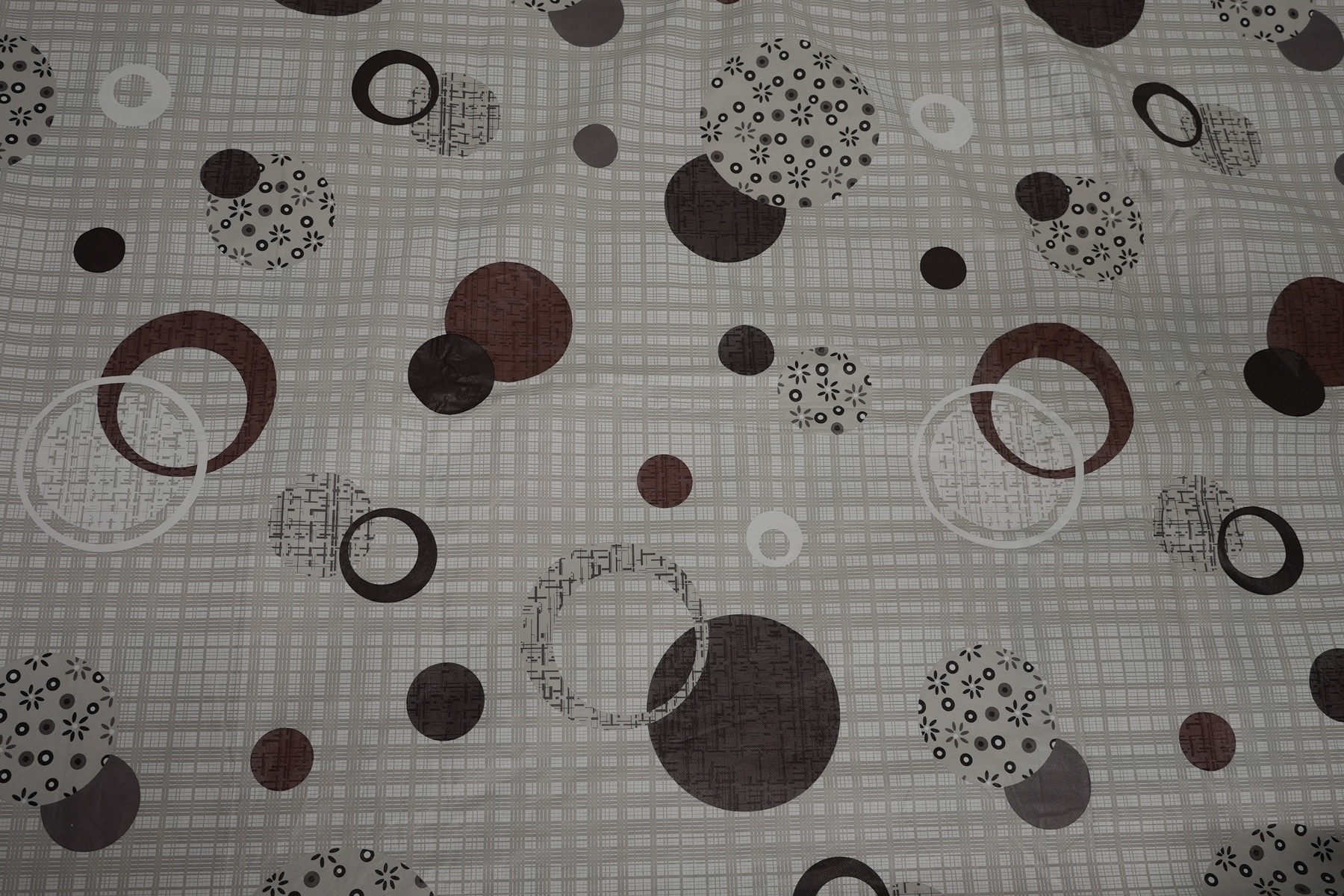}{2}\hspace{\gapH}%
        \objimg{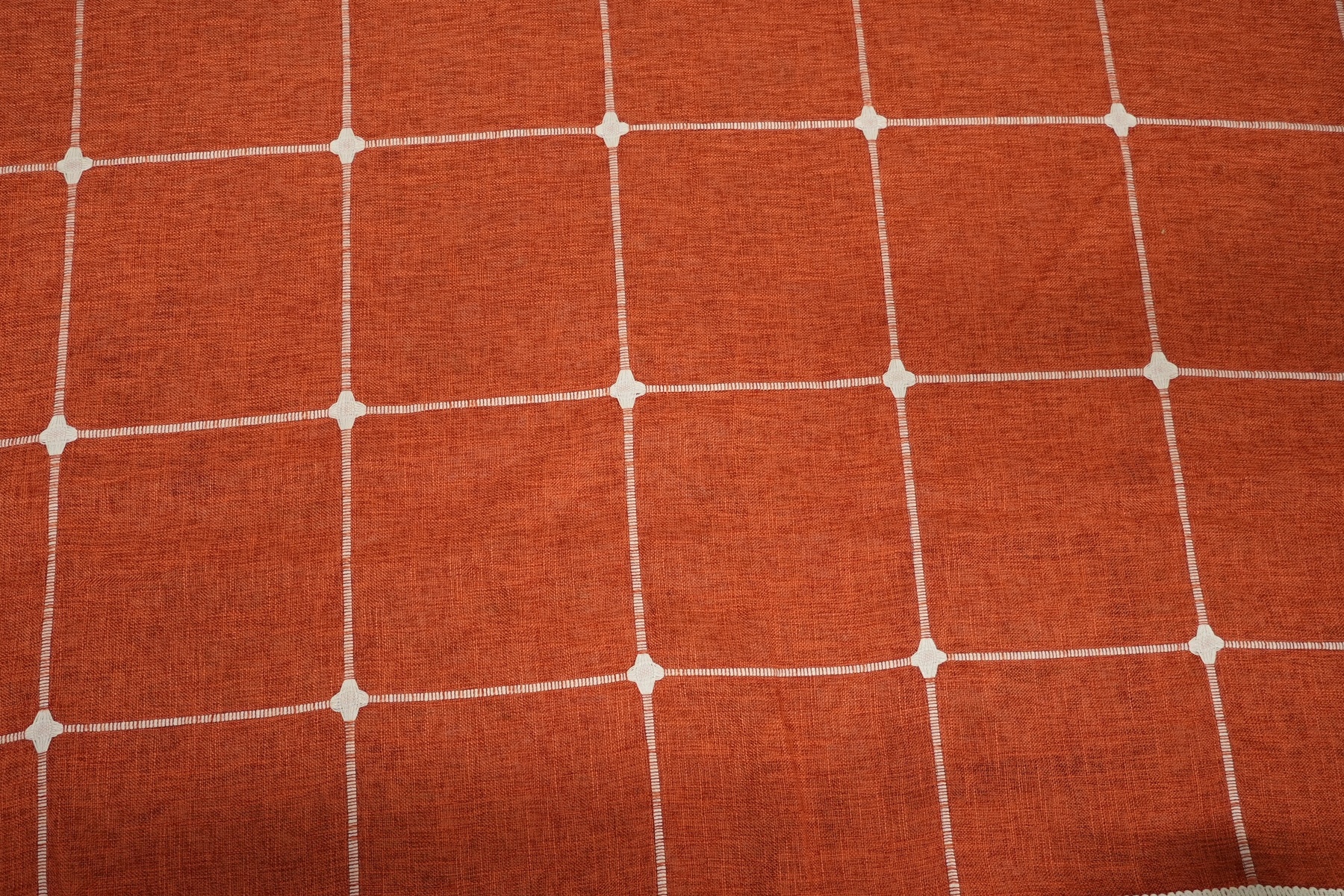}{3}\hspace{\gapH}%
        \objimg{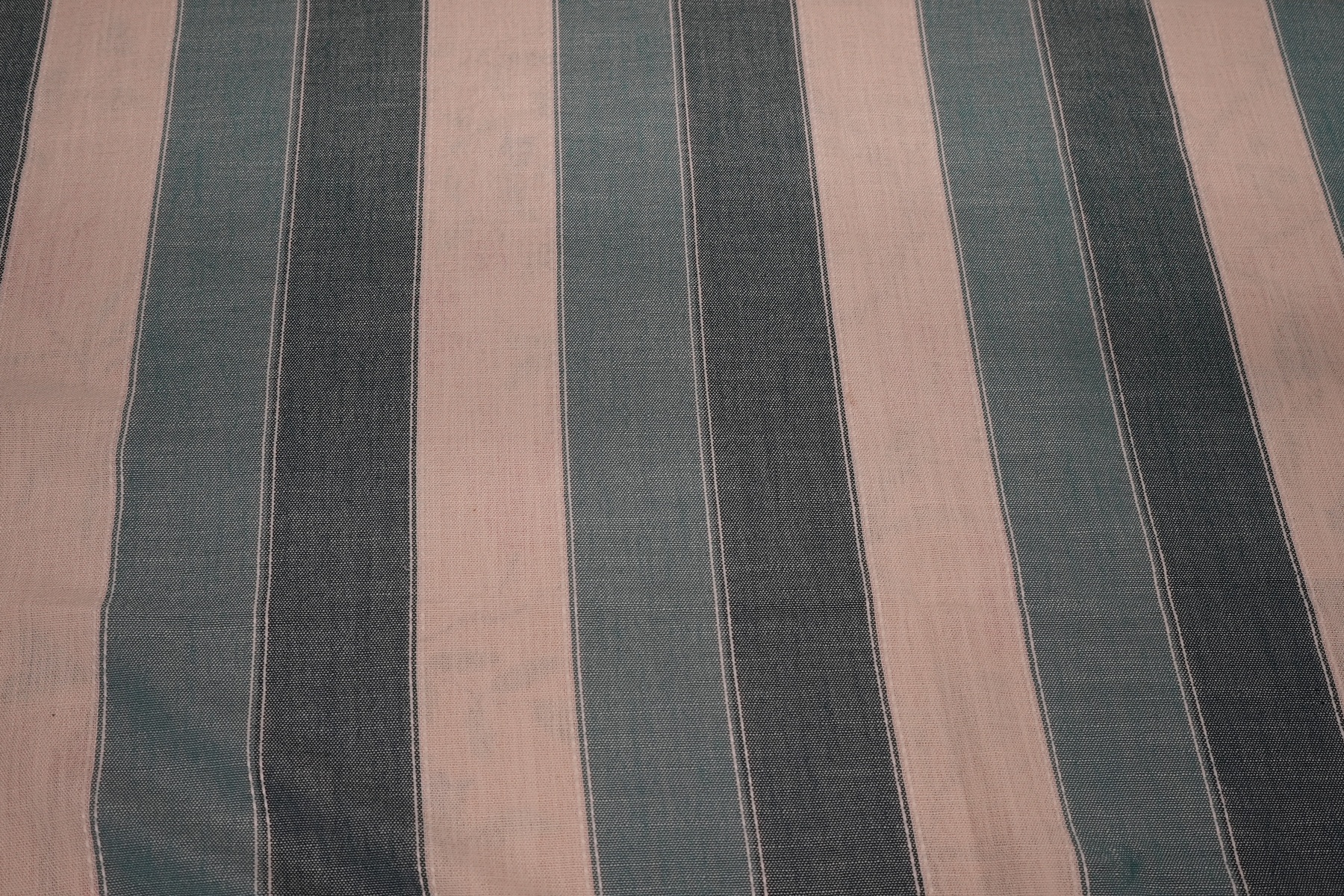}{4}\hspace{\gapH}%
        \objimg{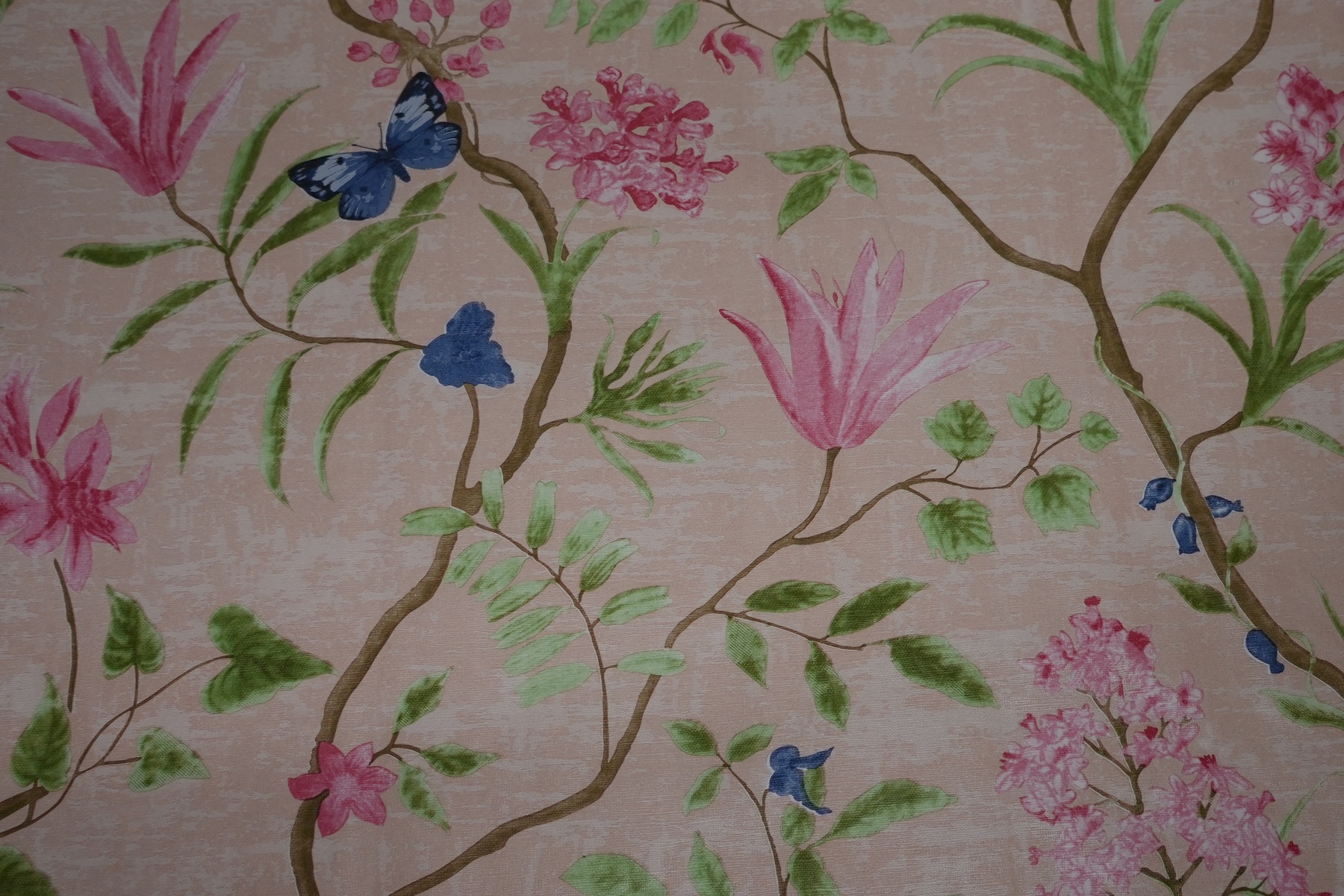}{5}%

        \noindent
        \objimg{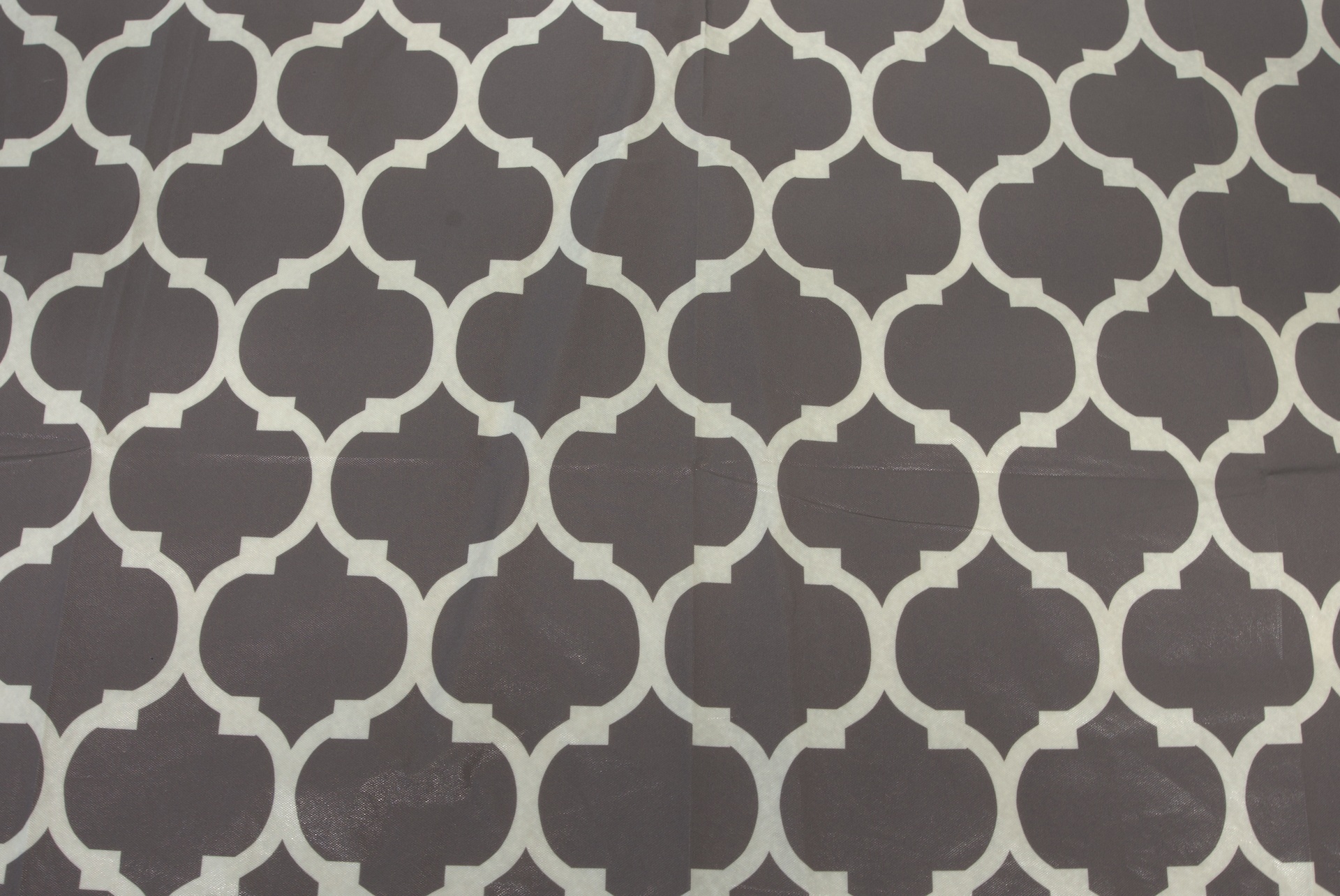}{6}\hspace{\gapH}%
        \objimg{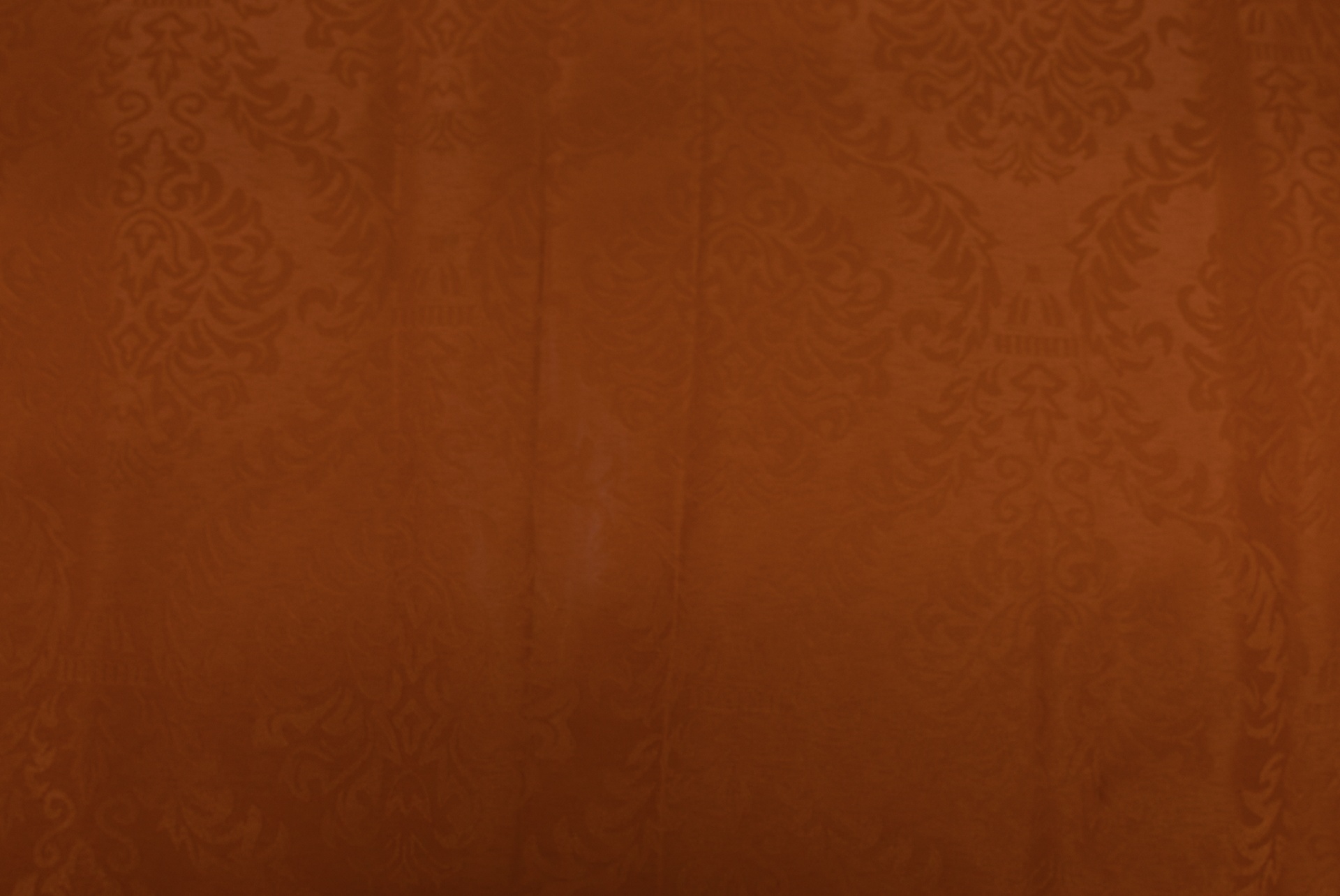}{7}\hspace{\gapH}%
        \objimg{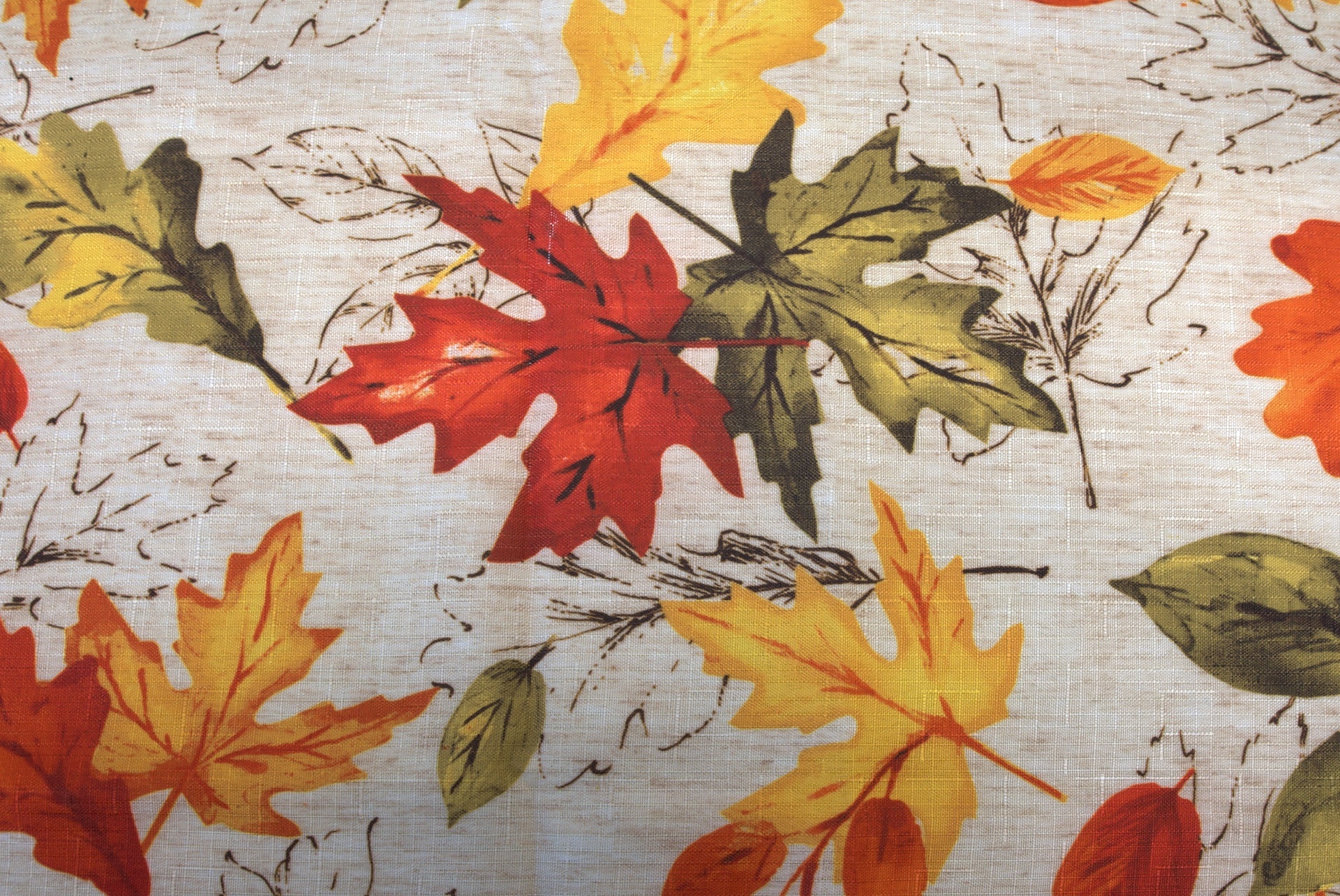}{8}\hspace{\gapH}%
        \objimg{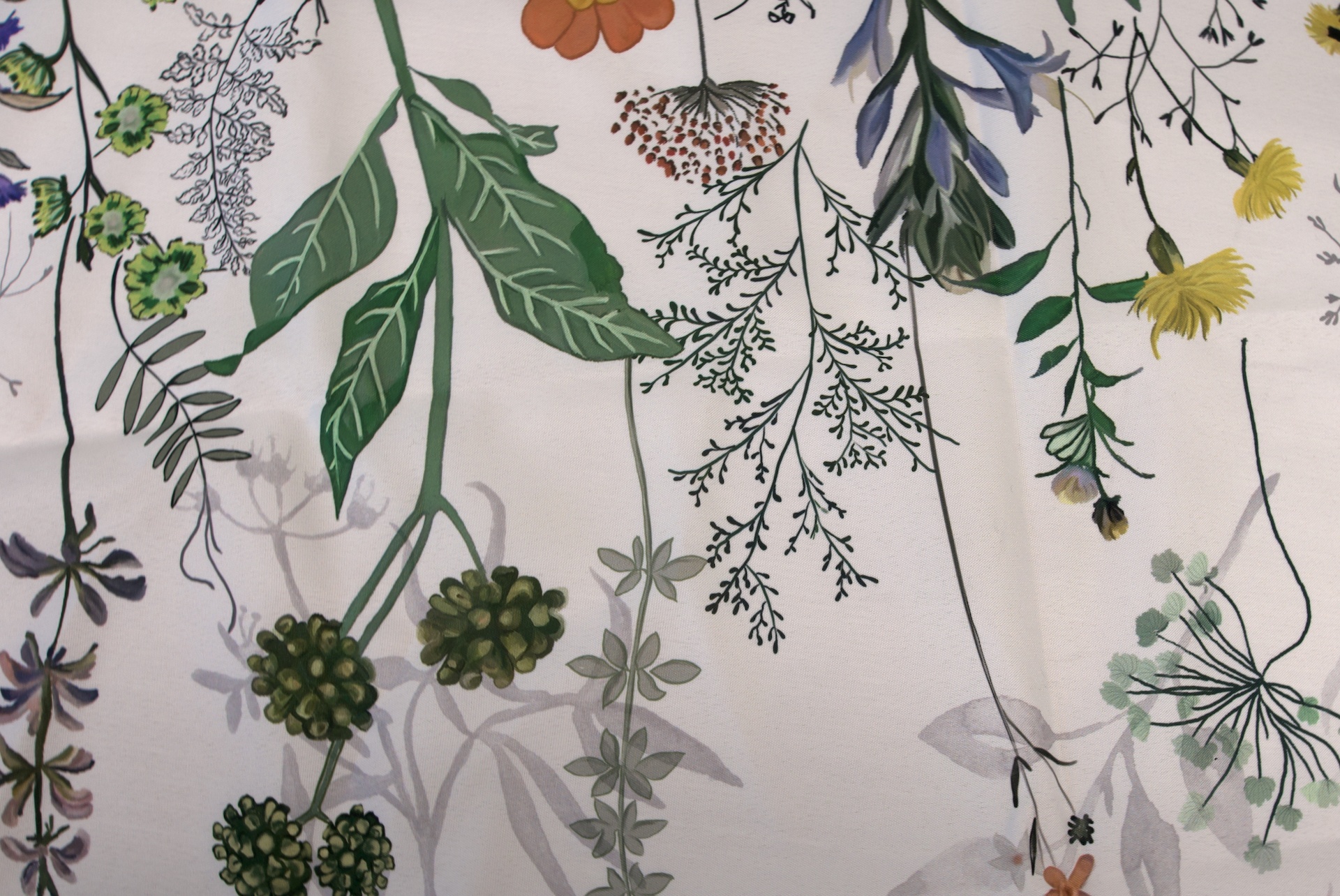}{9}\hspace{\gapH}%
        \objimg{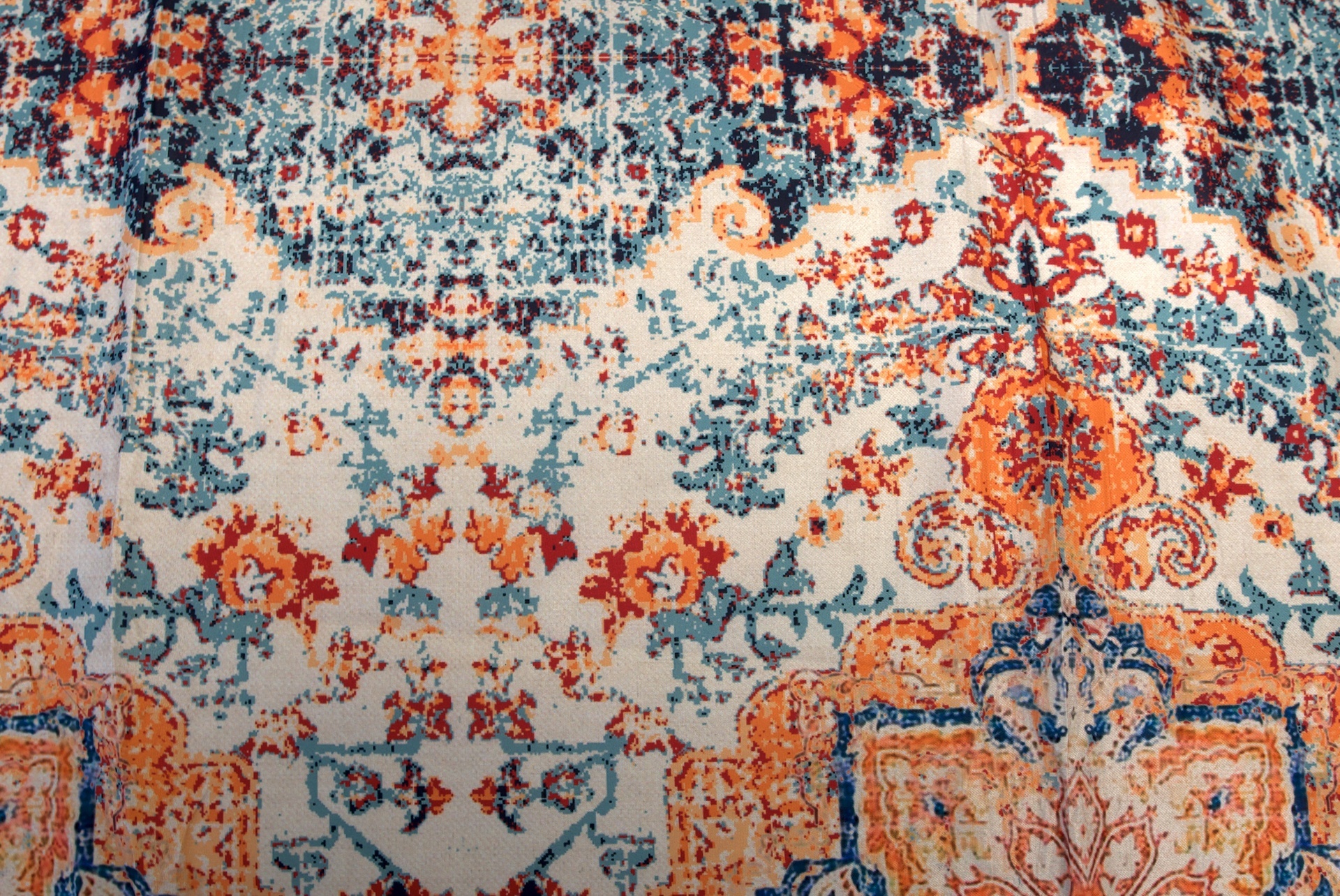}{10}%

        \par\vspace{\gapV}%
        \noindent
        \objimg{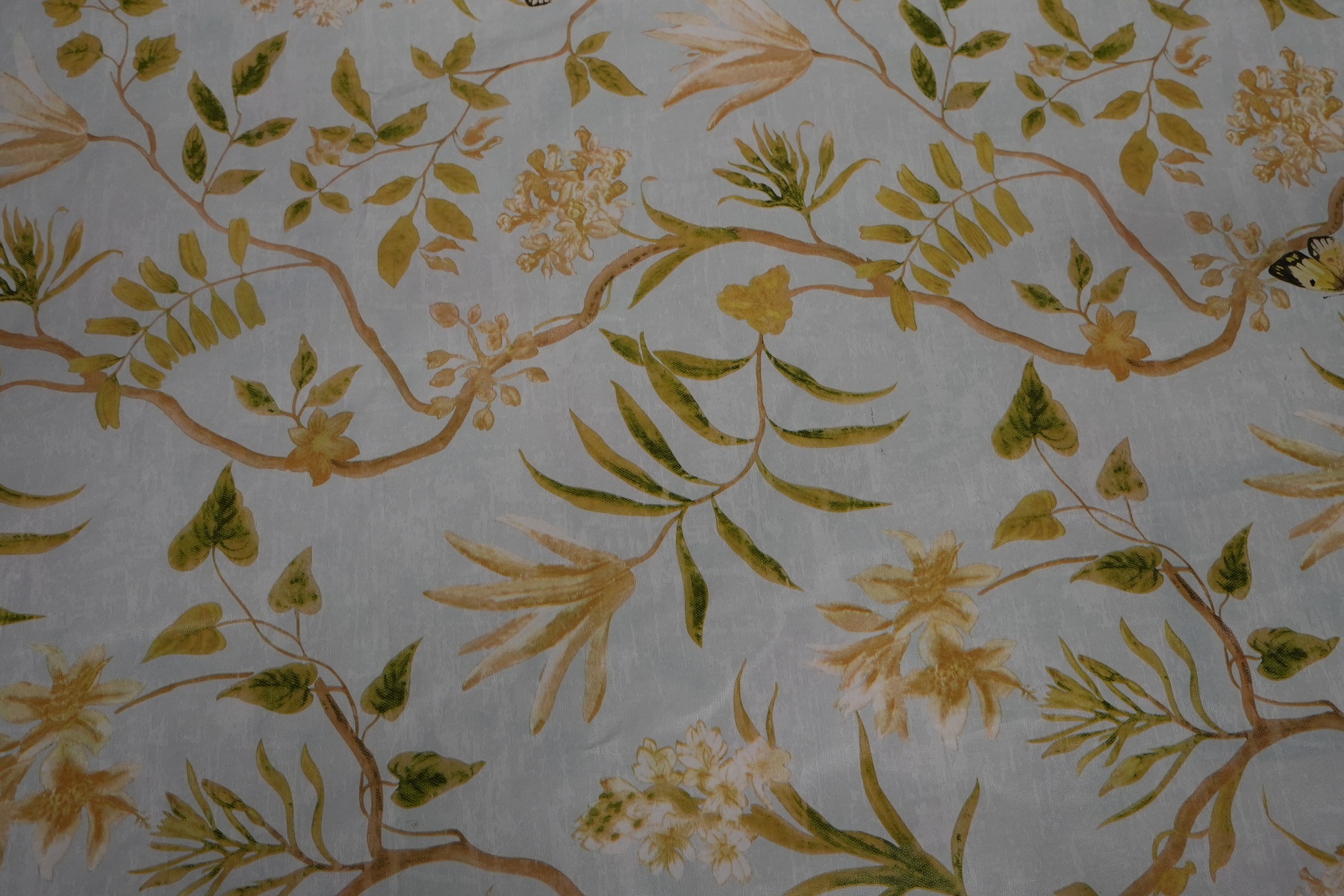}{11}\hspace{\gapH}%
        \objimg{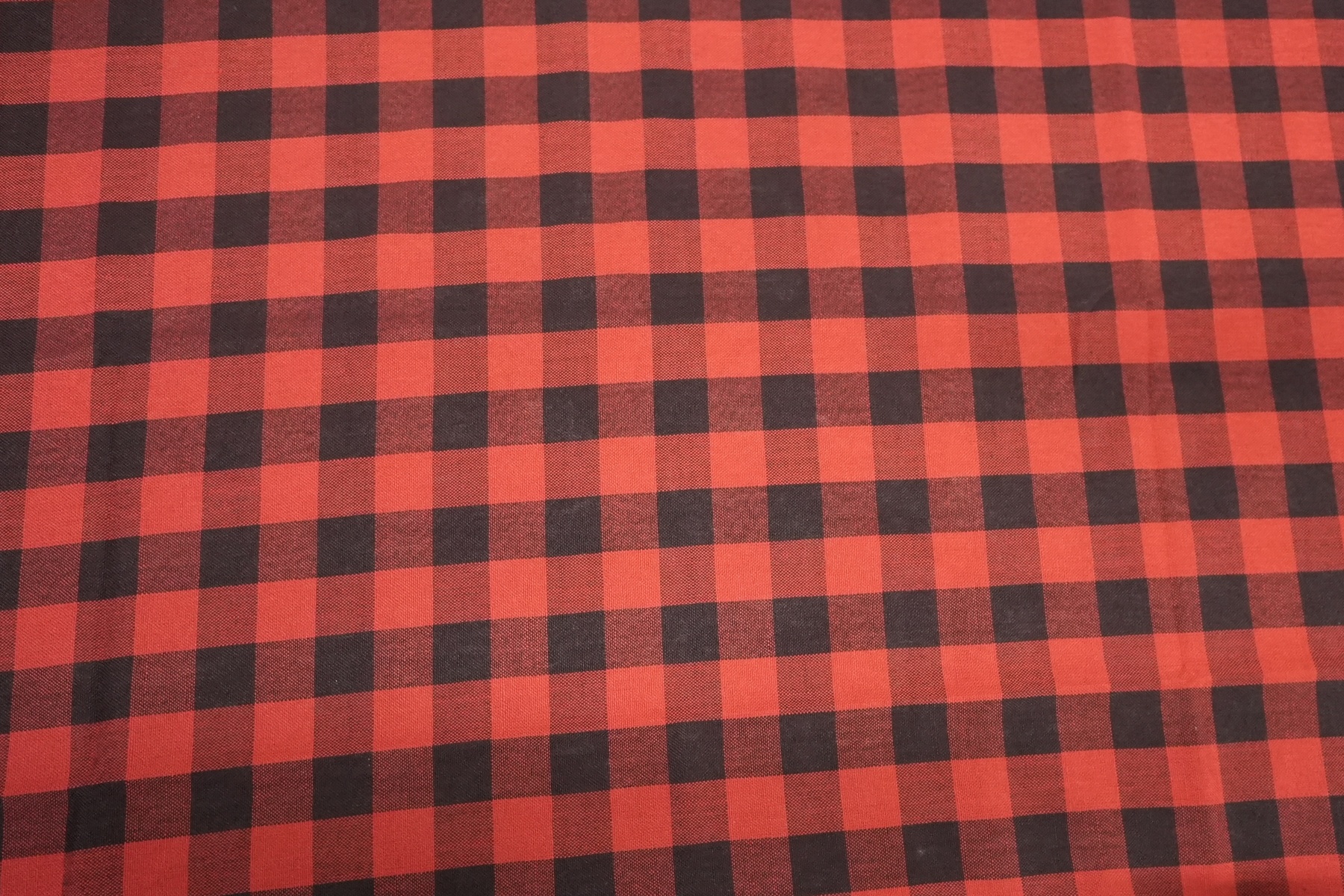}{12}%
      \end{minipage}%
    }

    \caption{Tablecloths used during training/testing (labelled 1--12). 1-4, 6-10 were used in training and 12 was used in testing for all six tasks. \textbf{Pick}, \textbf{Stack}, \textbf{Reorient} used 5 in training and 11 in testing, while \textbf{Pull}, \textbf{Book}, \textbf{Pour} used 11 in training and 5 in testing.}
    \label{fig:tablecloths-train-test}
\end{figure}

\begin{figure}[!t]
    \setlength{\imgw}{0.18\textwidth}
    \setlength{\gapH}{2pt}
    \setlength{\gapV}{1pt}
    \setlength{\totw}{5\imgw}
    \addtolength{\totw}{4\gapH}

    \fbox{%
      \begin{minipage}{\totw}
        \setlength{\parskip}{0pt}

        \noindent
        \objimg{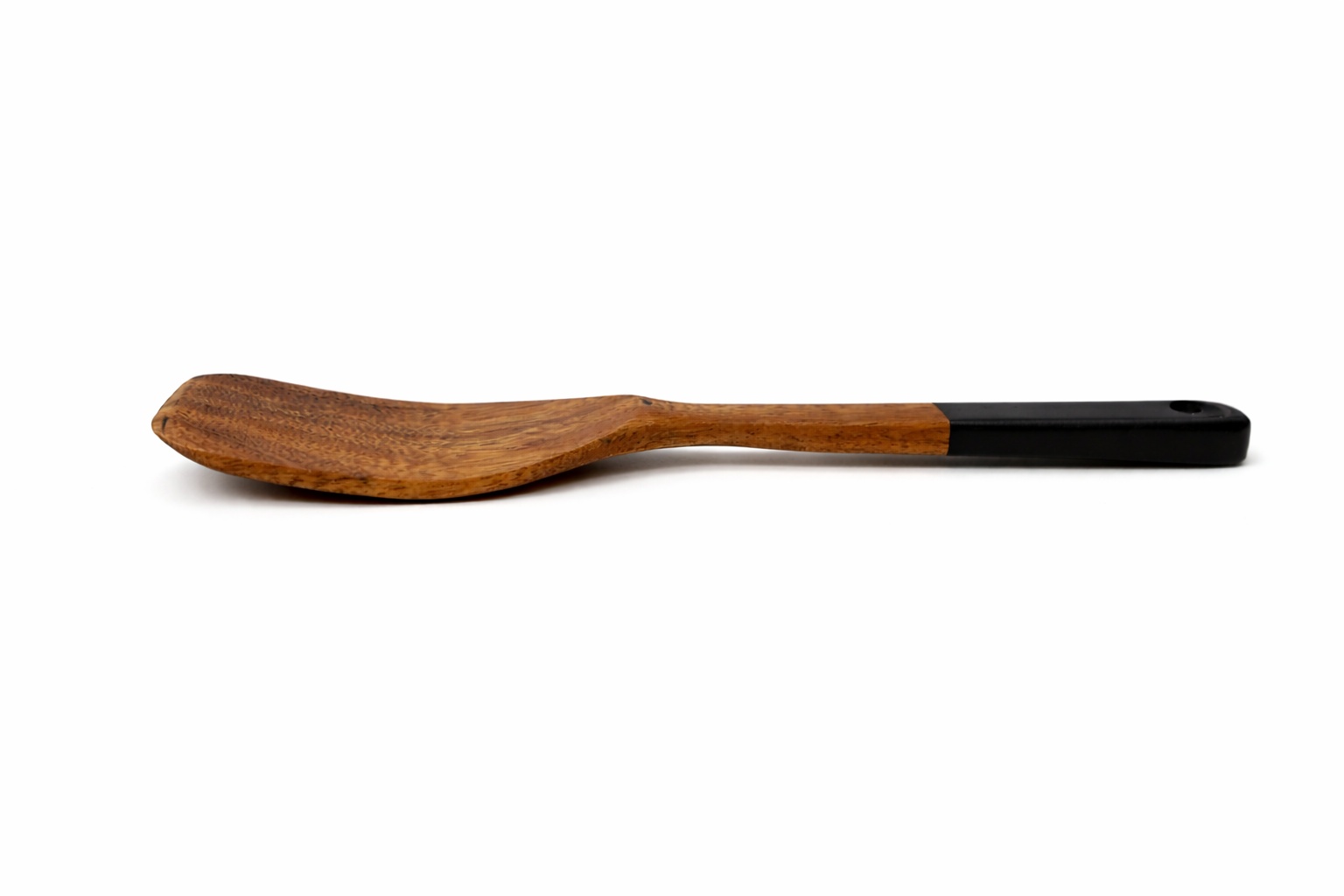}{1}\hspace{\gapH}%
        \objimg{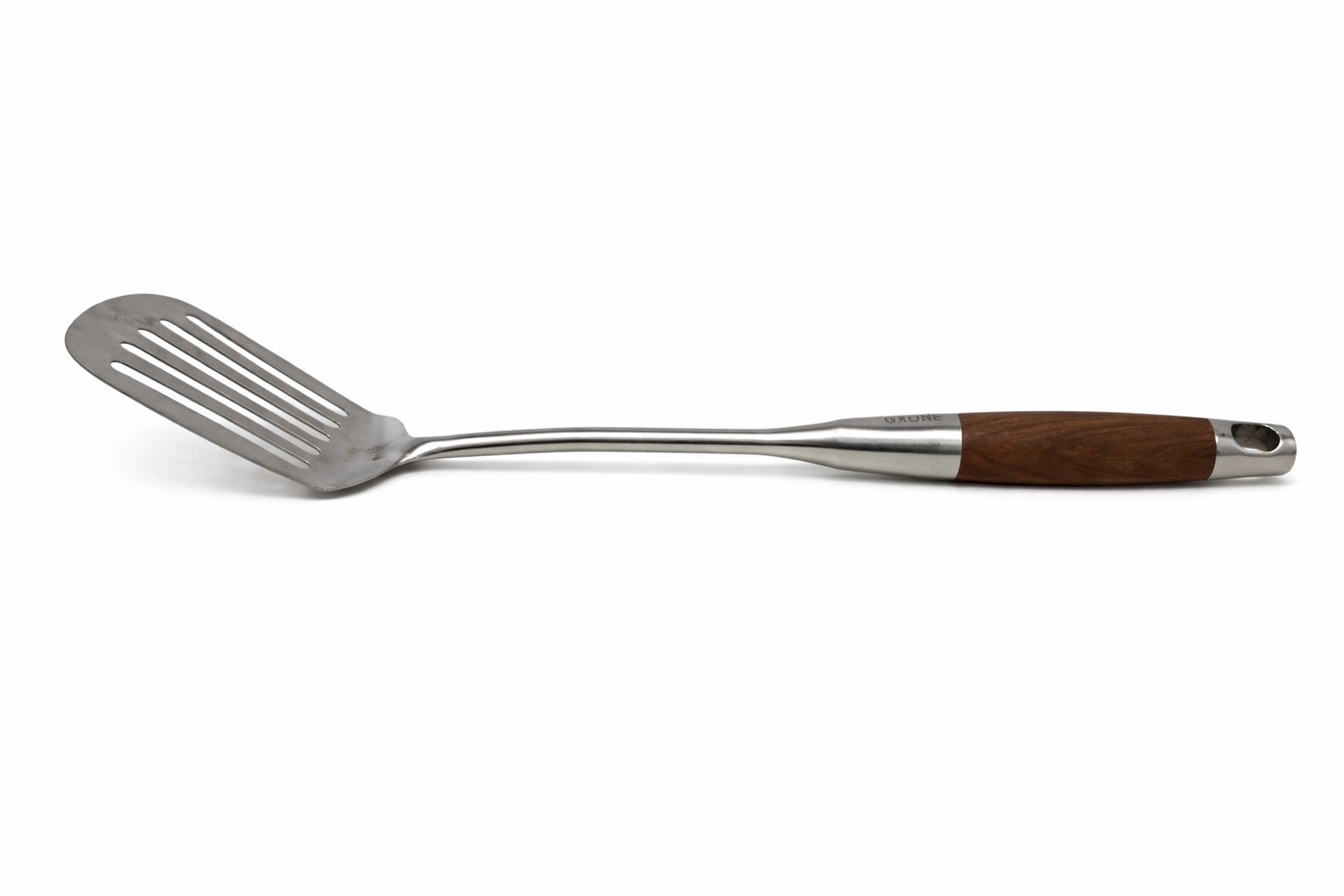}{2}\hspace{\gapH}%
        \objimg{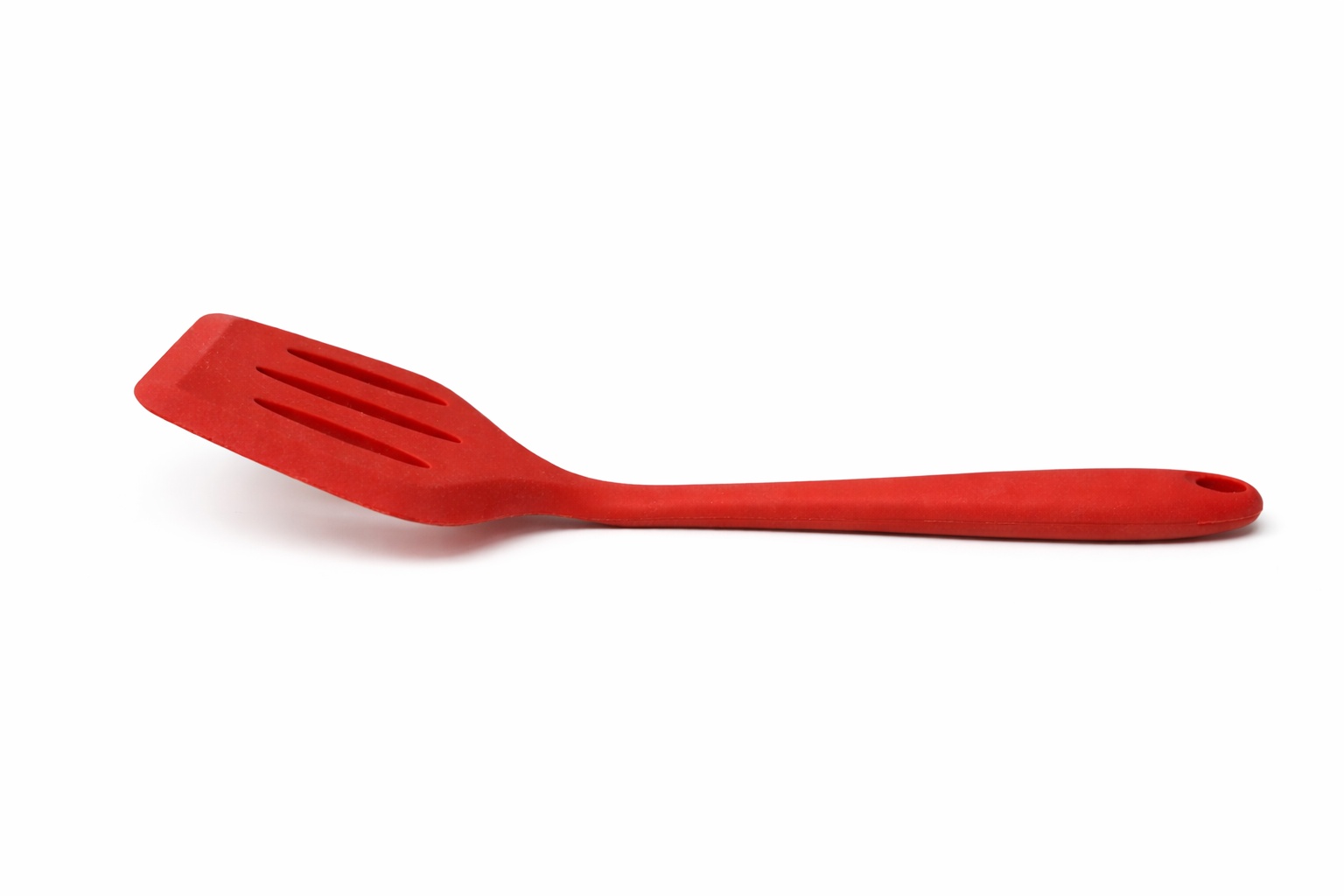}{3}\hspace{\gapH}%
        \objimg{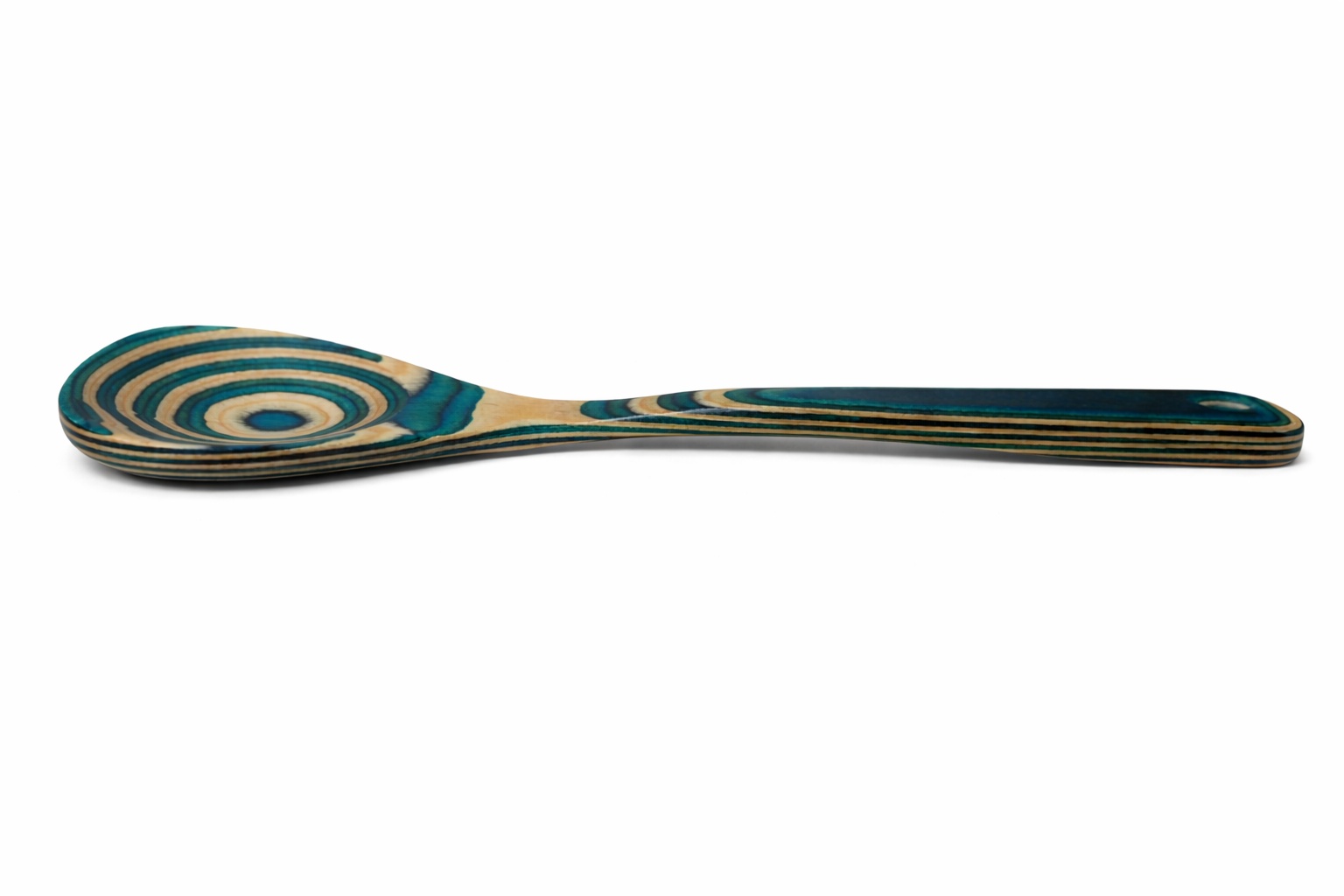}{4}\hspace{\gapH}%
        \objimg{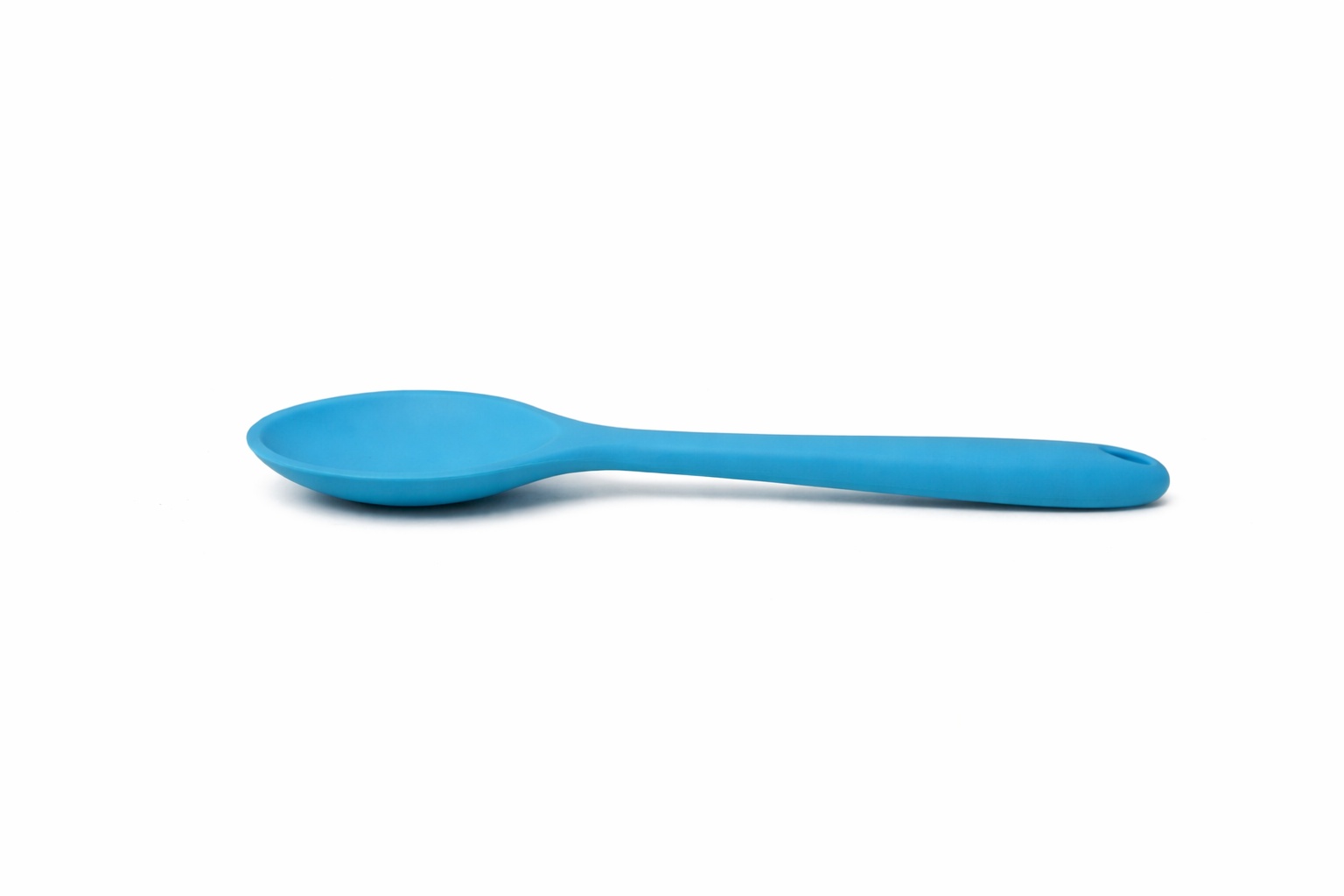}{5}%

        \noindent
        \objimg{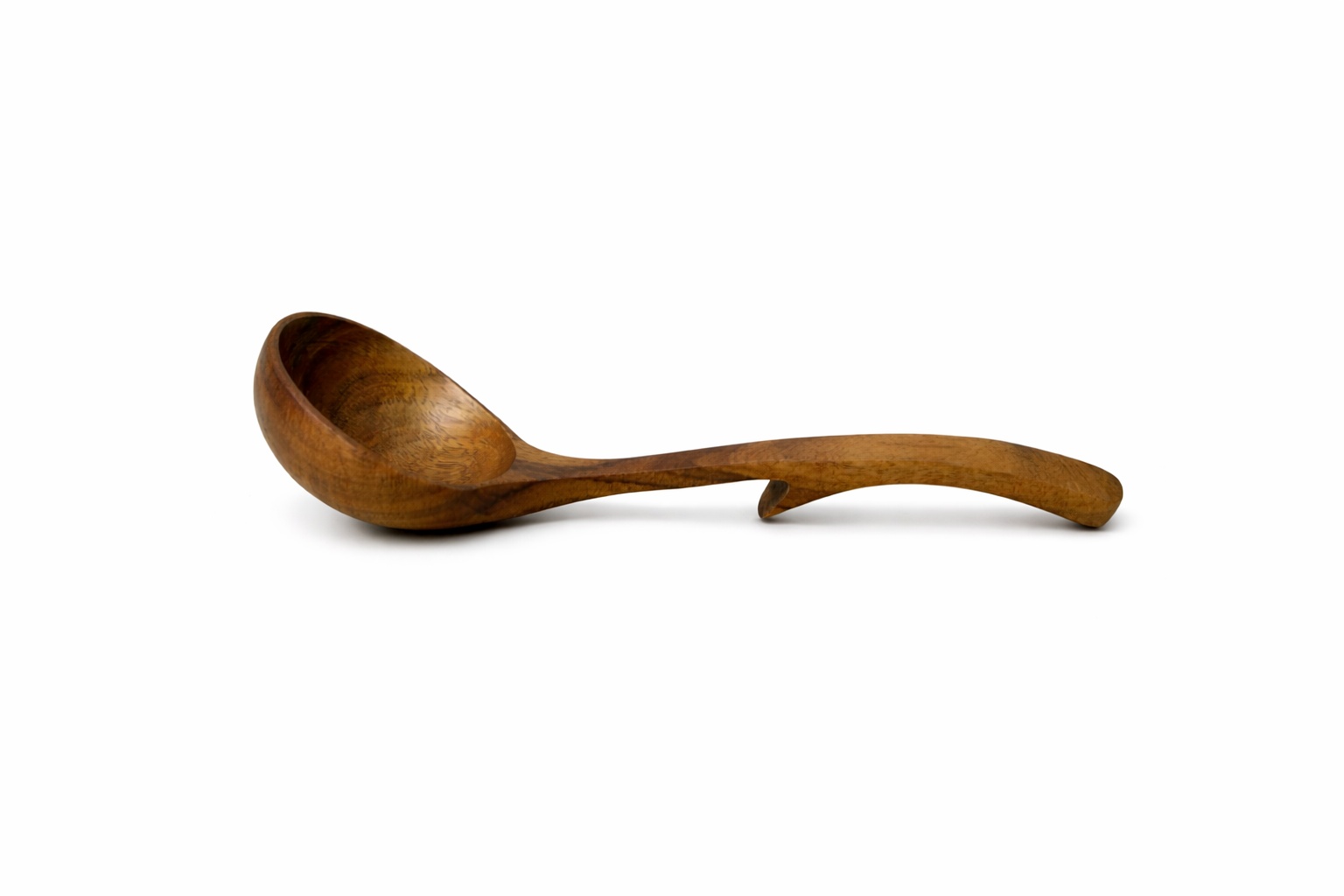}{6}\hspace{\gapH}%
        \objimg{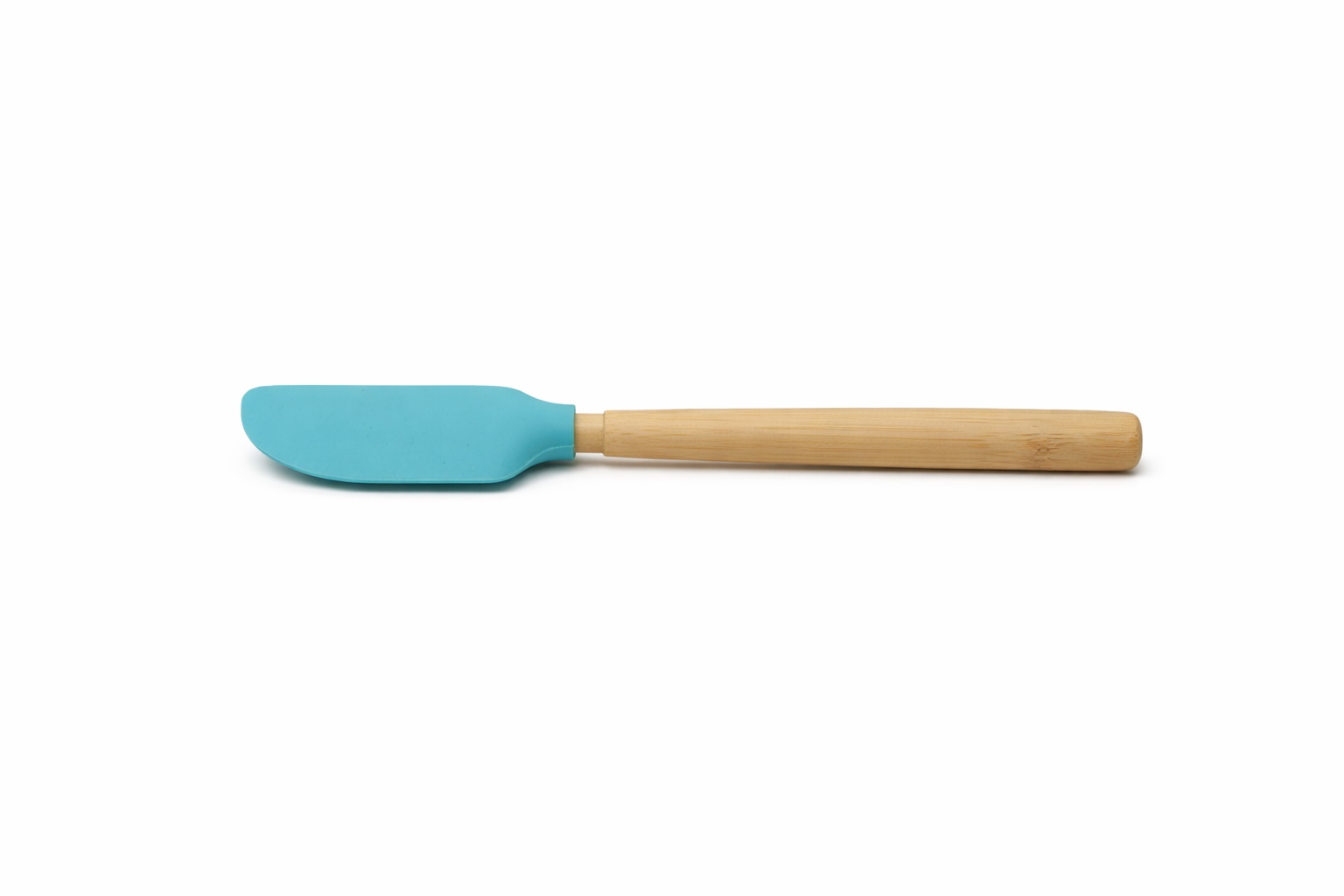}{7}\hspace{\gapH}%
        \objimg{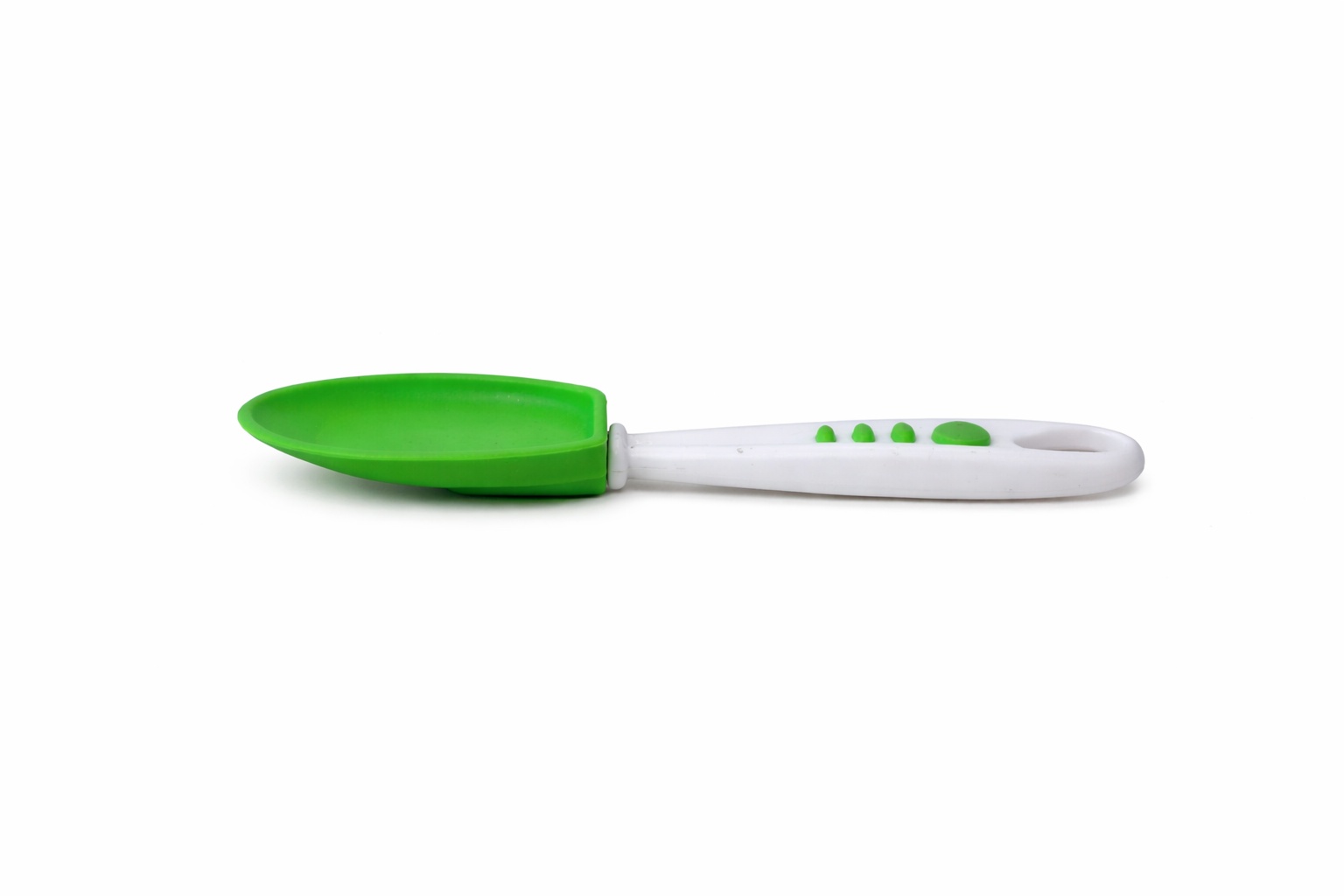}{8}\hspace{\gapH}%
        \objimg{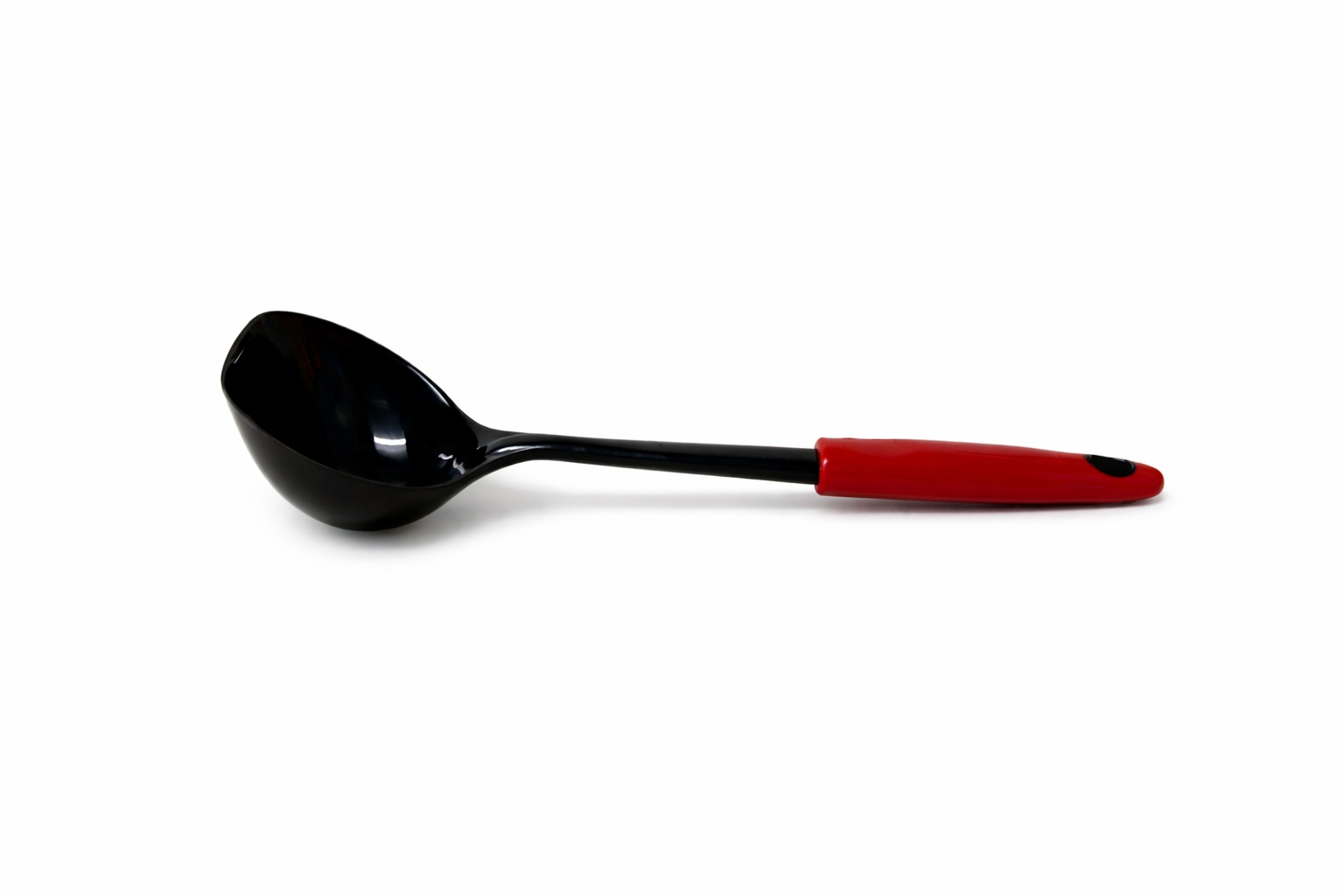}{9}\hspace{\gapH}%
        \objimg{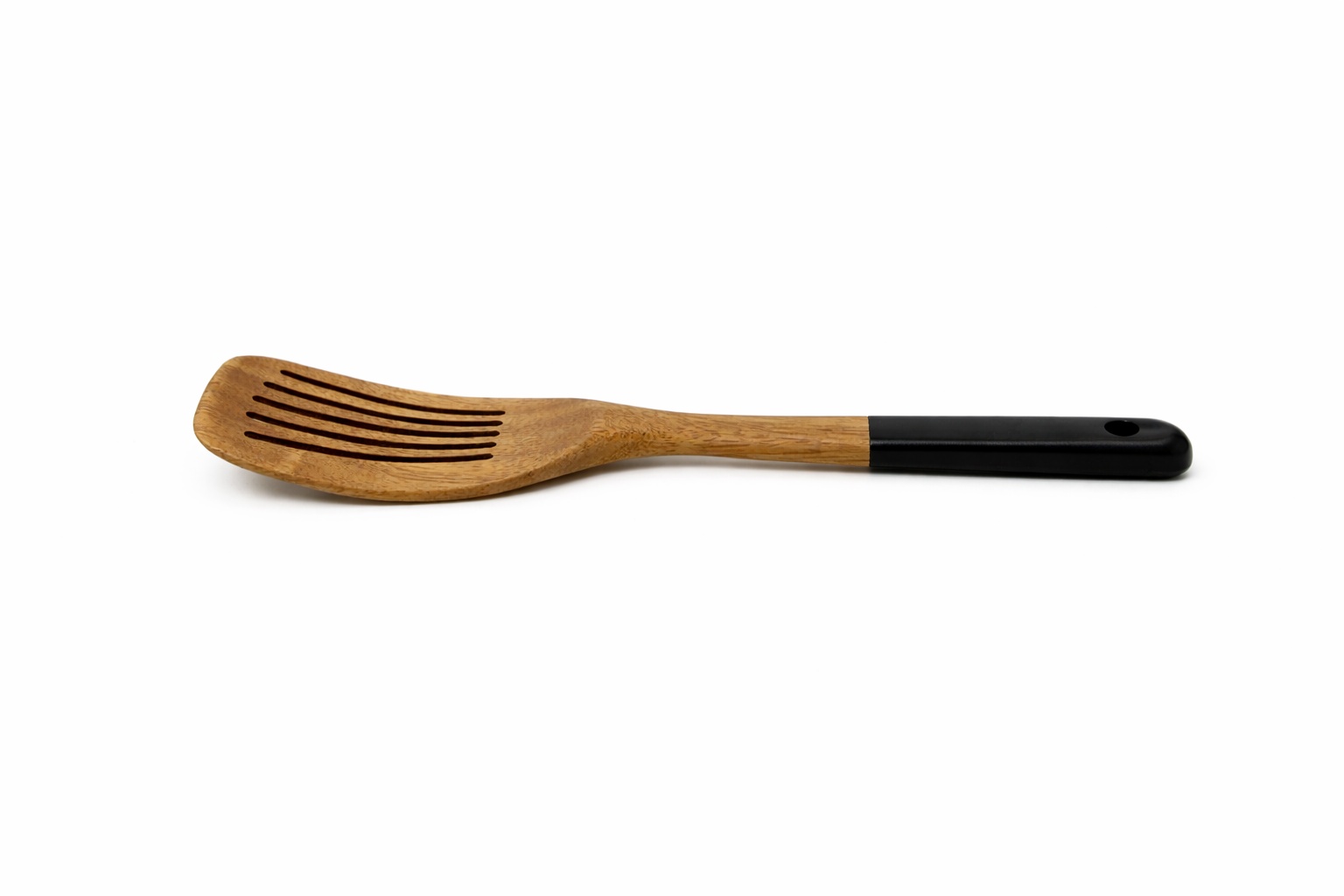}{10}%

        \par\vspace{\gapV}%
        \noindent
        \objimg{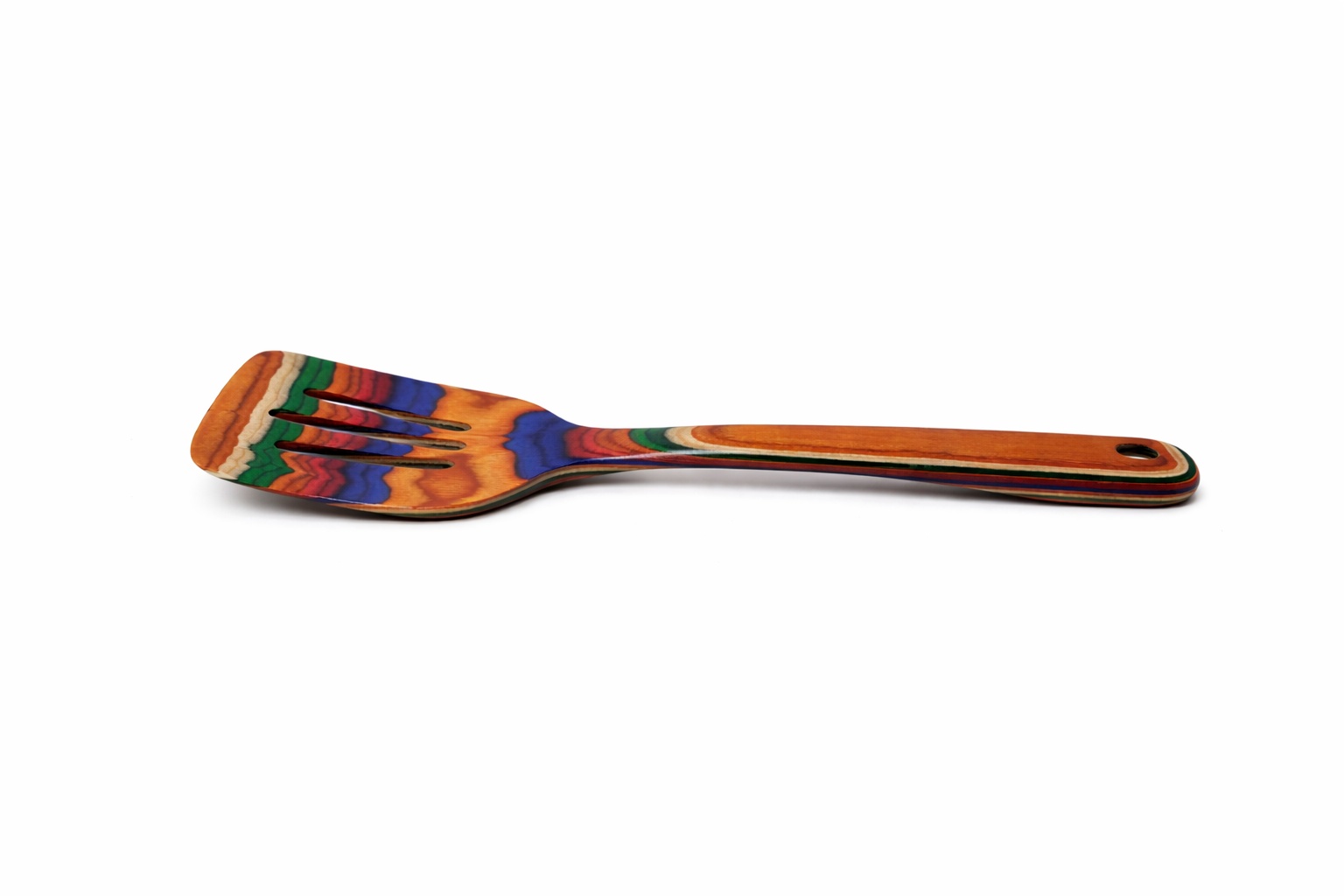}{11}\hspace{\gapH}%
        \objimg{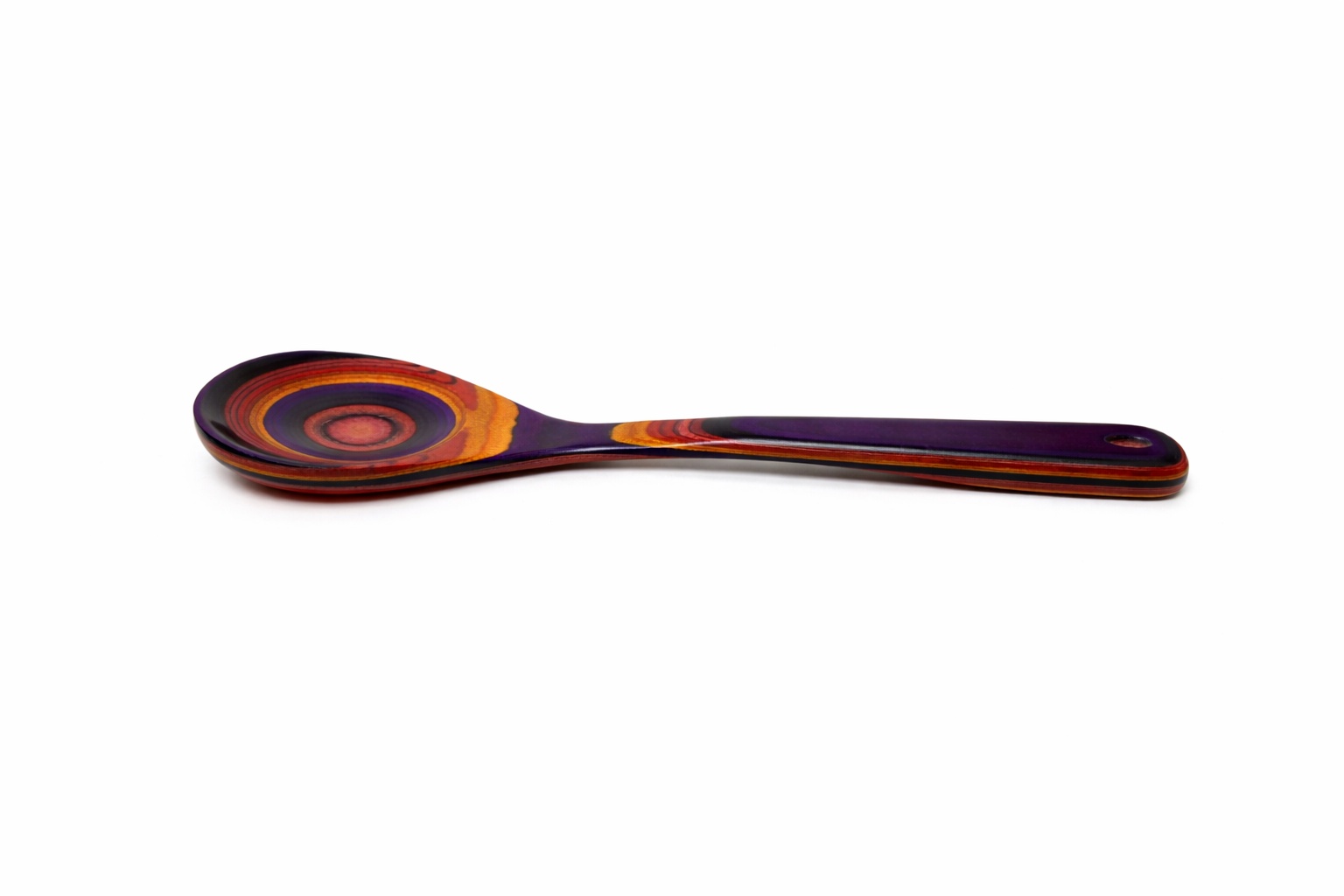}{12}%
      \end{minipage}%
    }

    \caption{Spatulas used during training/testing (labelled 1-12). 1-10 were used in training and 11-12 were used in testing for \textbf{Pick}.}
    \label{fig:spatulas-train-test}
\end{figure}

\begin{figure}[!t]
    \setlength{\imgw}{0.18\textwidth}
    \setlength{\gapH}{2pt}
    \setlength{\gapV}{1pt}
    \setlength{\totw}{5\imgw}
    \addtolength{\totw}{4\gapH}

    \fbox{%
      \begin{minipage}{\totw}
        \setlength{\parskip}{0pt}

        \noindent
        \objimg{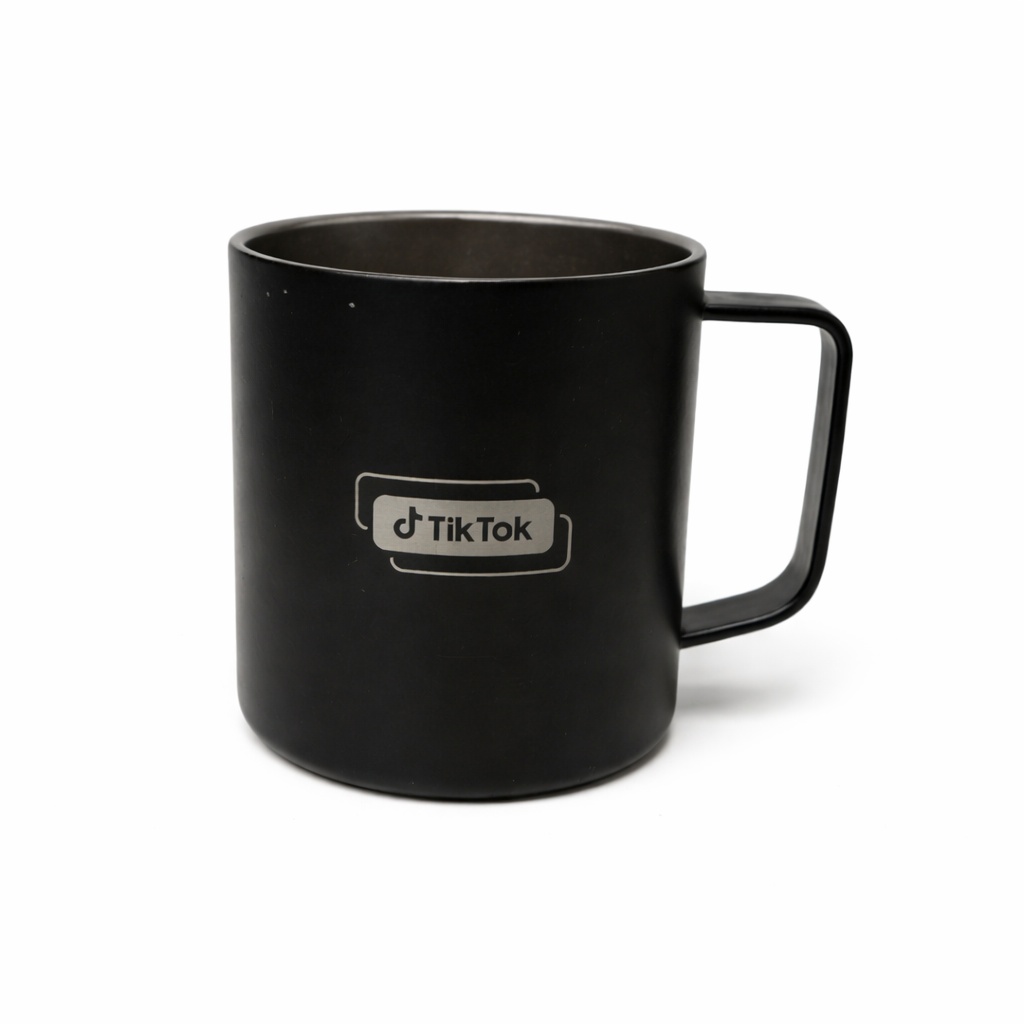}{1}\hspace{\gapH}%
        \objimg{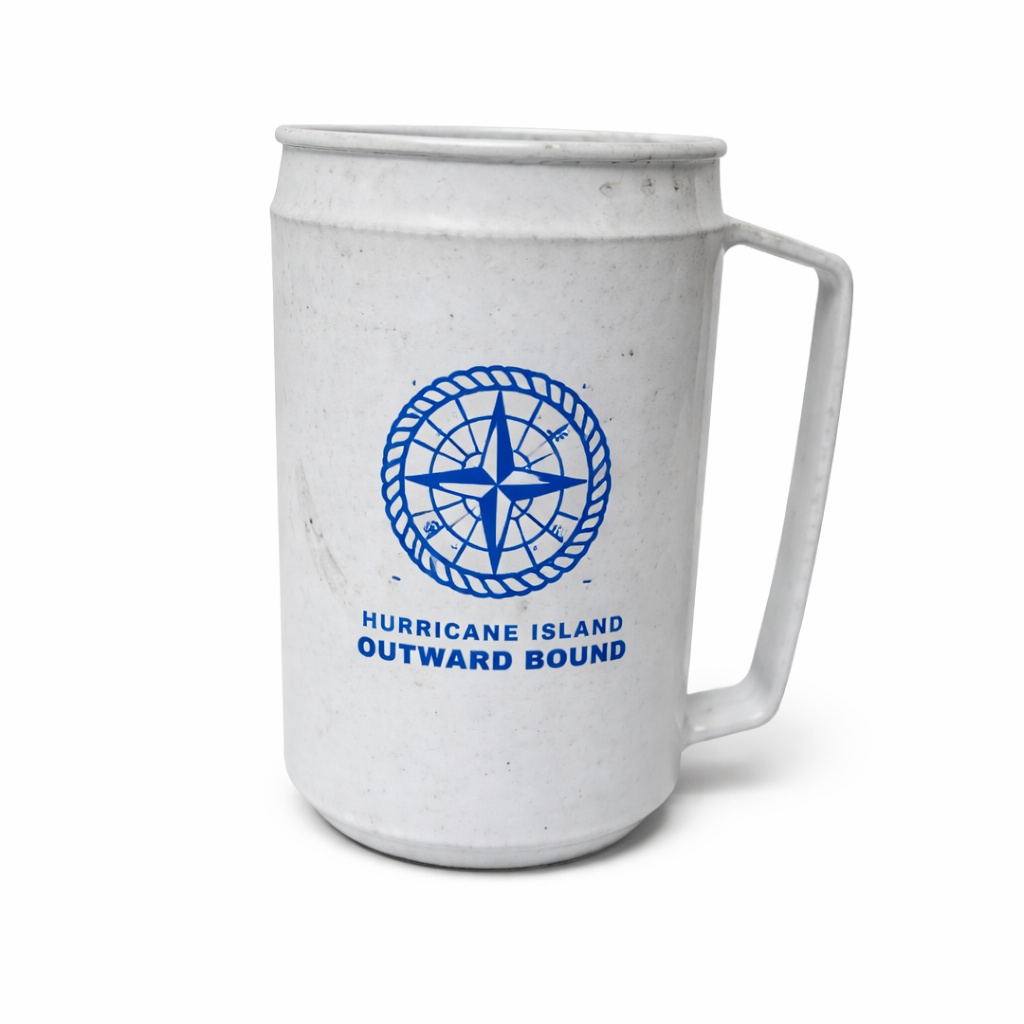}{2}\hspace{\gapH}%
        \objimg{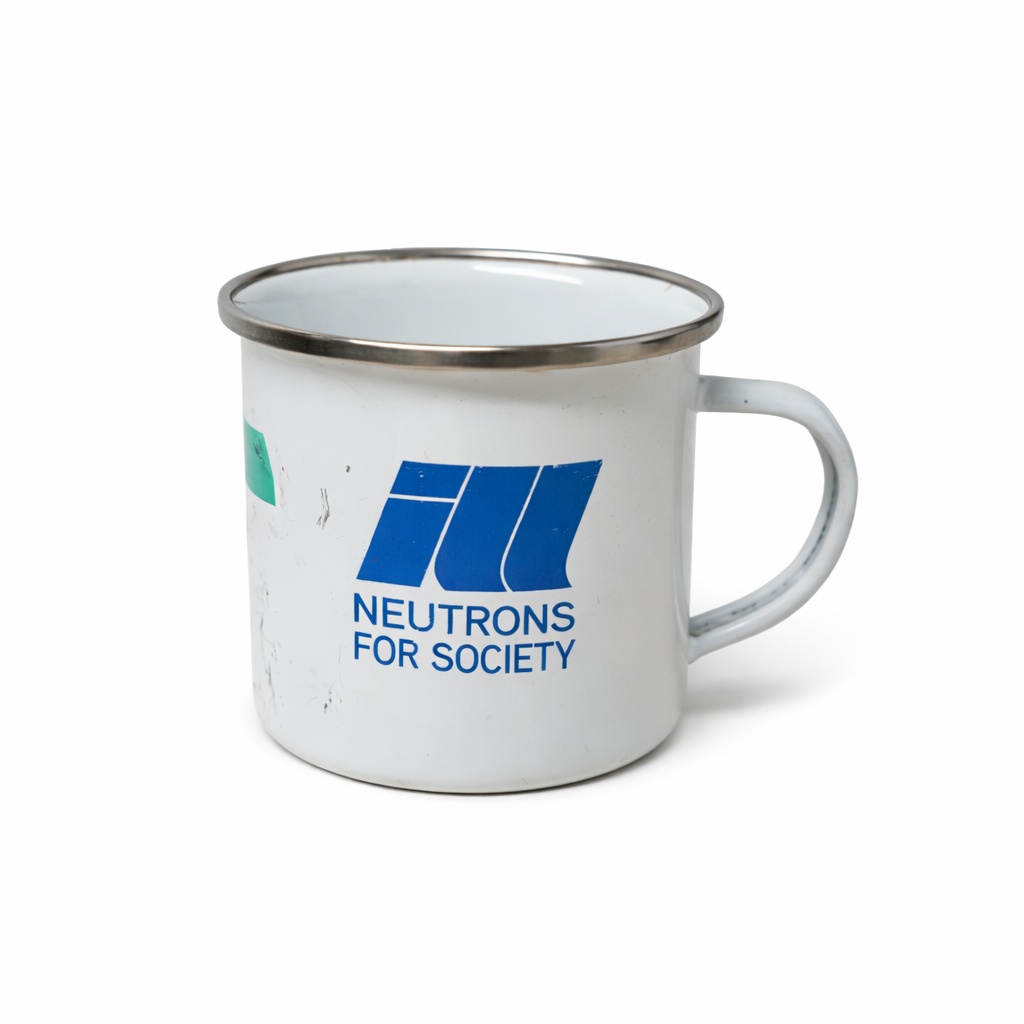}{3}\hspace{\gapH}%
        \objimg{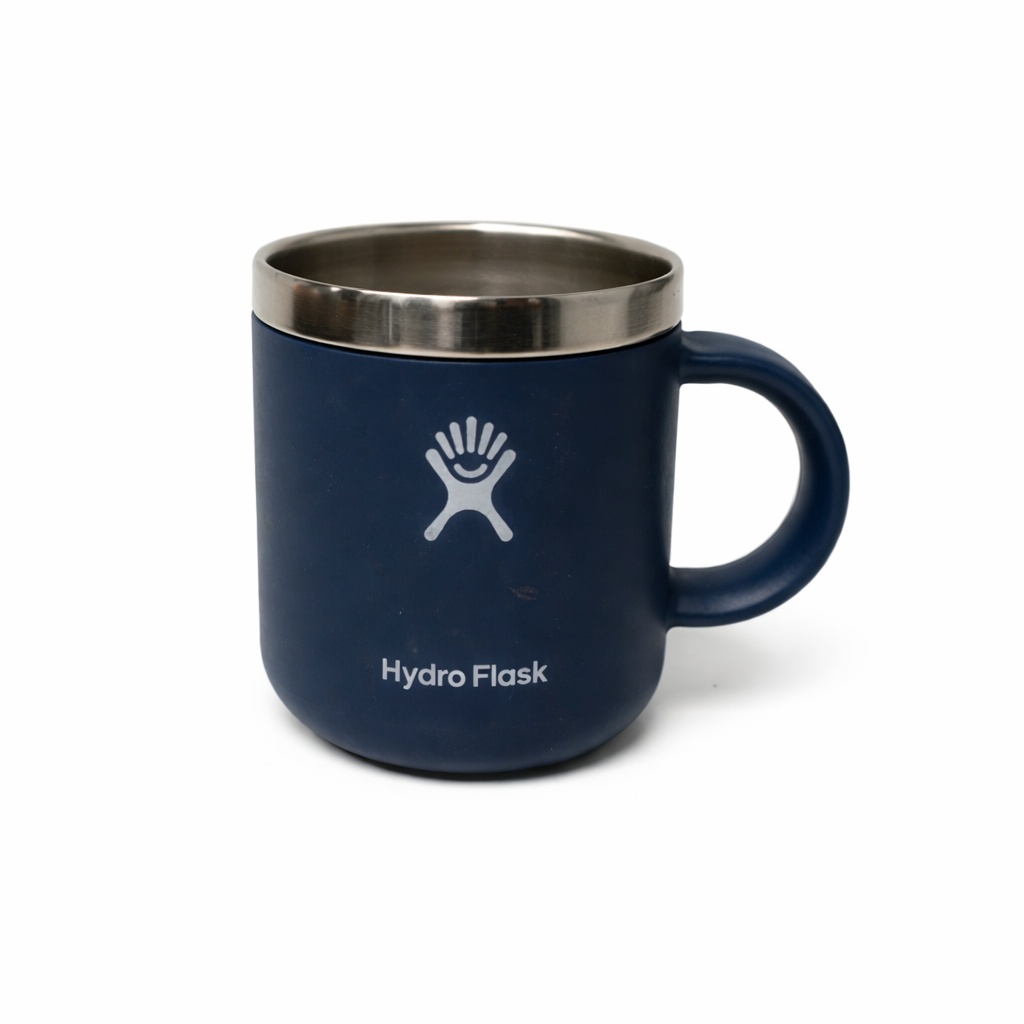}{4}\hspace{\gapH}%
        \objimg{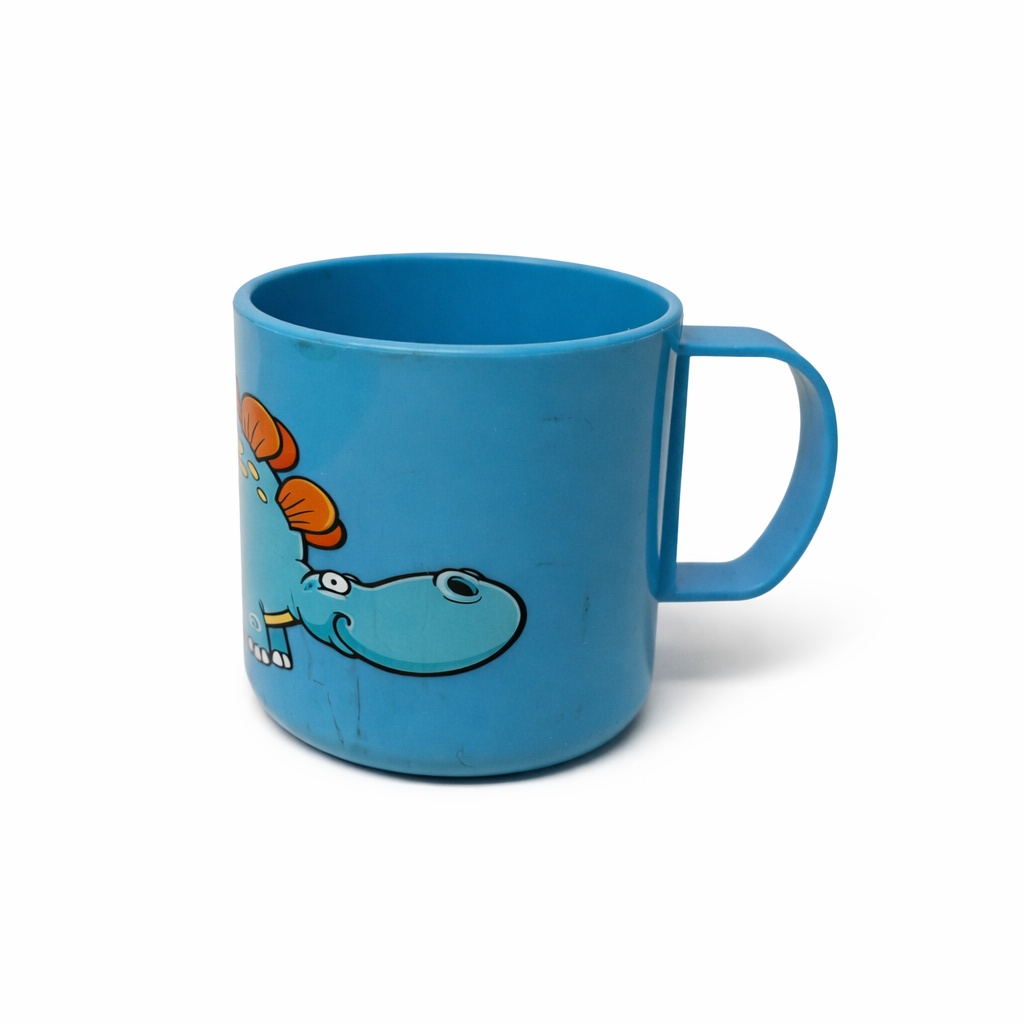}{5}%

        \noindent
        \objimg{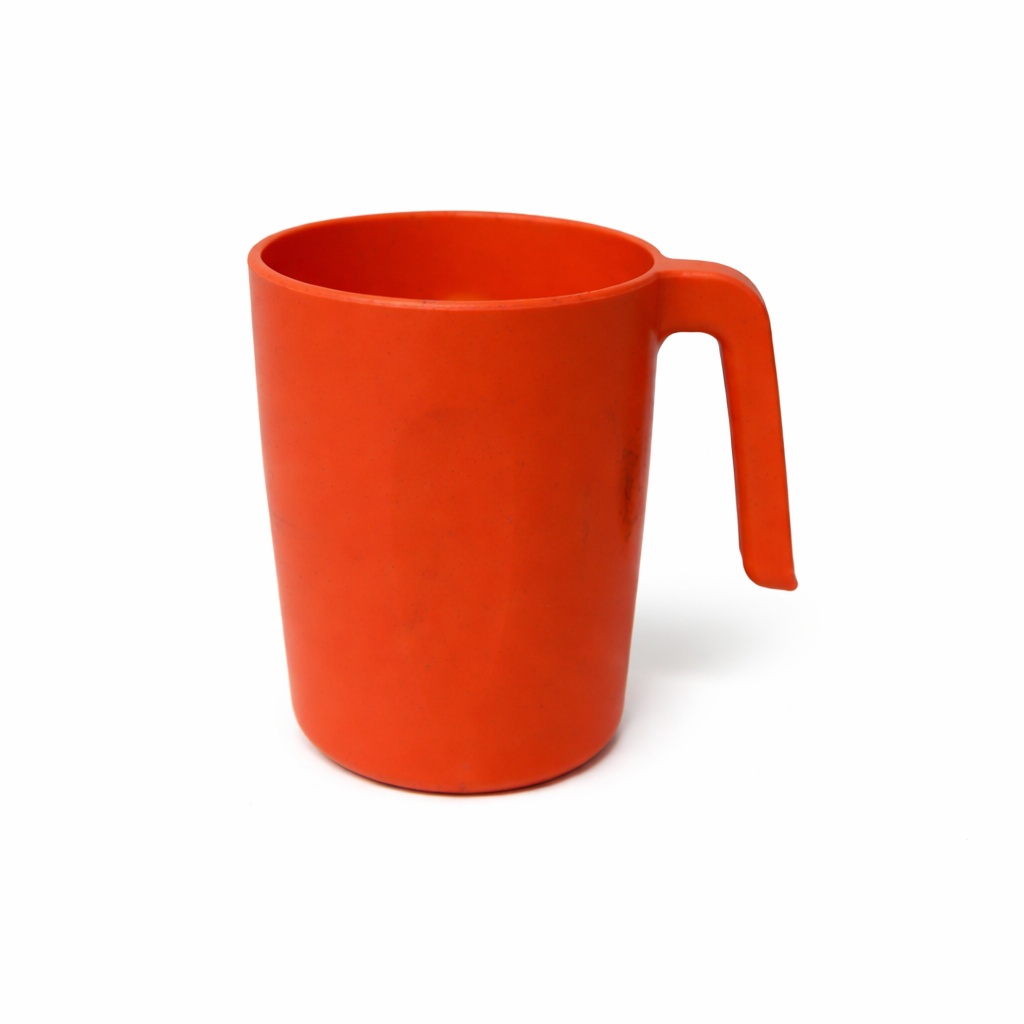}{6}\hspace{\gapH}%
        \objimg{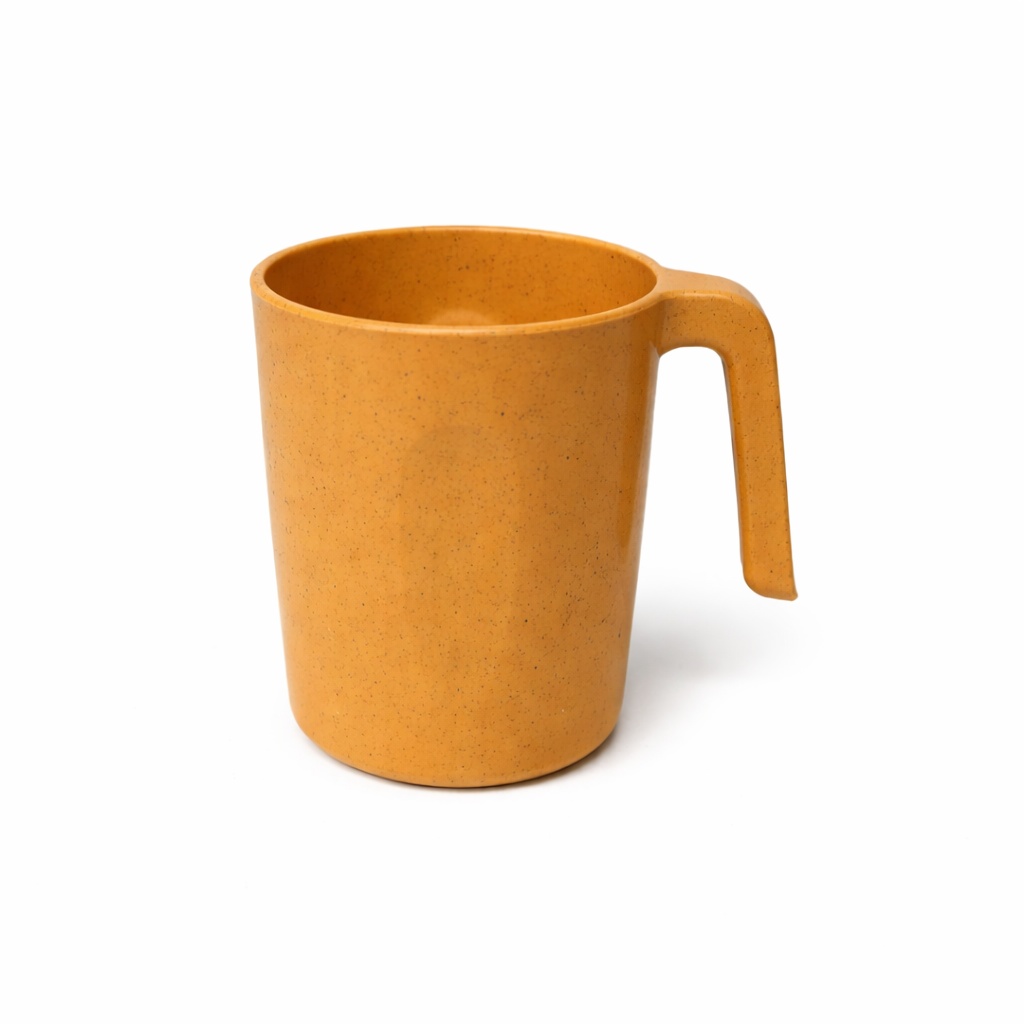}{7}\hspace{\gapH}%
        \objimg{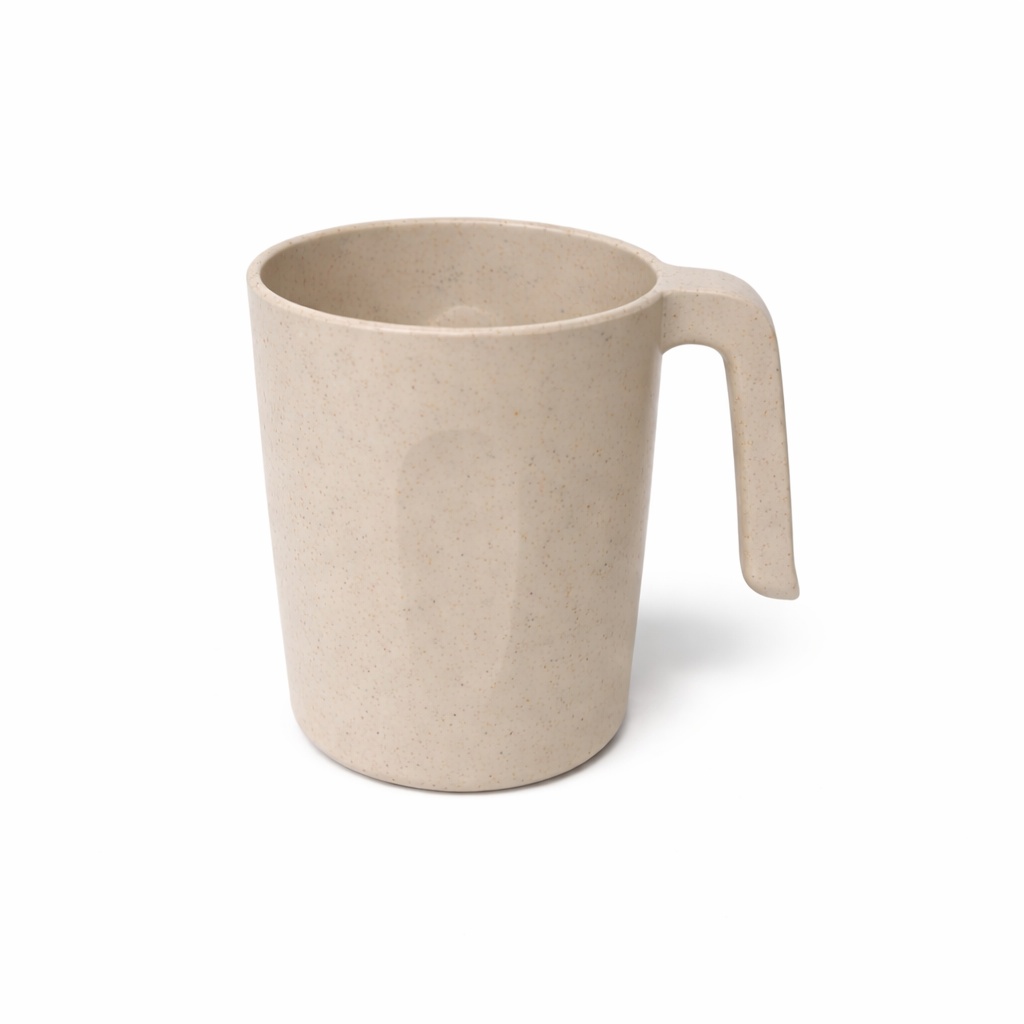}{8}\hspace{\gapH}%
        \objimg{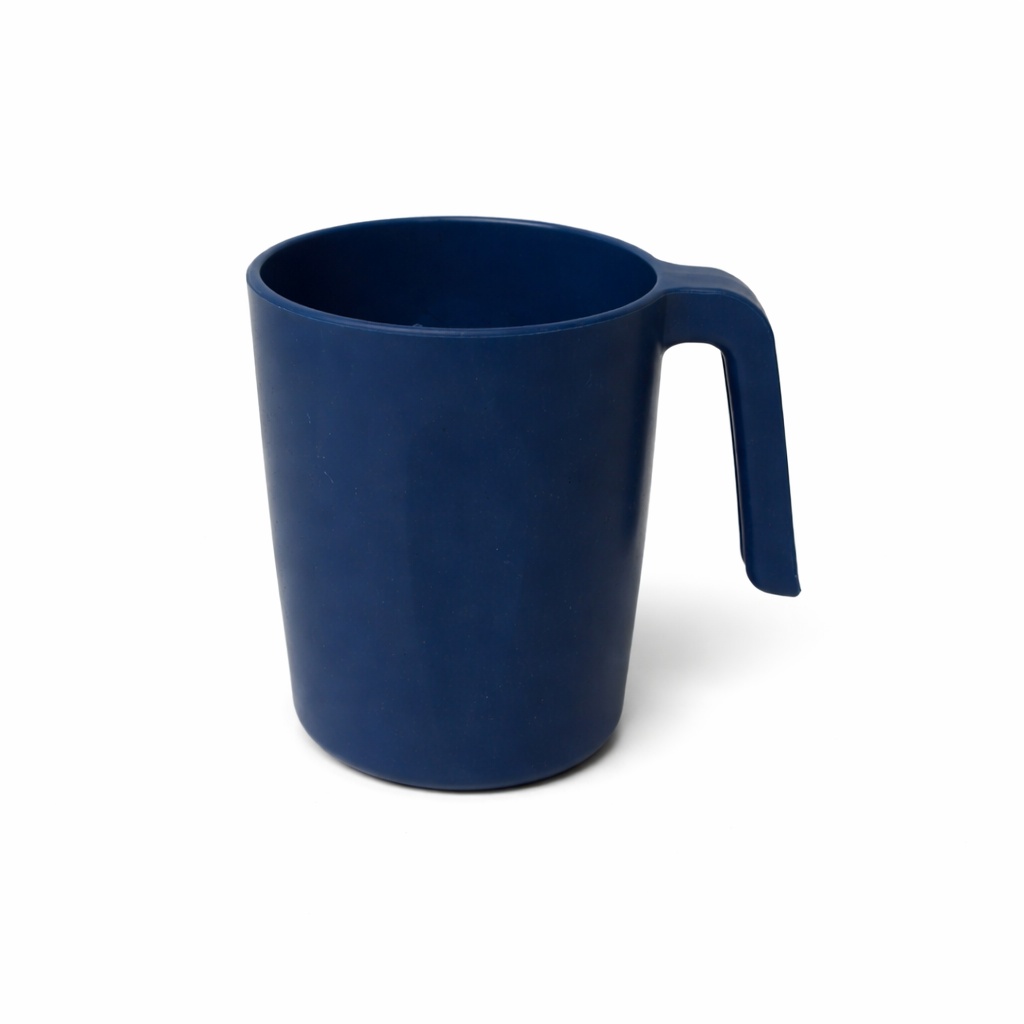}{9}\hspace{\gapH}%
        \objimg{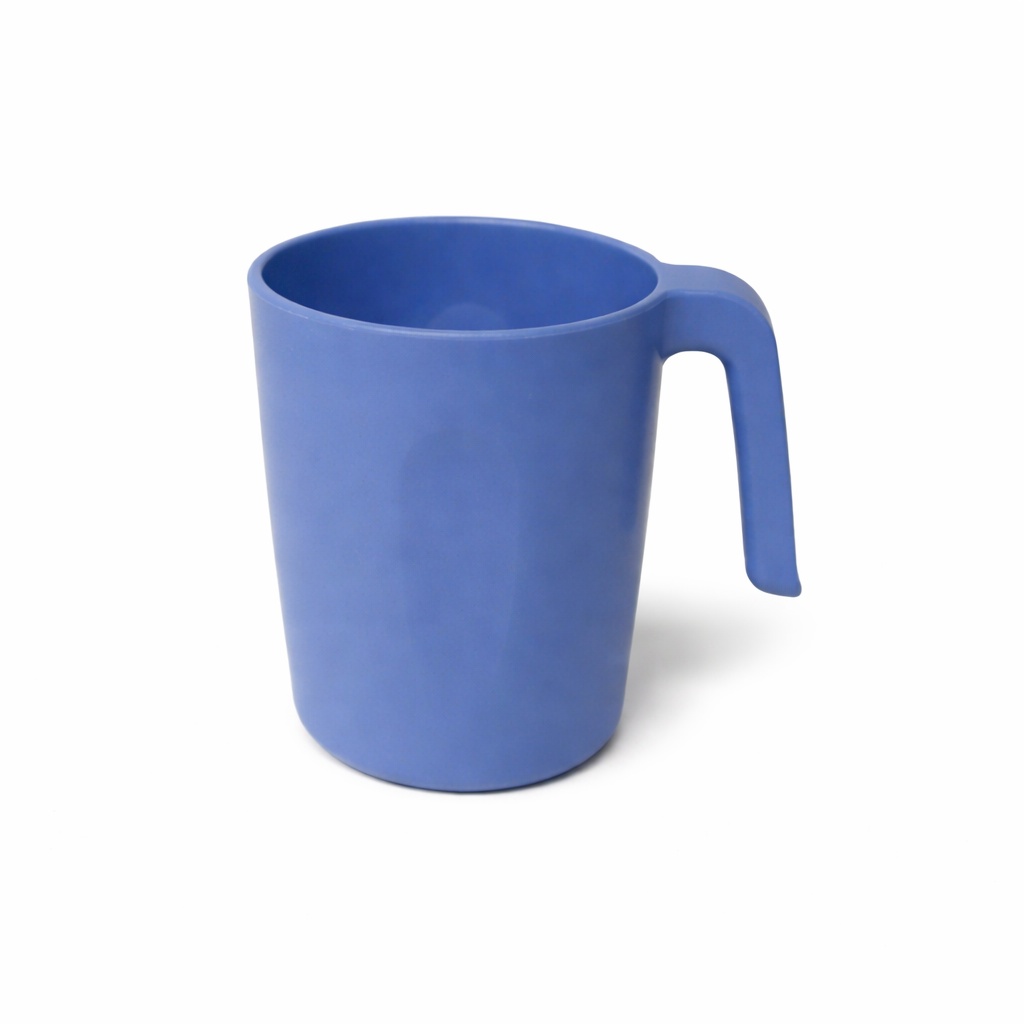}{10}%

        \par\vspace{\gapV}%
        \noindent
        \objimg{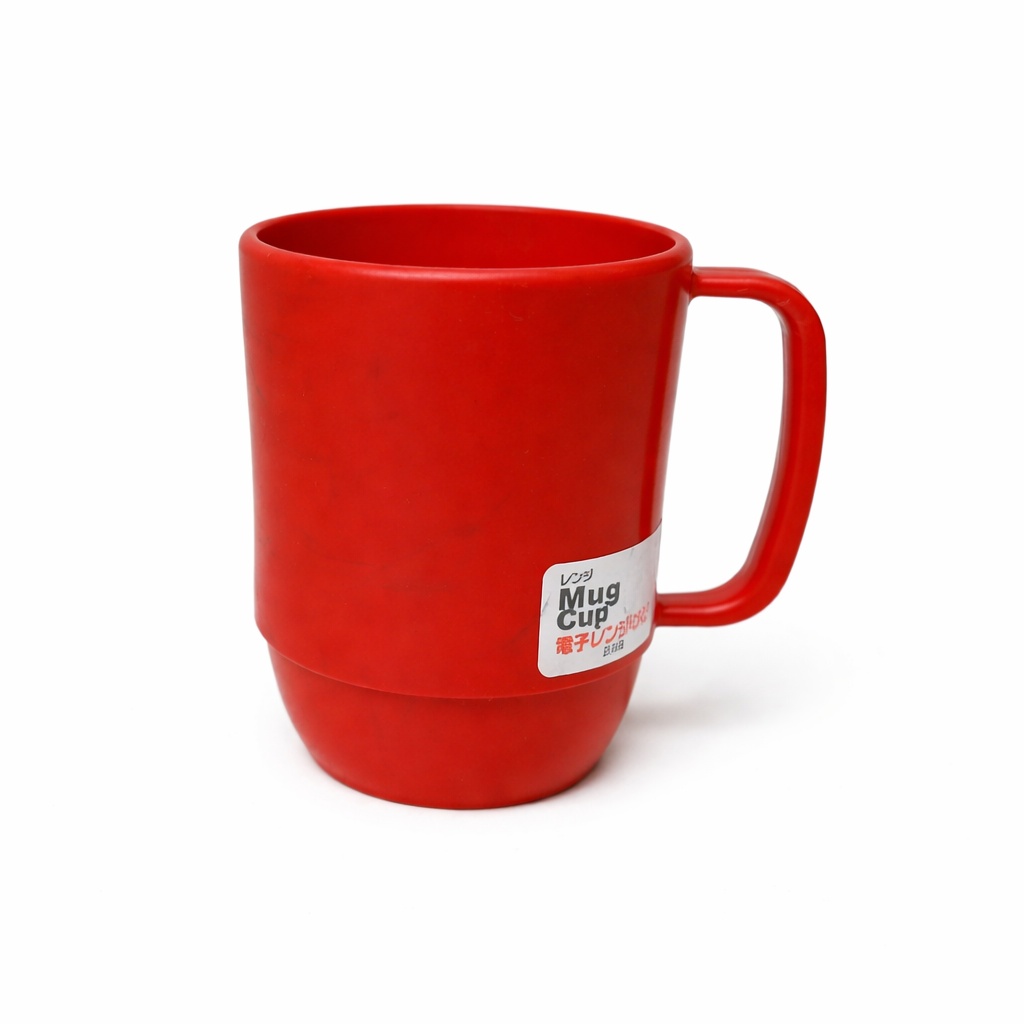}{11}\hspace{\gapH}%
        \objimg{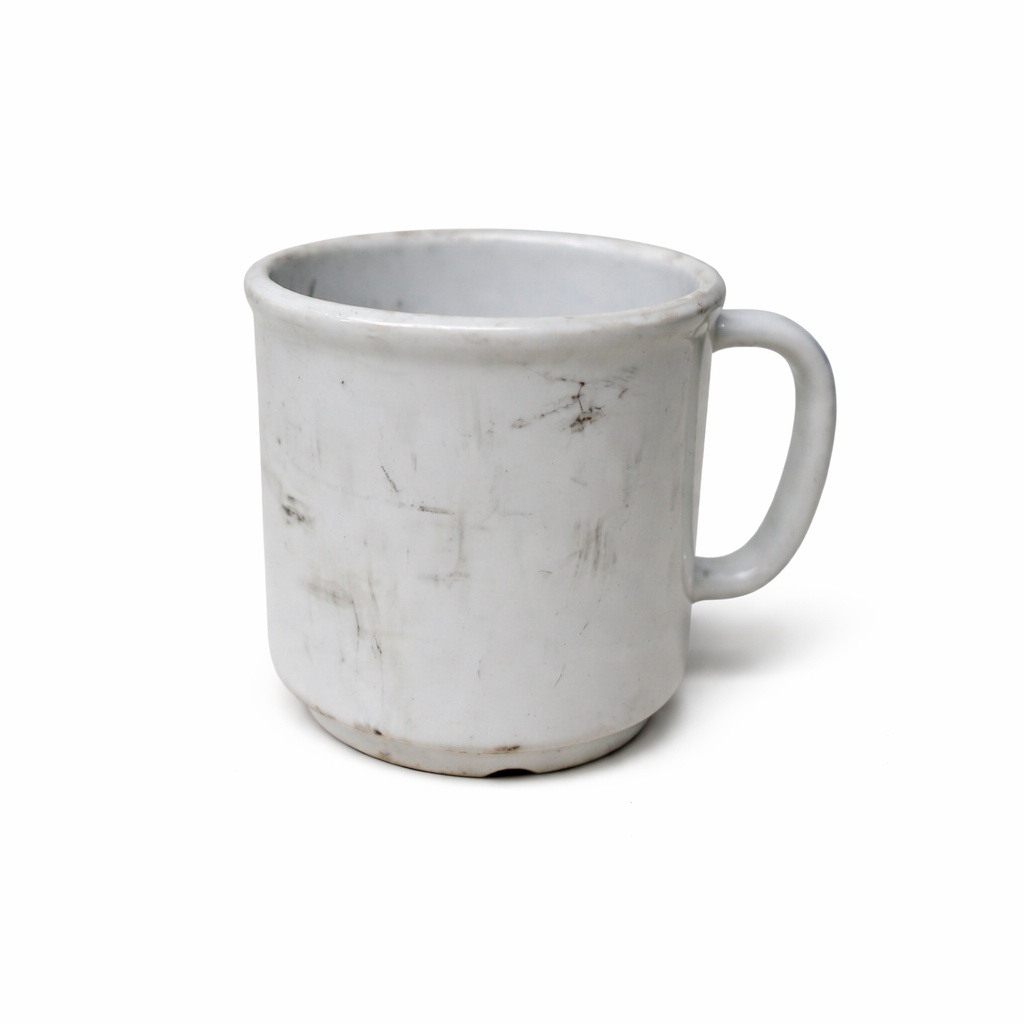}{12}\hspace{\gapH}%
        \objimg{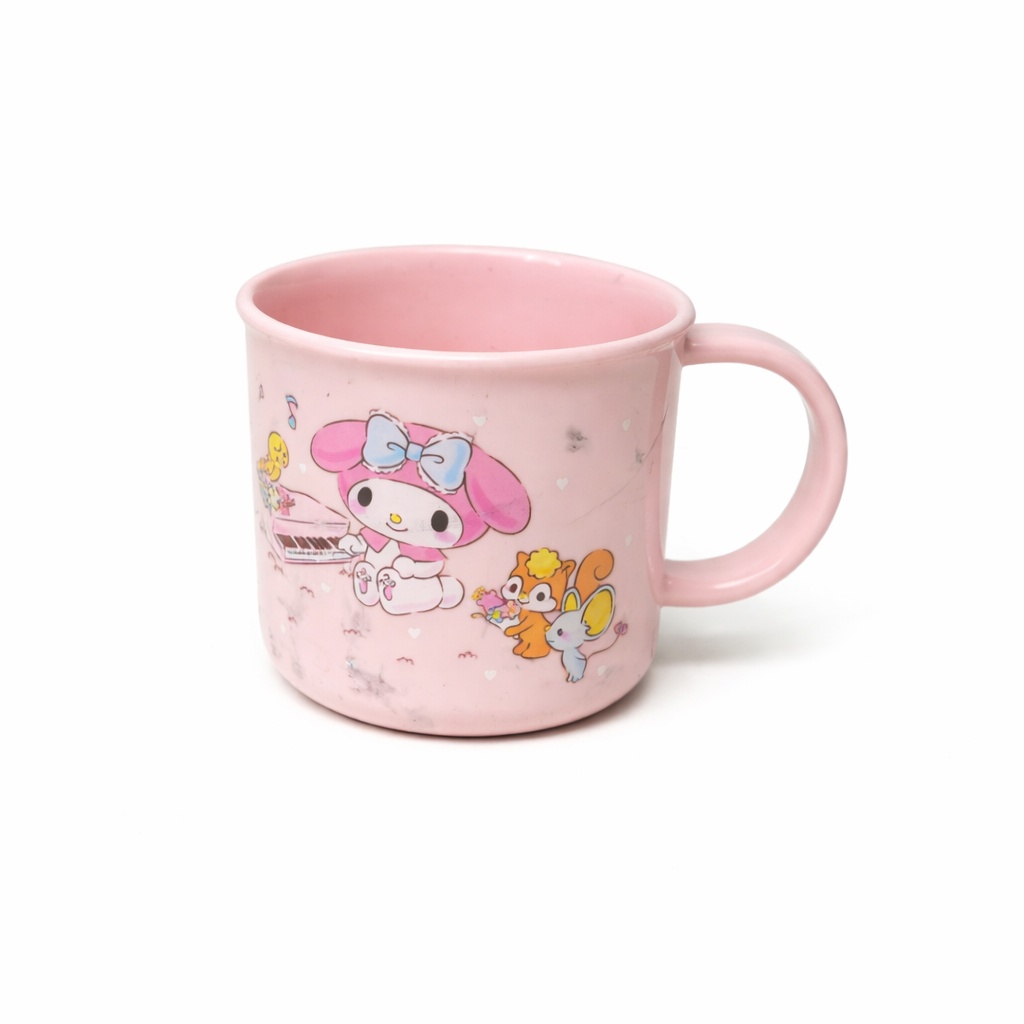}{13}
      \end{minipage}%
    }

    \caption{Mugs used during training/testing (labelled 1--13). 1-10 were used in training for both \textbf{Reorient} and \textbf{Pull}. 11-13 was used in testing for \textbf{Reorient}, and 11-12 was used in testing for \textbf{Pull}.}
    \label{fig:mugs-train-test}
\end{figure}

\begin{figure}[!t]
    \setlength{\imgw}{0.18\textwidth}
    \setlength{\gapH}{2pt}
    \setlength{\gapV}{1pt}
    \setlength{\totw}{5\imgw}
    \addtolength{\totw}{4\gapH}

    \fbox{%
      \begin{minipage}{\totw}
        \setlength{\parskip}{0pt}

        \noindent
        \objimg{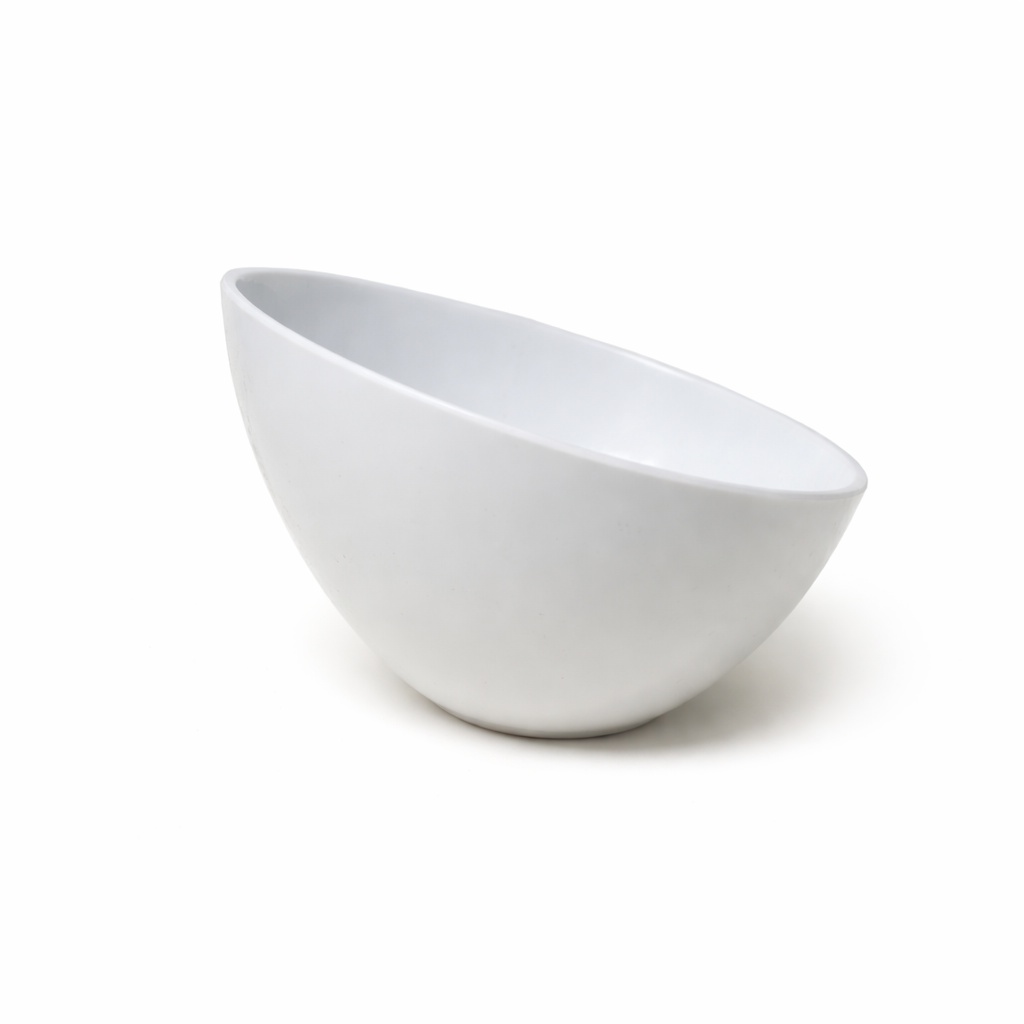}{1}\hspace{\gapH}%
        \objimg{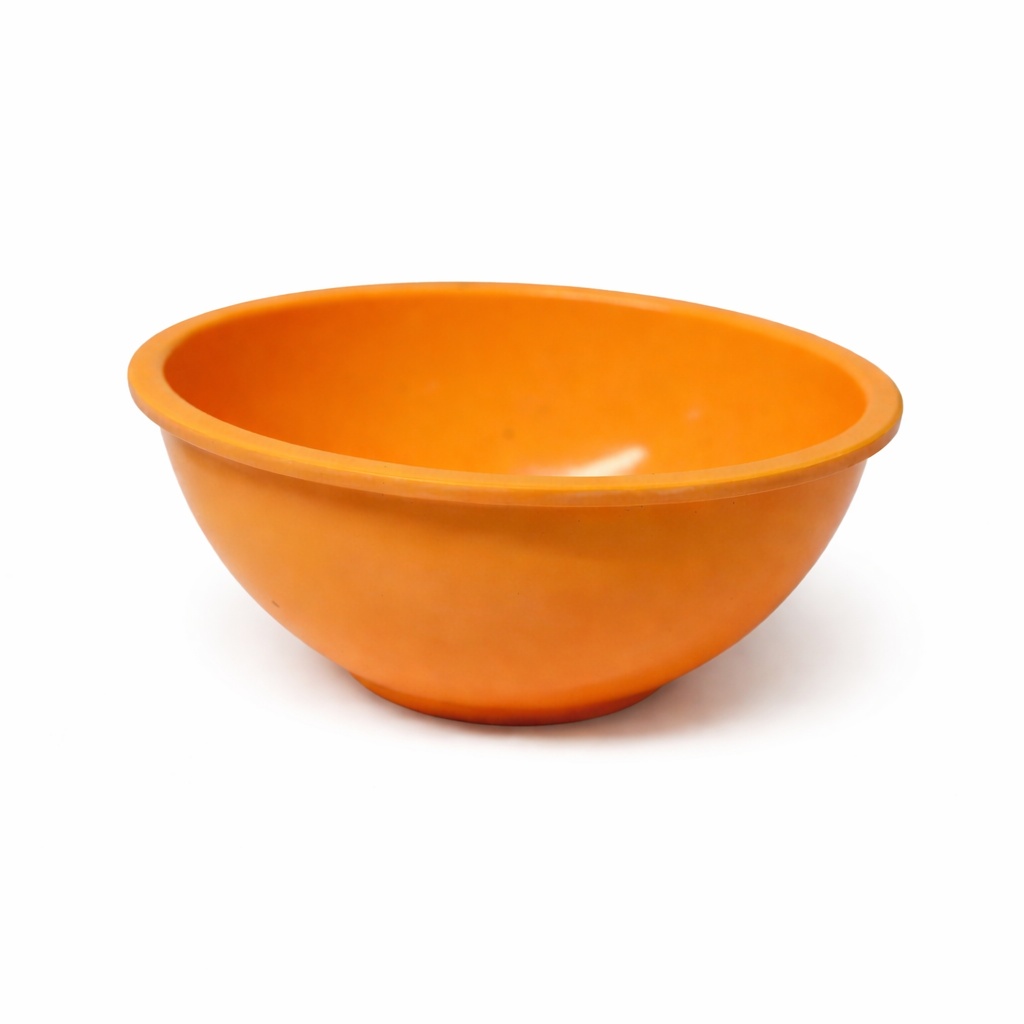}{2}\hspace{\gapH}%
        \objimg{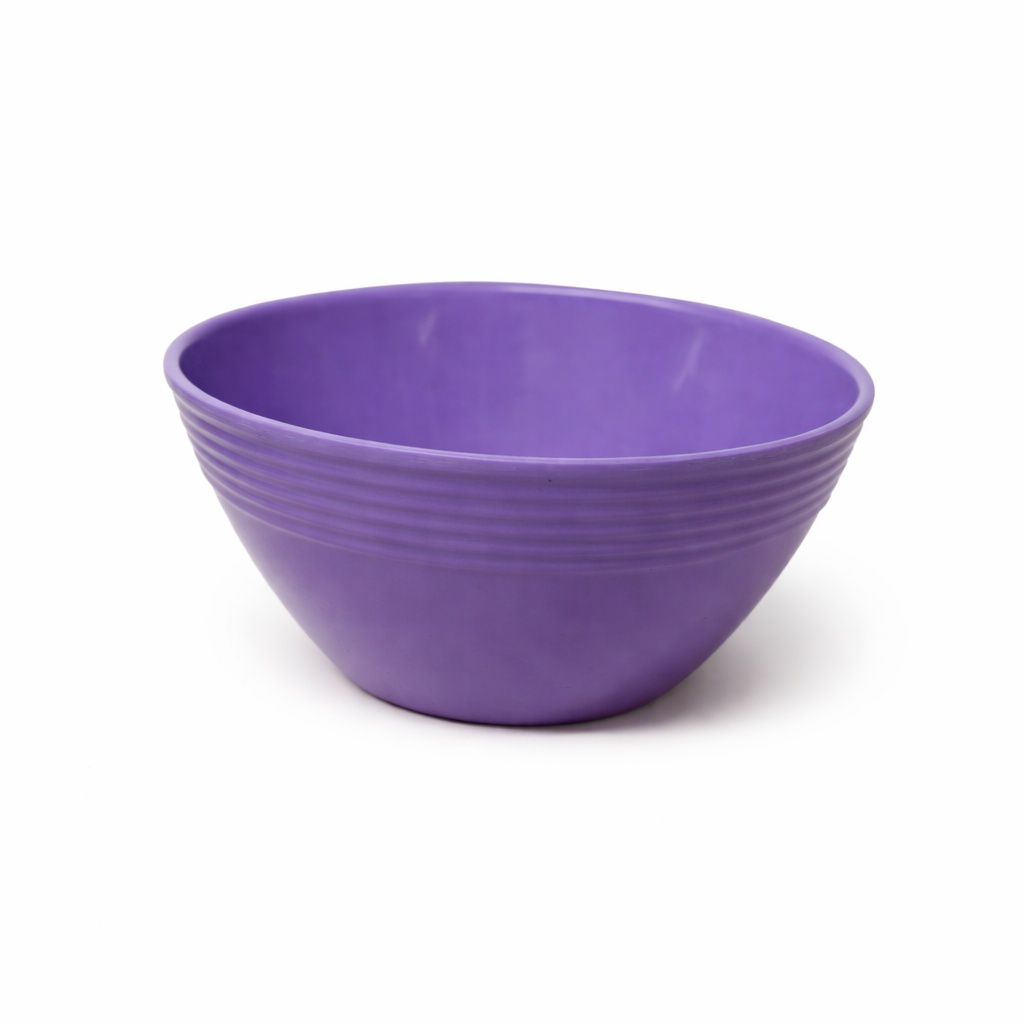}{3}\hspace{\gapH}%
        \objimg{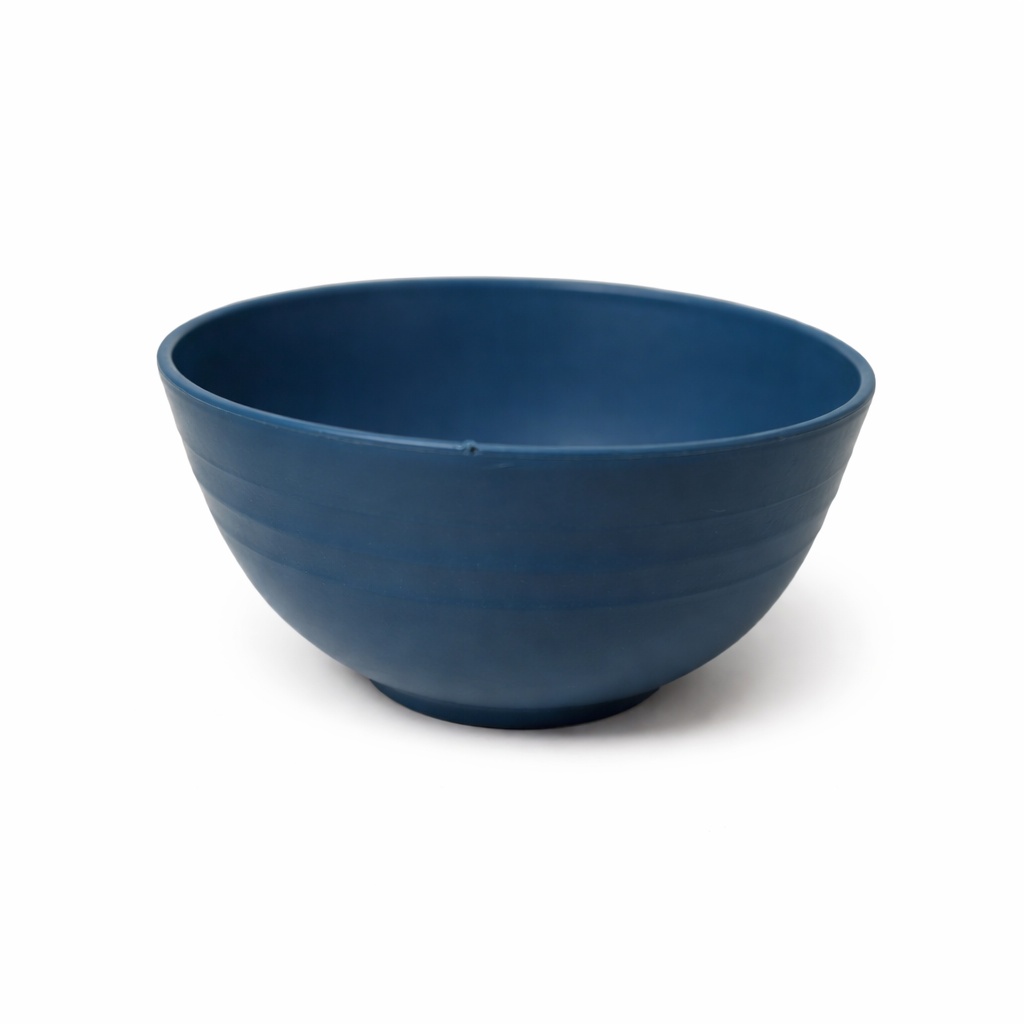}{4}\hspace{\gapH}%
        \objimg{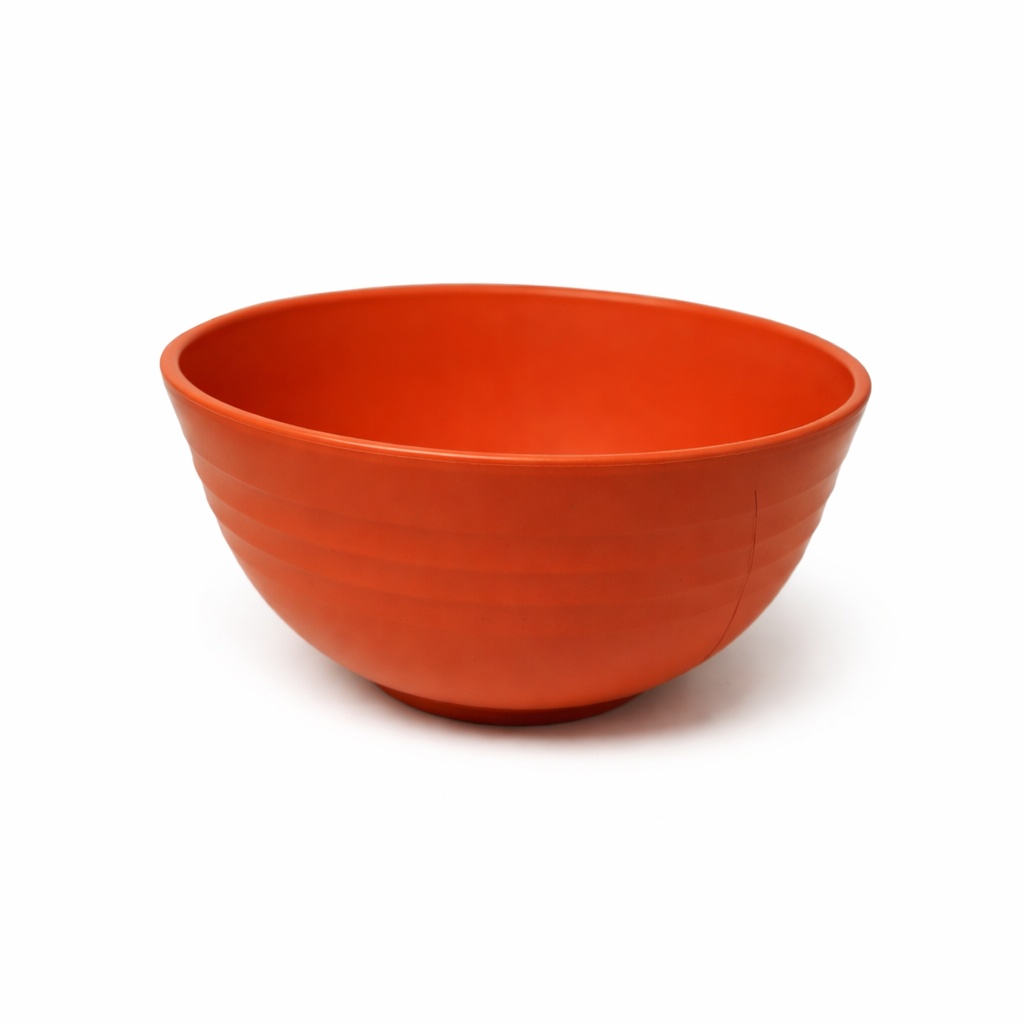}{5}

        \par\vspace{\gapV}%
        \noindent
        \objimg{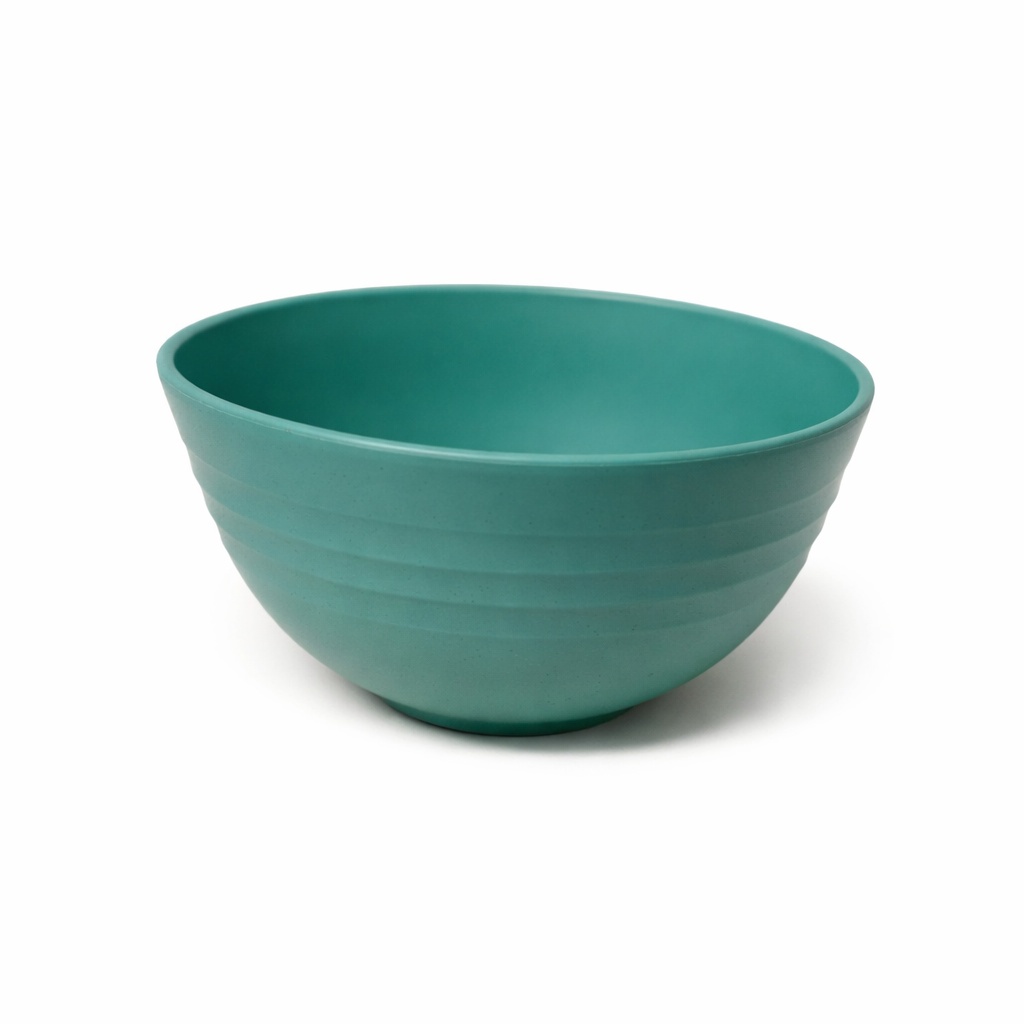}{6}\hspace{\gapH}%
        \objimg{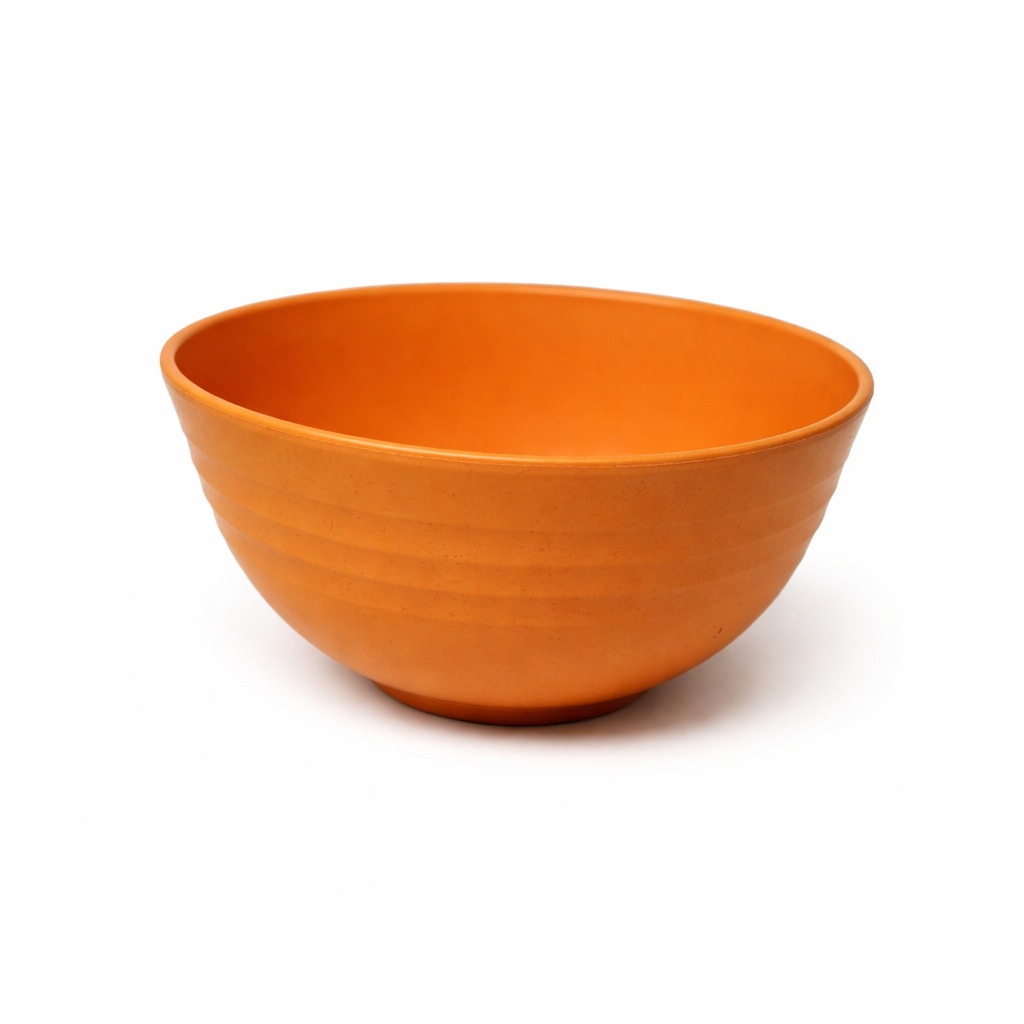}{7}\hspace{\gapH}%
        \objimg{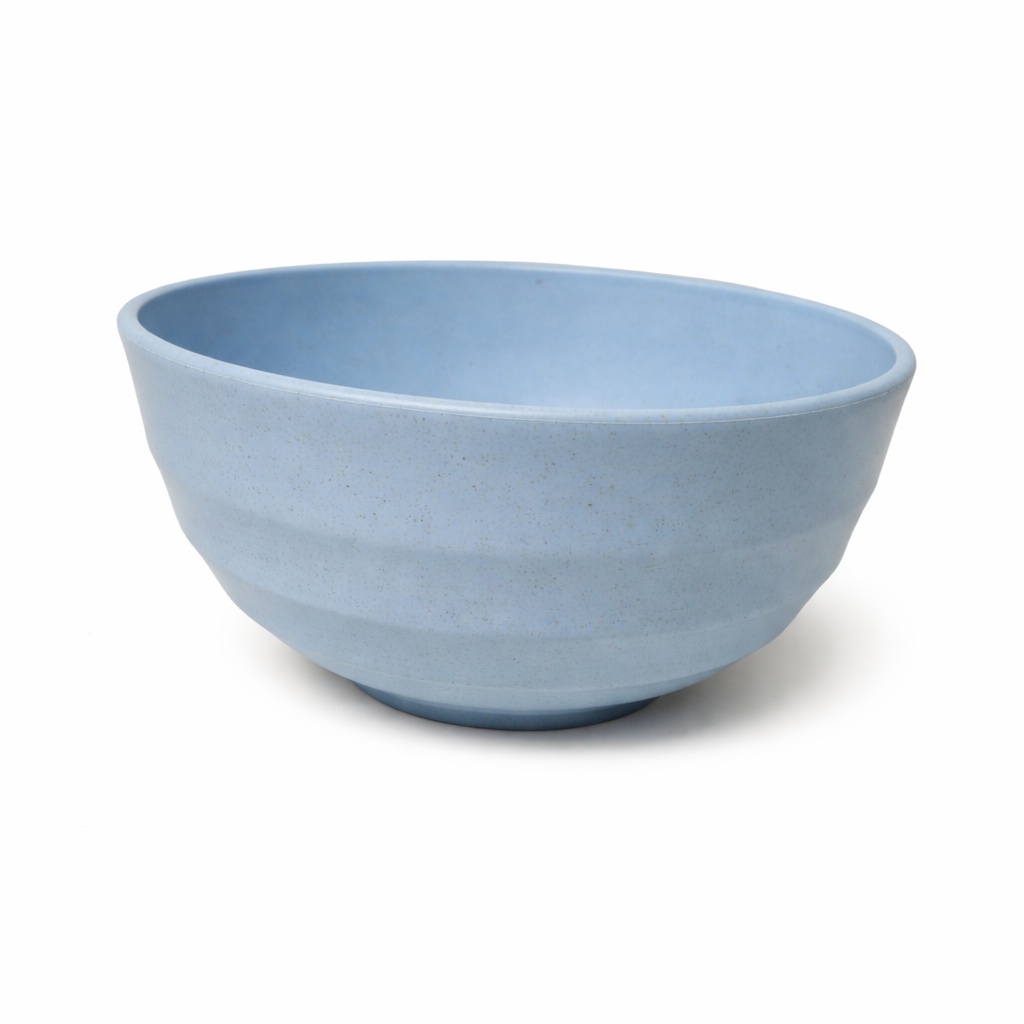}{8}\hspace{\gapH}%
        \objimg{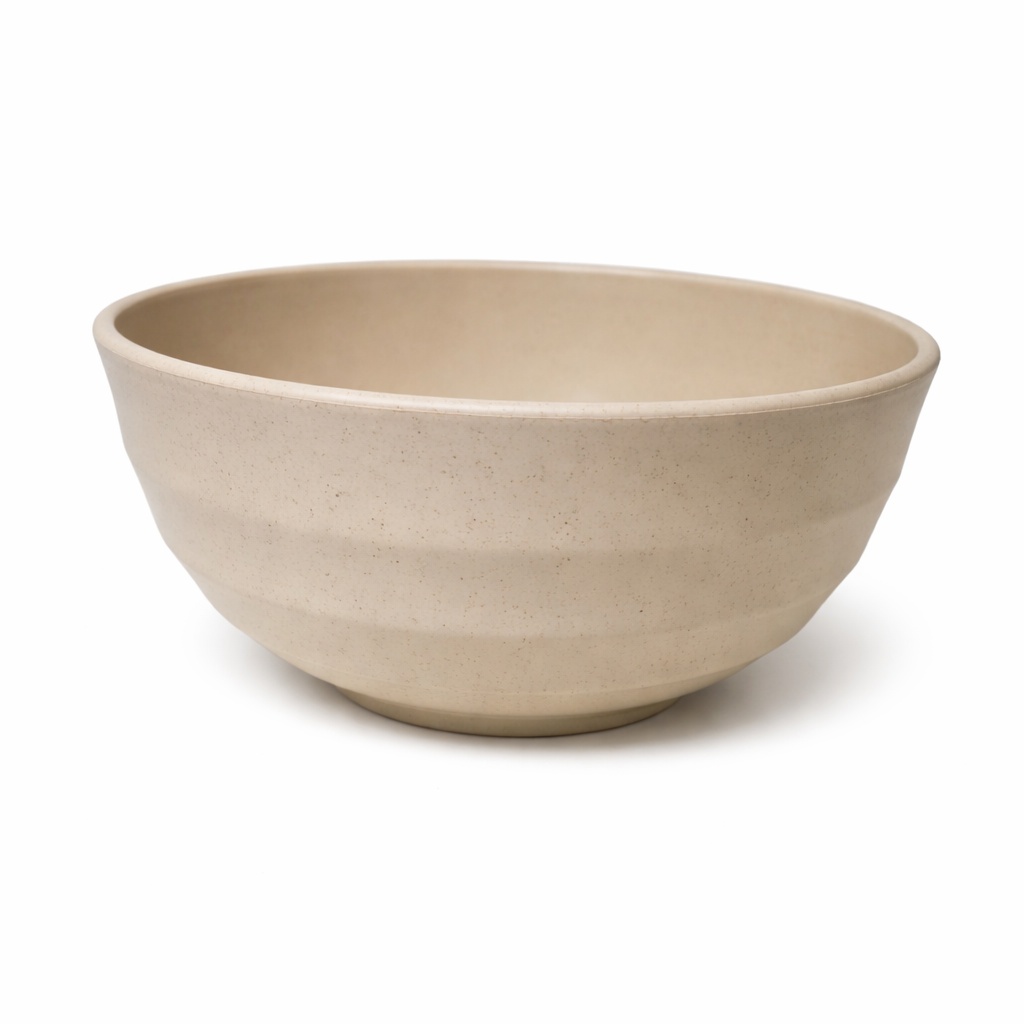}{9}\hspace{\gapH}%
        \objimg{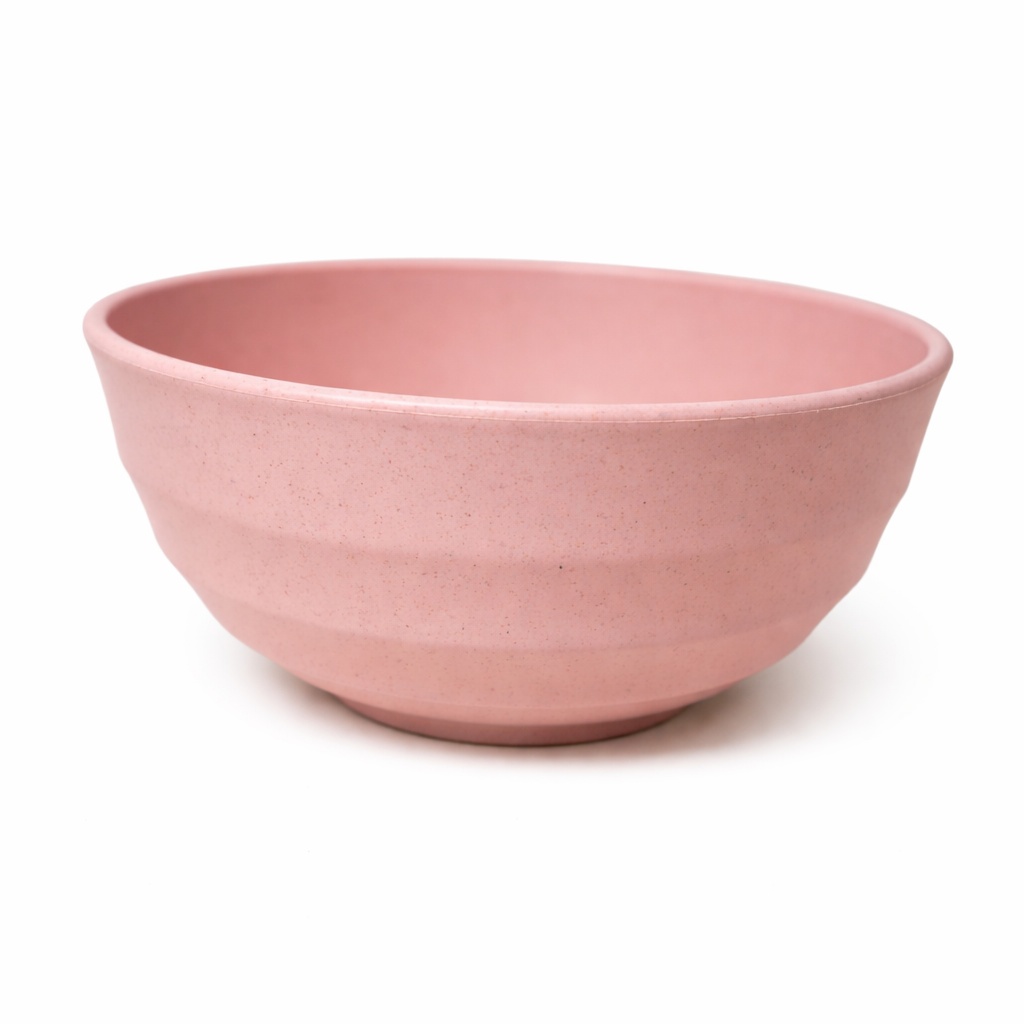}{10}%
      \end{minipage}%
    }

    \caption{Bowls used during training/testing (labelled 1--10). Pairs \{(2, 3), (2, 8), (9, 10), (3, 9), (1, 3), (5, 6), (4, 5), (6, 7), (4, 7), (8, 9)\} and pairs \{(5, 7), (4, 6)\} were used in training and testing, respectively, for \textbf{Stack}. Pairs \{(4, 8), (2, 10), (5, 9), (1, 9), (3, 8), (5, 8), (3, 9), (4, 9), (5, 10), (6, 10)\} and pairs \{(2, 8) (3, 10)\} were used in training and testing, respectively, for \textbf{Pour}.}
    \label{fig:bowls-train-test}
\end{figure}

\begin{figure}[!t]
    \setlength{\imgw}{0.18\textwidth}
    \setlength{\gapH}{2pt}
    \setlength{\gapV}{1pt}
    \setlength{\totw}{5\imgw}
    \addtolength{\totw}{4\gapH}

    \fbox{%
      \begin{minipage}{\totw}
        \setlength{\parskip}{0pt}

        \noindent
        \objimg{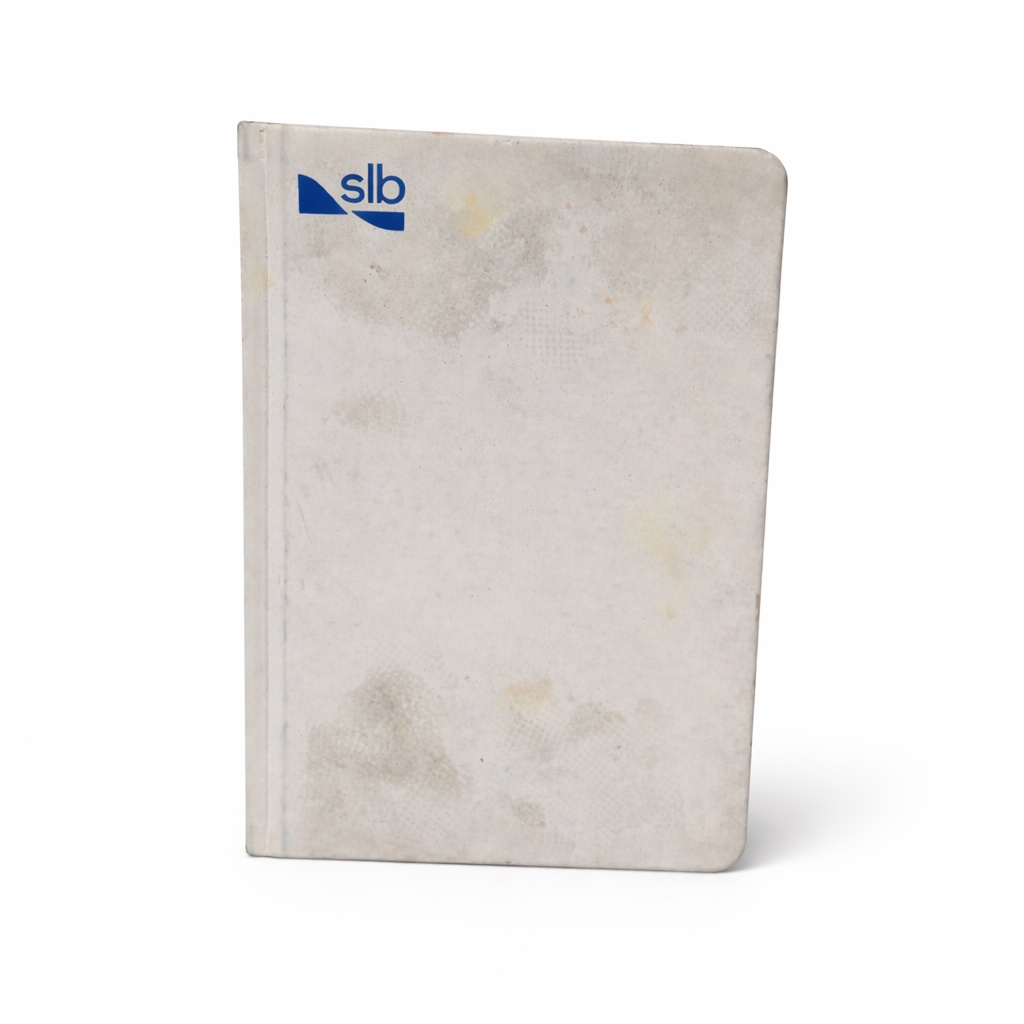}{1}\hspace{\gapH}%
        \objimg{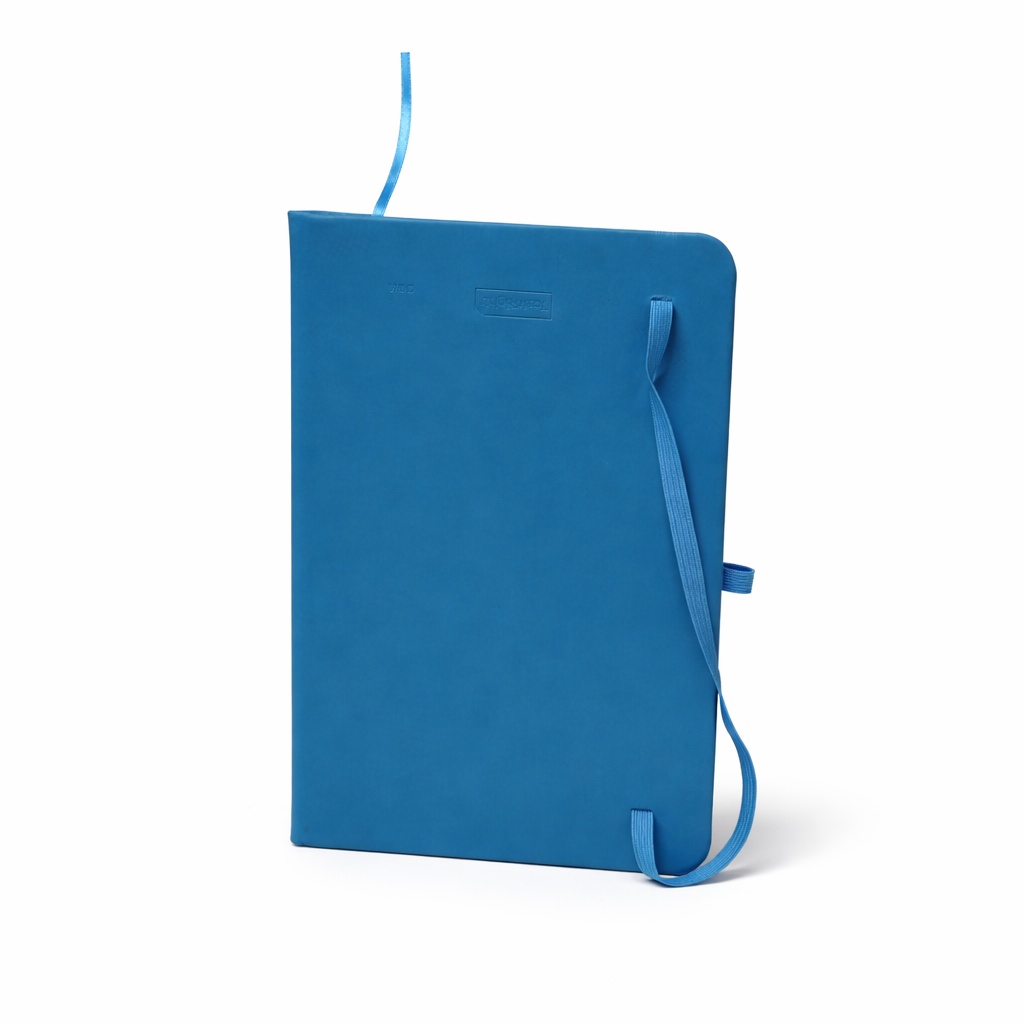}{2}\hspace{\gapH}%
        \objimg{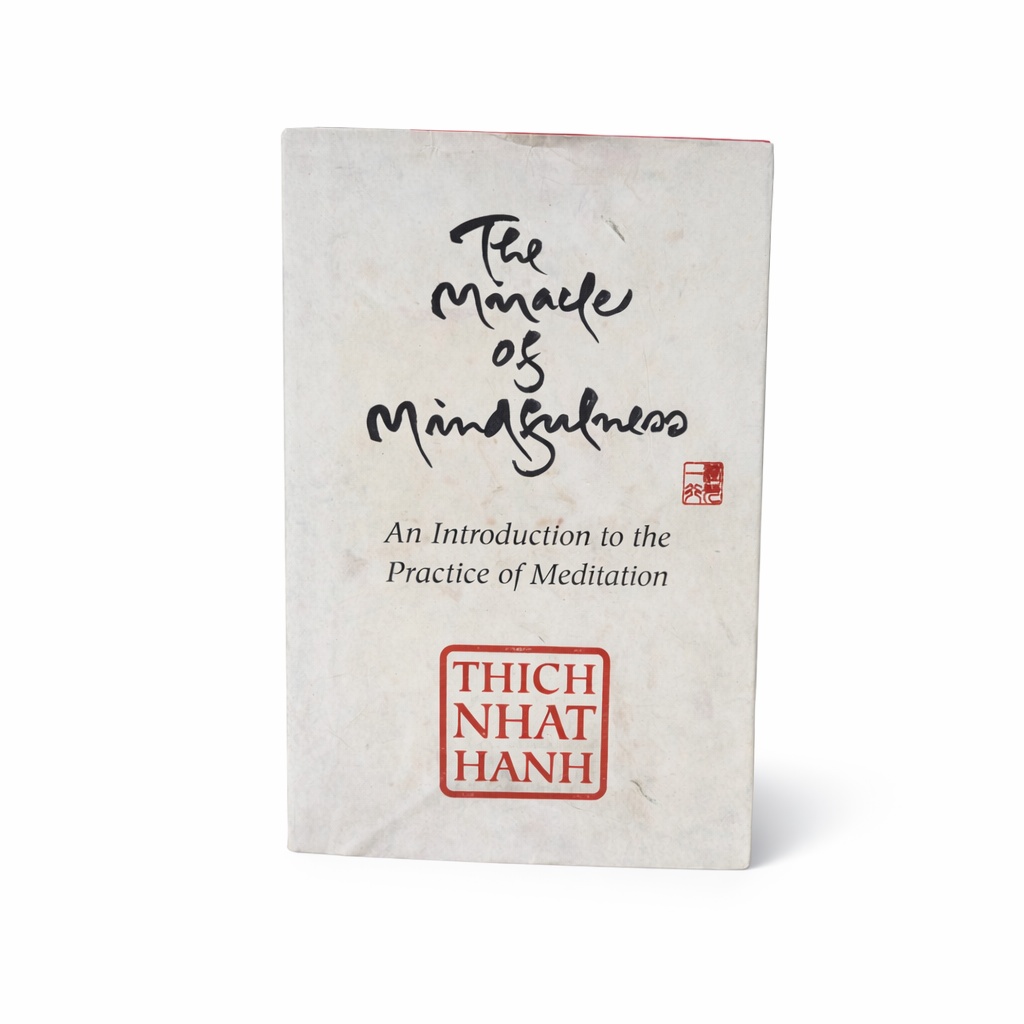}{3}\hspace{\gapH}%
        \objimg{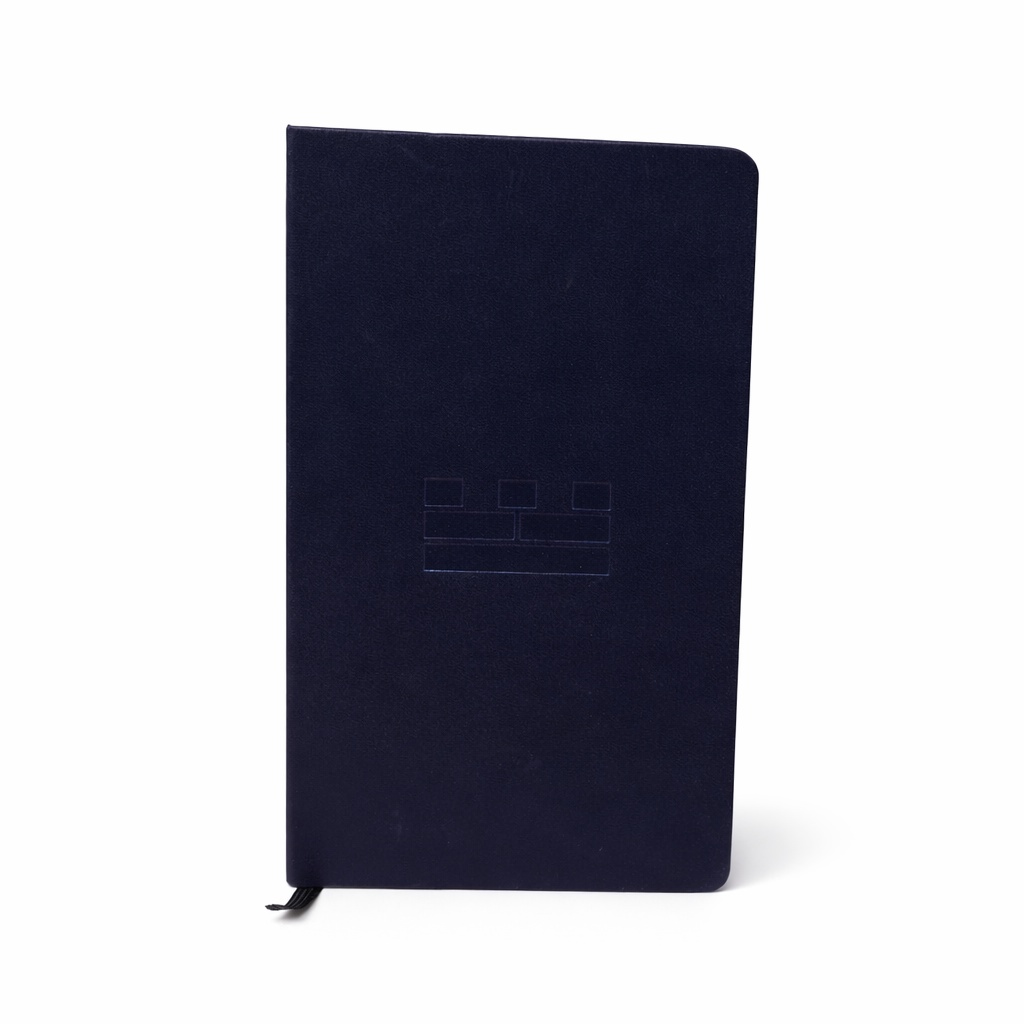}{4}\hspace{\gapH}%
        \objimg{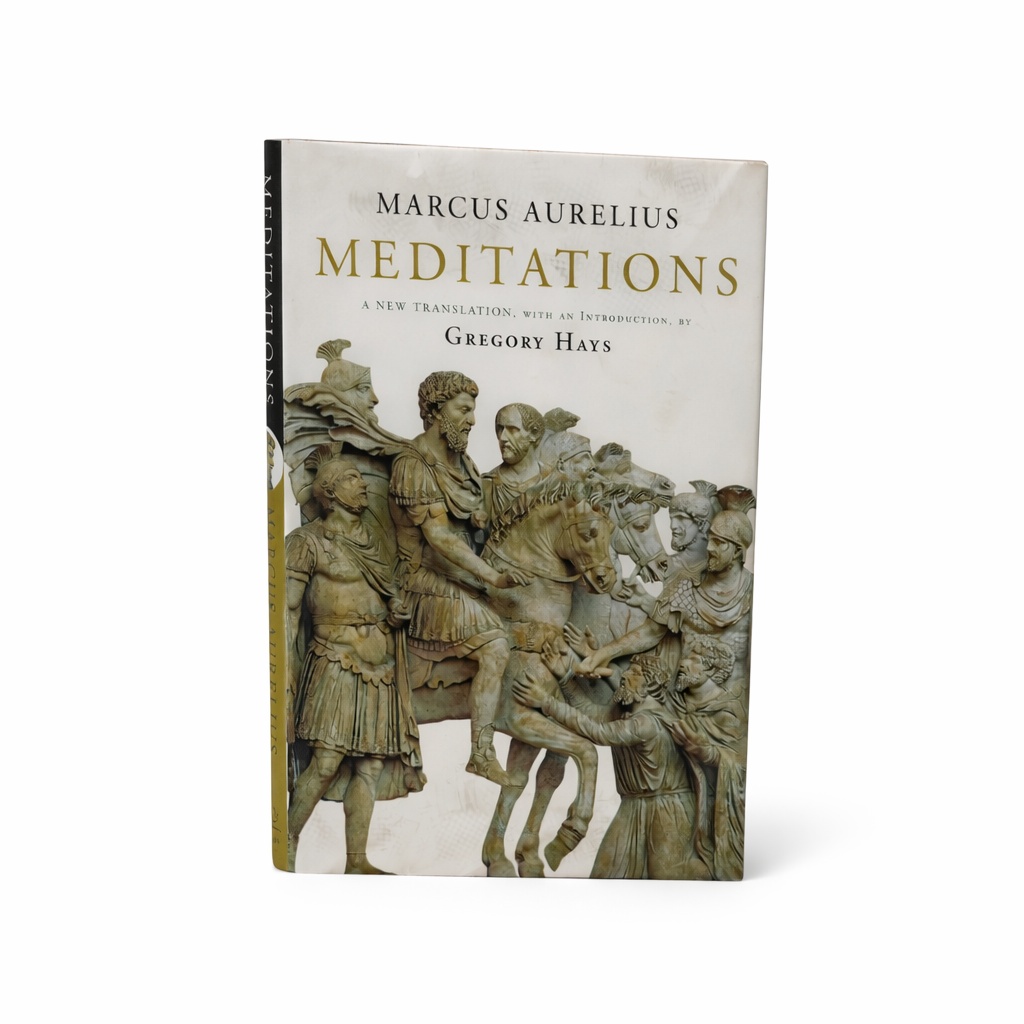}{5}%

        \noindent
        \objimg{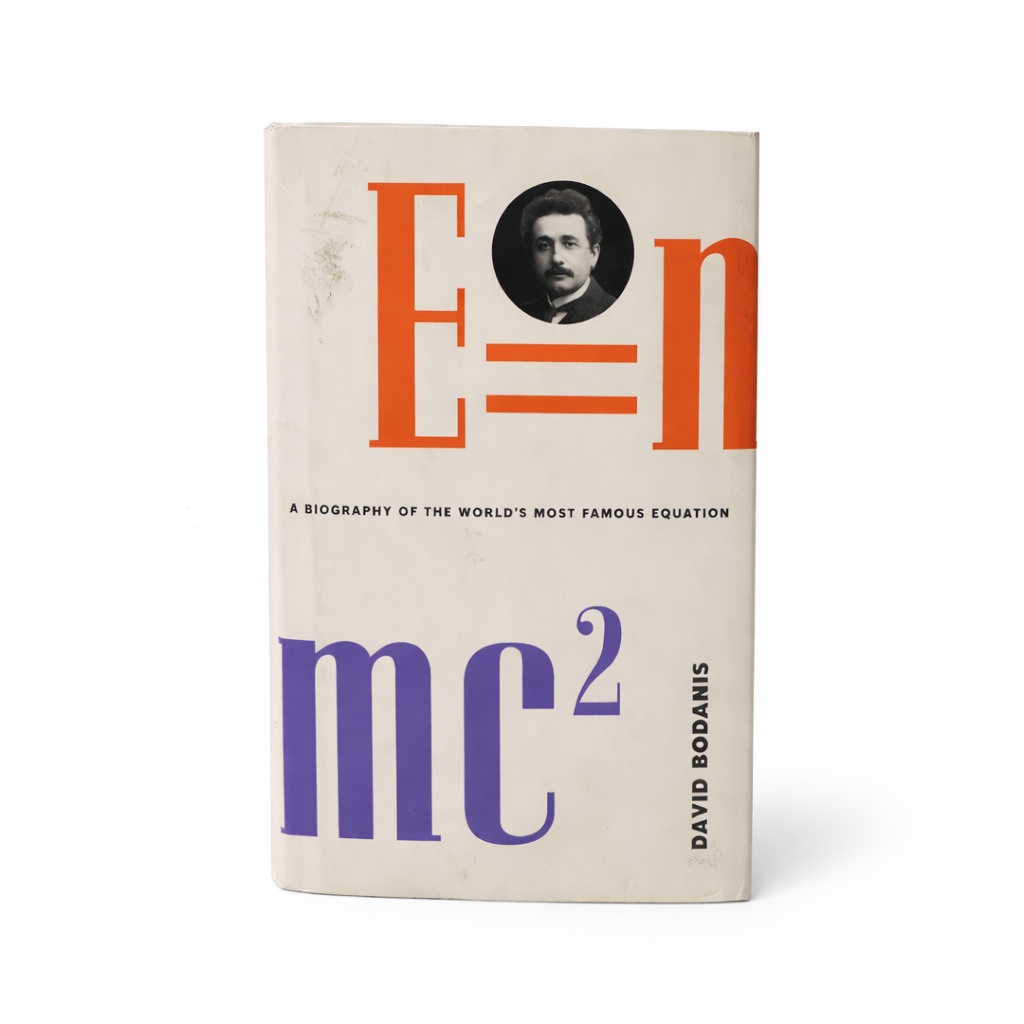}{6}\hspace{\gapH}%
        \objimg{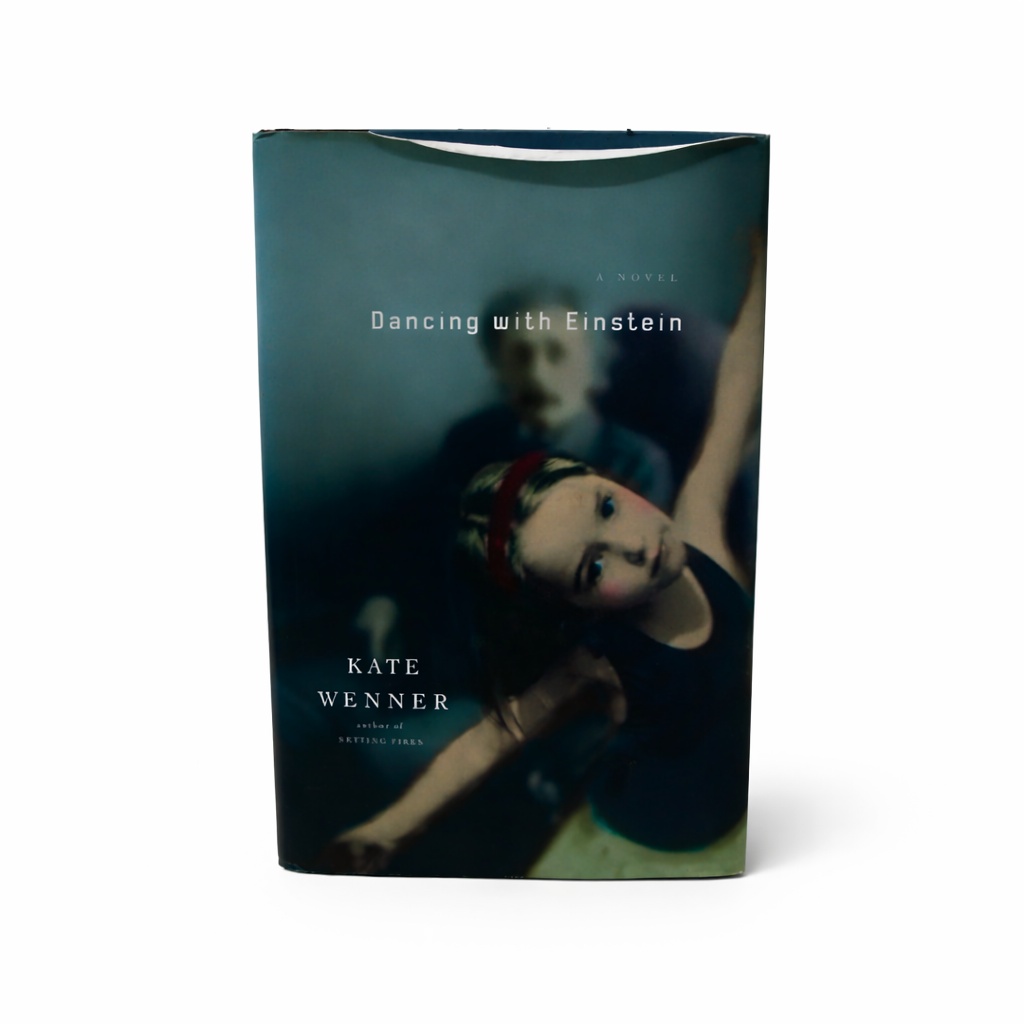}{7}\hspace{\gapH}%
        \objimg{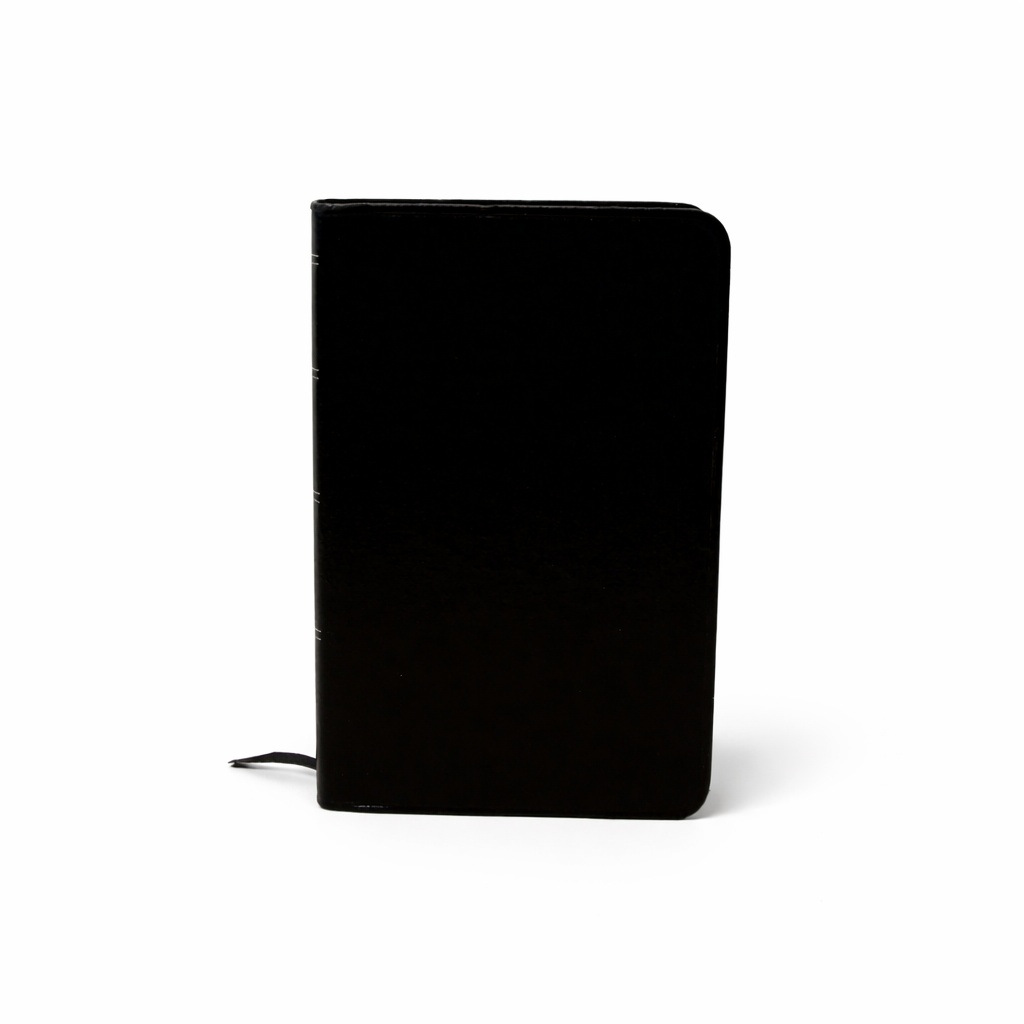}{8}\hspace{\gapH}%
        \objimg{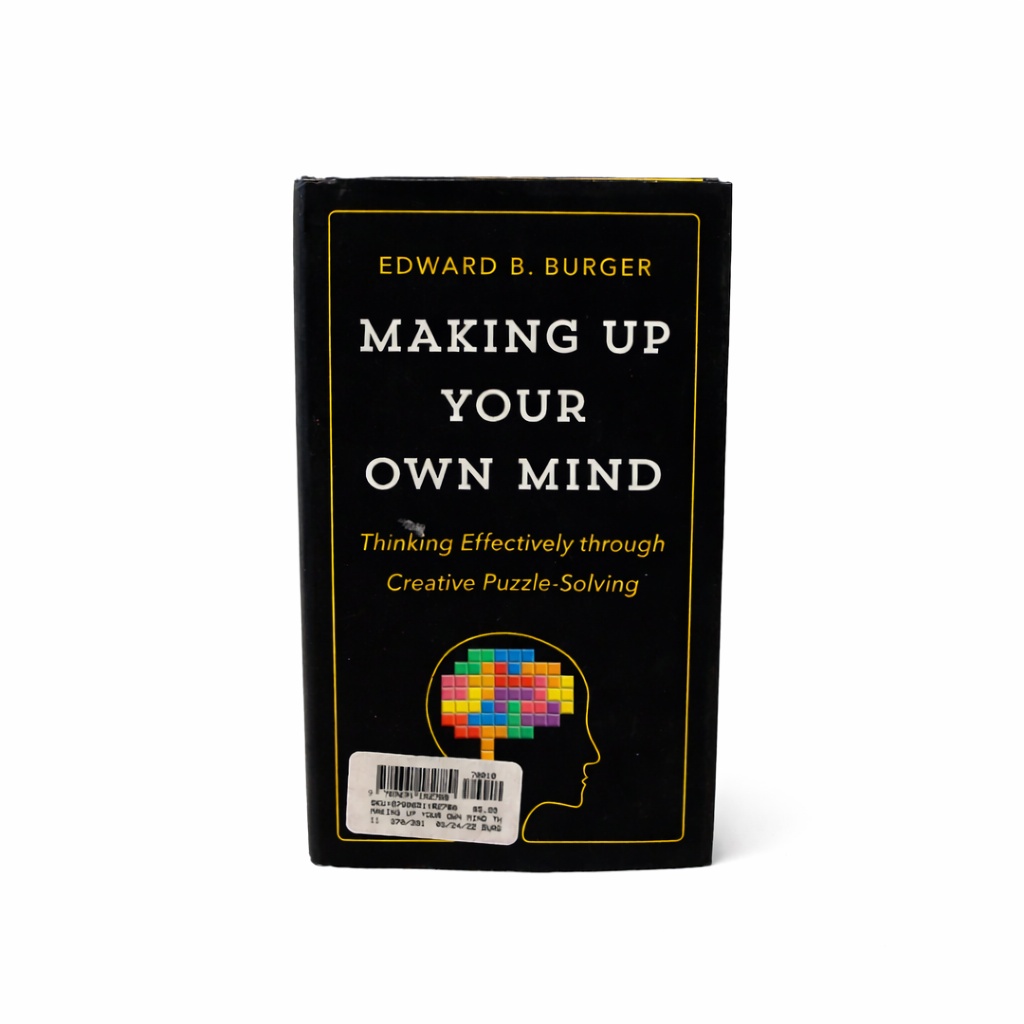}{9}\hspace{\gapH}%
        \objimg{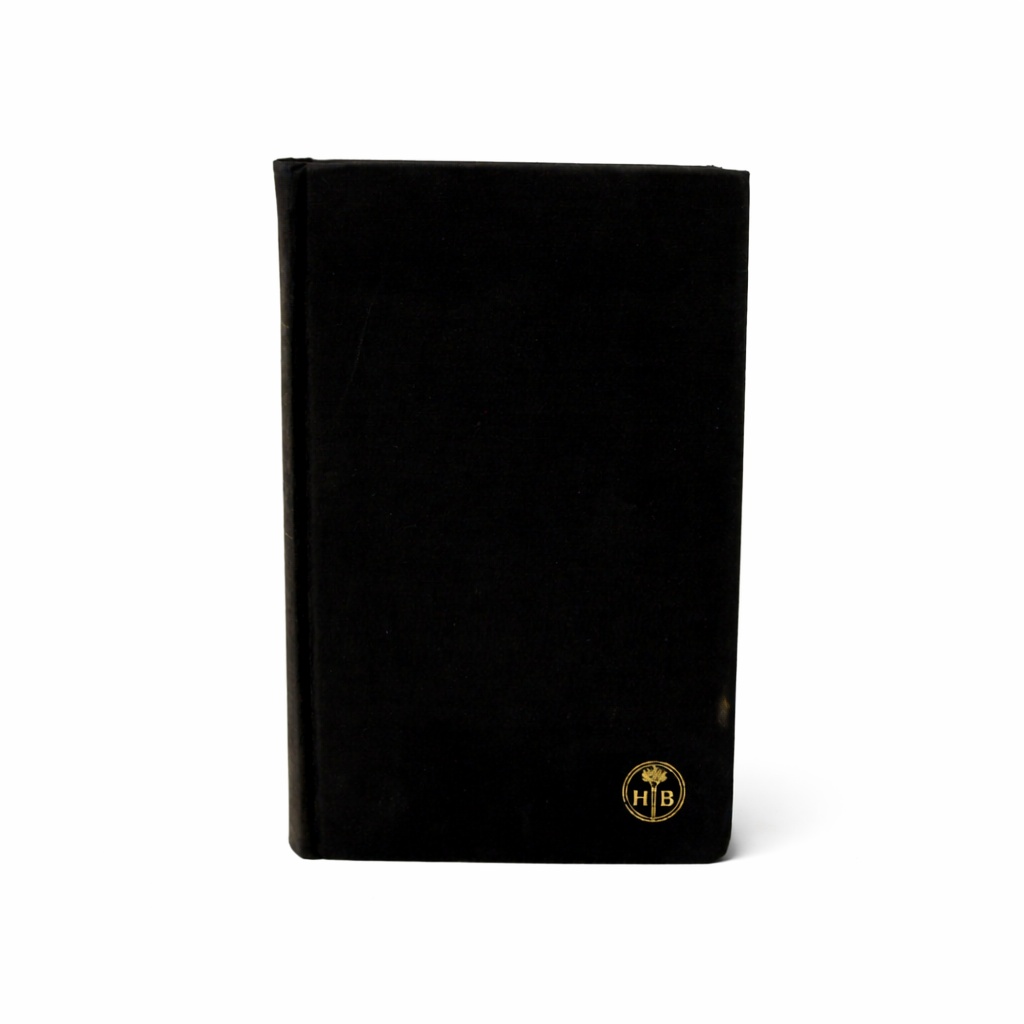}{10}%

        \par\vspace{\gapV}%
        \noindent
        \objimg{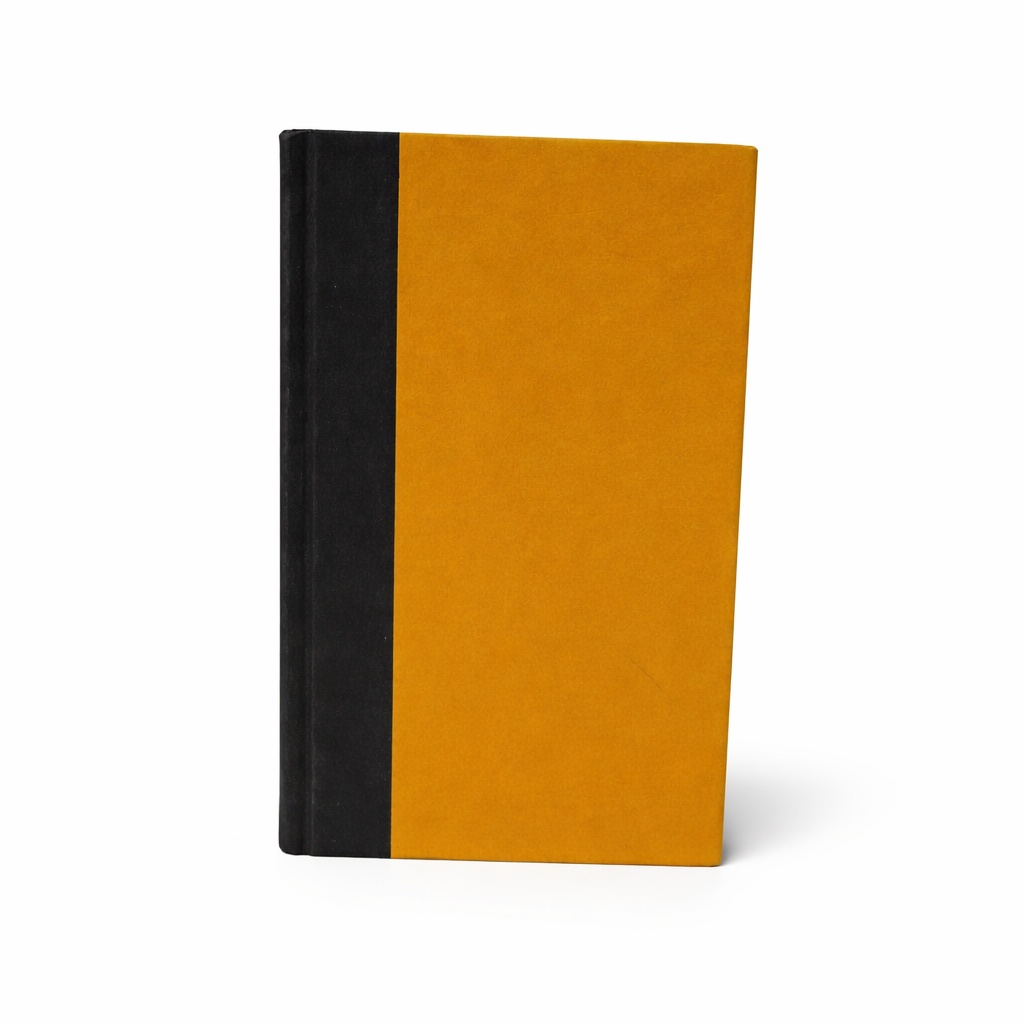}{11}\hspace{\gapH}%
        \objimg{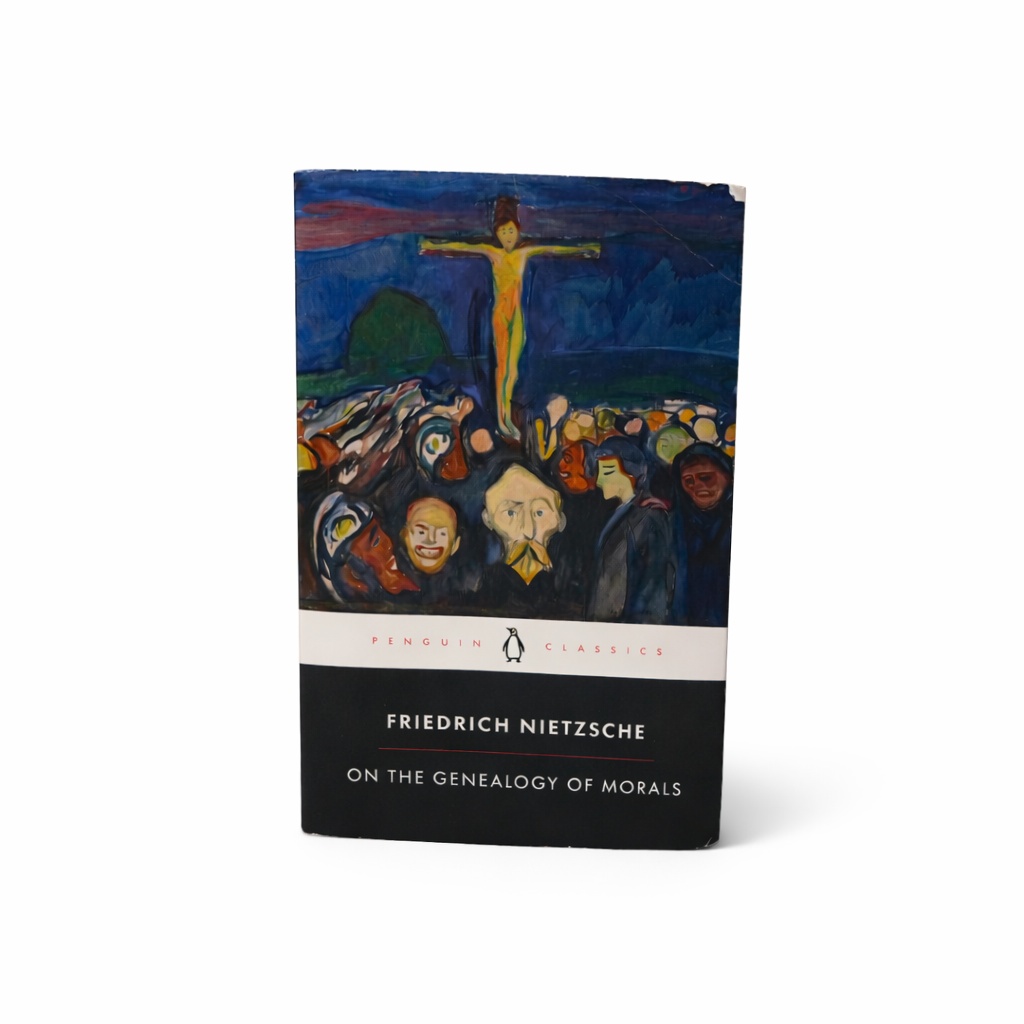}{12}%
      \end{minipage}%
    }

    \caption{Books used during training/testing (labelled 1--12). 1-10 were used in training and 11-12 were used in testing for \textbf{book}.}
    \label{fig:books-train-test}
\end{figure}

\section{Results}
\label{supp:results}
We report policy performance under a unified evaluation setup across all six tasks and 15 trials each across four unseen test environments (Sec. \ref{supp:tasks}, \ref{supp:environment_vis}). For each training setting (3, 5, or 10 training environments), we select the best checkpoint. Tables \ref{table:results_1}, \ref{table:results_robot_only} summarize per-task success rates and overall trends in how increased training diversity affects generalization.

\begin{table}[t]
\caption{\textbf{Human Cotraining success rate per task.} This table gives the quantitative results for the Human Cotraining method from Figure \ref{fig:human_vs_robot_scaling} broken down by task. Mean success rates are reported with 95\% Clopper--Pearson confidence intervals.}
\label{table:results_1}
\centering
\setlength{\tabcolsep}{5.5pt}
\small
\begin{tabular}{lccccc}
\cmidrule(lr){1-6}
\textbf{} & \textbf{Mean [95\% CI]} & \textbf{Env 1} & \textbf{Env 2} & \textbf{Env 3} & \textbf{Env 4}\\
\cmidrule(lr){1-6}

\multicolumn{6}{l}{\textbf{Pick}} \\
\cmidrule(lr){1-6}
3  & 50.0\% [36.8, 63.2] & 53.3\% & 53.3\% & 46.7\% & 46.7\% \\
5  & 68.3\% [55.9, 78.8] & 66.7\% & 80.0\% & 53.3\% & 73.3\% \\
10 & 88.3\% [77.4, 95.2] & 86.7\% & 100\%  & 86.7\% & 80.0\% \\
\cmidrule(lr){1-6}

\multicolumn{6}{l}{\textbf{Stack}} \\
\cmidrule(lr){1-6}
3  & 48.3\% [35.3, 61.5] & 26.7\% & 46.7\% & 73.3\% & 46.7\% \\
5  & 56.7\% [43.2, 69.4] & 66.7\% & 33.3\% & 80.0\% & 46.7\% \\
10 & 55.0\% [41.6, 67.9] & 60.0\% & 46.7\% & 60.0\% & 53.3\% \\
\cmidrule(lr){1-6}

\multicolumn{6}{l}{\textbf{Pull}} \\
\cmidrule(lr){1-6}
3  & 63.3\% [50.0, 75.2] & 46.7\% & 53.3\% & 73.3\% & 80.0\% \\
5  & 71.7\% [58.6, 82.5] & 66.7\% & 73.3\% & 80.0\% & 66.7\% \\
10 & 71.7\% [58.6, 82.5] & 73.3\% & 53.3\% & 86.7\% & 73.3\% \\
\cmidrule(lr){1-6}

\multicolumn{6}{l}{\textbf{Reorient}} \\
\cmidrule(lr){1-6}
3  & 38.3\% [26.2, 51.8] & 40.0\% & 53.3\% & 26.7\% & 33.3\% \\
5  & 55.0\% [41.6, 67.9] & 33.3\% & 66.7\% & 66.7\% & 53.3\% \\
10 & 68.3\% [55.9, 78.8] & 66.7\% & 80.0\% & 73.3\% & 53.3\% \\
\cmidrule(lr){1-6}

\multicolumn{6}{l}{\textbf{Book}} \\
\cmidrule(lr){1-6}
3  & 23.3\% [13.6, 36.6] & 26.7\% & 33.3\% & 13.3\% & 20.0\% \\
5  & 38.3\% [26.2, 51.8] & 46.7\% & 40.0\% & 40.0\% & 26.7\% \\
10 & 60.0\% [46.7, 72.4] & 80.0\% & 40.0\% & 73.3\% & 46.7\% \\
\cmidrule(lr){1-6}

\multicolumn{6}{l}{\textbf{Pour}} \\
\cmidrule(lr){1-6}
3  & 26.7\% [16.4, 40.3] & 40.0\% & 33.3\% & 0.00\% & 33.3\% \\
5  & 31.7\% [20.3, 45.9] & 46.7\% & 53.3\% & 13.3\% & 13.3\% \\
10 & 53.3\% [40.0, 66.3] & 53.3\% & 46.7\% & 66.7\% & 46.7\% \\
\bottomrule
\end{tabular}
\end{table}

\begin{table}[t]
\caption{\textbf{Robot Only success rate per task.} This table gives the quantitative results for the Robot Only method from Figure \ref{fig:human_vs_robot_scaling} broken down by task. Mean success rates are reported with 95\% Clopper--Pearson confidence intervals.}
\label{table:results_robot_only}
\centering
\setlength{\tabcolsep}{5.5pt}
\small
\begin{tabular}{lccccc}
\cmidrule(lr){1-6}
\textbf{} & \textbf{Mean [95\% CI]} & \textbf{Env 1} & \textbf{Env 2} & \textbf{Env 3} & \textbf{Env 4}\\
\cmidrule(lr){1-6}

\multicolumn{6}{l}{\textbf{Pick}} \\
\cmidrule(lr){1-6}
3  & 30.0\% [19.0, 43.1] & 33.3\% & 40.0\% & 26.7\% & 20.0\% \\
5  & 66.7\% [53.3, 78.3] & 80.0\% & 73.3\% & 66.7\% & 46.7\% \\
10 & 71.7\% [58.6, 82.5] & 66.7\% & 53.3\% & 80.0\% & 86.7\% \\
\cmidrule(lr){1-6}

\multicolumn{6}{l}{\textbf{Stack}} \\
\cmidrule(lr){1-6}
3  & 10.0\% [4.1, 19.5] & 6.67\% & 13.3\% & 13.3\% & 6.67\% \\
5  & 35.0\% [23.1, 48.4] & 40.0\% & 20.0\% & 33.3\% & 46.7\% \\
10 & 61.7\% [48.2, 73.8] & 46.7\% & 80.0\% & 46.7\% & 73.3\% \\
\cmidrule(lr){1-6}

\multicolumn{6}{l}{\textbf{Pull}} \\
\cmidrule(lr){1-6}
3  & 15.0\% [7.3, 26.5] & 0.00\% & 6.67\% & 13.3\% & 40.0\% \\
5  & 31.7\% [20.3, 45.9] & 26.7\% & 20.0\% & 40.0\% & 40.0\% \\
10 & 38.3\% [26.2, 51.8] & 26.7\% & 33.3\% & 26.7\% & 66.7\% \\
\cmidrule(lr){1-6}

\multicolumn{6}{l}{\textbf{Reorient}} \\
\cmidrule(lr){1-6}
3  & 11.7\% [5.6, 21.8] & 6.67\% & 6.67\% & 13.3\% & 20.0\% \\
5  & 25.0\% [15.1, 37.9] & 20.0\% & 26.7\% & 20.0\% & 33.3\% \\
10 & 71.7\% [58.6, 82.5] & 60.0\% & 60.0\% & 86.7\% & 80.0\% \\
\cmidrule(lr){1-6}

\multicolumn{6}{l}{\textbf{Book}} \\
\cmidrule(lr){1-6}
3  & 0.00\% [0, 5.98] & 0.00\% & 0.00\% & 0.00\% & 0.00\% \\
5  & 18.3\% [10.6, 29.8] & 13.3\% & 20.0\% & 20.0\% & 20.0\% \\
10 & 35.0\% [23.1, 48.4] & 53.3\% & 20.0\% & 46.7\% & 20.0\% \\
\cmidrule(lr){1-6}

\multicolumn{6}{l}{\textbf{Pour}} \\
\cmidrule(lr){1-6}
3  & 5.00\% [1.7, 13.7] & 0.00\% & 6.67\% & 6.67\% & 6.67\% \\
5  & 18.3\% [10.6, 29.8] & 13.3\% & 20.0\% & 6.67\% & 33.3\% \\
10 & 43.3\% [30.6, 56.8] & 13.3\% & 26.7\% & 66.7\% & 66.7\% \\
\bottomrule
\end{tabular}
\end{table}

\end{document}